\documentclass[runningheads]{llncs}

\usepackage{eccv}

\usepackage{eccvabbrv}

\usepackage{graphicx}
\usepackage{booktabs}

\usepackage{amsmath}
\usepackage{amssymb}
\usepackage{algorithm, algpseudocode}
\usepackage{bbm}
\usepackage{amsfonts}
\usepackage{mathrsfs}
\usepackage{array}
\usepackage{makecell}
\usepackage{multirow}
\usepackage{bm}
\usepackage{float}
\usepackage[export]{adjustbox} 
\usepackage{sidecap}
\sidecaptionvpos{figure}{t}
\usepackage{marvosym}

\usepackage[accsupp]{axessibility}  

\usepackage{hyperref}

\usepackage{orcidlink}

\floatname{algorithm}{\footnotesize Algorithm}

\newcommand{\PreserveBackslash}[1]{\let\temp=\\#1\let\\=\temp}
\newcolumntype{C}[1]{>{\PreserveBackslash\centering}p{#1}}
\newcolumntype{R}[1]{>{\PreserveBackslash\raggedleft}p{#1}}
\newcolumntype{L}[1]{>{\PreserveBackslash\raggedright}p{#1}}

\def\k{\bm{k}}
\def\n{\bm{n}}
\def\w{\bm{w}}
\def\x{\bm{x}}
\def\y{\bm{y}}
\def\z{\bm{z}}

\begin{document}

\title{Blind Image Deconvolution by Generative-based Kernel Prior and Initializer via Latent Encoding} 

\titlerunning{BID by Generative-based Kernel Prior and Initializer via Latent Encoding}

\author{Jiangtao Zhang\inst{1} \and
	Zongsheng Yue\inst{2}\orcidlink{0000-0002-9178-671X} \and
	Hui Wang\inst{1} \and\\Qian Zhao\inst{1}$^\textrm{\Letter}$\orcidlink{0000-0001-9956-0064} \and Deyu Meng\inst{1,3}\orcidlink{0000-0002-1294-8283}}

\authorrunning{J.~Zhang et al.}

\institute{Xi'an Jiaotong University, Shaanxi, China \and S-Lab, Nanyang Technological University, Singapore \and Pazhou Laboratory (Huangpu), Guangdong, China\\
\email{zhangjt2021@stu.xjtu.edu.cn, zsyzam@gmail.com, huiwang2019@stu.xjtu.edu.cn, \{timmy.zhaoqian, dymeng\}@xjtu.edu.cn}}

\maketitle

\renewcommand{\thefootnote}{\fnsymbol{footnote}}
\footnotetext[0]{\hspace{-2mm}Code is available at \url{https://github.com/jtaoz/GKPILE-Deconvolution}.}
\renewcommand{\thefootnote}{\arabic{footnote}}
\thispagestyle{empty}

\begin{abstract}
Blind image deconvolution (BID) is a classic yet challenging problem in the field of image processing. Recent advances in deep image prior (DIP) have motivated a series of DIP-based approaches, demonstrating remarkable success in BID. 
However, due to the high non-convexity of the inherent optimization process, these methods are notorious for their sensitivity to the initialized kernel. To alleviate this issue and further improve their performance, we propose a new framework for BID that better considers the prior modeling and the initialization for blur kernels, 
leveraging a deep generative model. The proposed approach pre-trains a generative adversarial network-based kernel generator that aptly characterizes the kernel priors and a kernel initializer that facilitates a well-informed initialization for the blur kernel through latent space encoding. With the pre-trained kernel generator and initializer, one can obtain a high-quality initialization of the blur kernel, and enable optimization within a compact latent kernel manifold. Such a framework results in an evident performance improvement over existing DIP-based BID methods. Extensive experiments on different datasets demonstrate the effectiveness of the proposed method. 
\keywords{Generative kernel prior \and Kernel initializer \and DIP}
\end{abstract}

\section{Introduction}
\label{sec:intro}	
Blind image deconvolution (BID), also known as deblurring, is a classical problem in image processing \cite{fergus2006removing, chan1998total, cho2009fast, chen2019blind, krishnan2011blind, levin2009understanding}, which aims at recovering the latent clean image from the observed blurry counterpart. This blurry counterpart can be approximately modeled in mathematics as (assuming the blur process is uniform and spatially invariant) 
\begin{equation}\label{eq_model}
    \y=\k \otimes \x+\n,
\end{equation}
where $\y$ is the blurry image, $\x$ the underlying clean image,  $\k$ the blur kernel, $\n$ usually the additive white Gaussian noise (AWGN), and $\otimes$ denotes the 2D convolution operator. Then the goal of BID is to estimate $\x$ under this degradation model without knowing the ground truth $\k$. As is well-known, this problem is highly ill-posed, since different pairs of $\x$ and $\k$ may result in the same $\y$.

The BID task was conventionally formulated as an optimization problem, corresponding to a Bayesian posterior estimation problem from the probabilistic perspective. Thus, the maximum a postriori (MAP) \cite{krishnan2009fast, krishnan2011blind, tog2008_shan_high_quality, xu2013unnatural, cvpr2014_pan_l0text, michaeli2014blind, pan2016blind} and variational inference (VI) \cite{fergus2006removing, tip2015_zhou_variational_dirichlet_deblur, book2017_bishop_review} approaches were developed for point-wise and distribution-wise estimation, respectively. The core challenge of these approaches is to properly design the priors for both the clean image \cite{krishnan2009fast, xu2013unnatural, cvpr2014_pan_l0text, michaeli2014blind, pan2016blind} and blur kernel \cite{icassp1997_molina_dirichlet_deblur, tip2015_zhou_variational_dirichlet_deblur}. However, these handcrafted priors are relatively
subjective and may not precisely characterize the intrinsic distributions of natural images and blur kernels in the real world, which limits the performance of the deblurring algorithm. Besides, the resulting optimization is typically endowed with highly non-convex characteristics, leading to potential convergence issues such as reaching a trivial solution of delta kernel or unsatisfactory local minimum \cite{levin2009understanding}. 

Owing to their wide and successful applications, deep learning technologies have been introduced to solve the BID problem. Early explorations~\cite{chakrabarti2016neural, sun2015learning, gong2017motion} embedded deep neural networks (DNNs) into the traditional methods to facilitate the estimation of blur kernels. Recent works have pursued a more direct approach, wherein the mapping from a blurry image to its clean counterpart is directly learned \cite{nah2017deep, tao2018scale, kupyn2018deblurgan, kupyn2019deblurgan, zamir2021multi, zamir2022restormer}. The latter strategies have achieved promising results on several benchmarks, mainly attributed to the well-designed network architectures. However, these methods tend to overfit the training datasets, and thus may not generalize well to unseen scenarios, particularly for images with large and complex motion blur kernels.

Recently, deep image prior (DIP) \cite{ulyanov2018deep} has exhibited great potential in addressing the task of BID~\cite{ren2020neural, huo2023blind}, particularly under the scenario of motion blur. The core idea is to parameterize both the blur kernel and clean image as DNNs with random inputs, and then optimize these DNNs through an energy minimization problem conducted by Eq.~\eqref{eq_model}. Compared with the traditional BID methods, these approaches employed more effective DNN-oriented priors and thus achieved superior performance.
However, these methods did not fully explore the statistical characteristics of blur kernels, resulting in inaccurate kernel estimations, particularly in cases involving large kernel sizes. Additionally, since the optimization is highly non-convex, the random initialization of the blur kernel often leads to algorithmic instability and compromises the overall performance.

\begin{figure}[t]
    \centering
    \includegraphics[width=0.93\columnwidth]{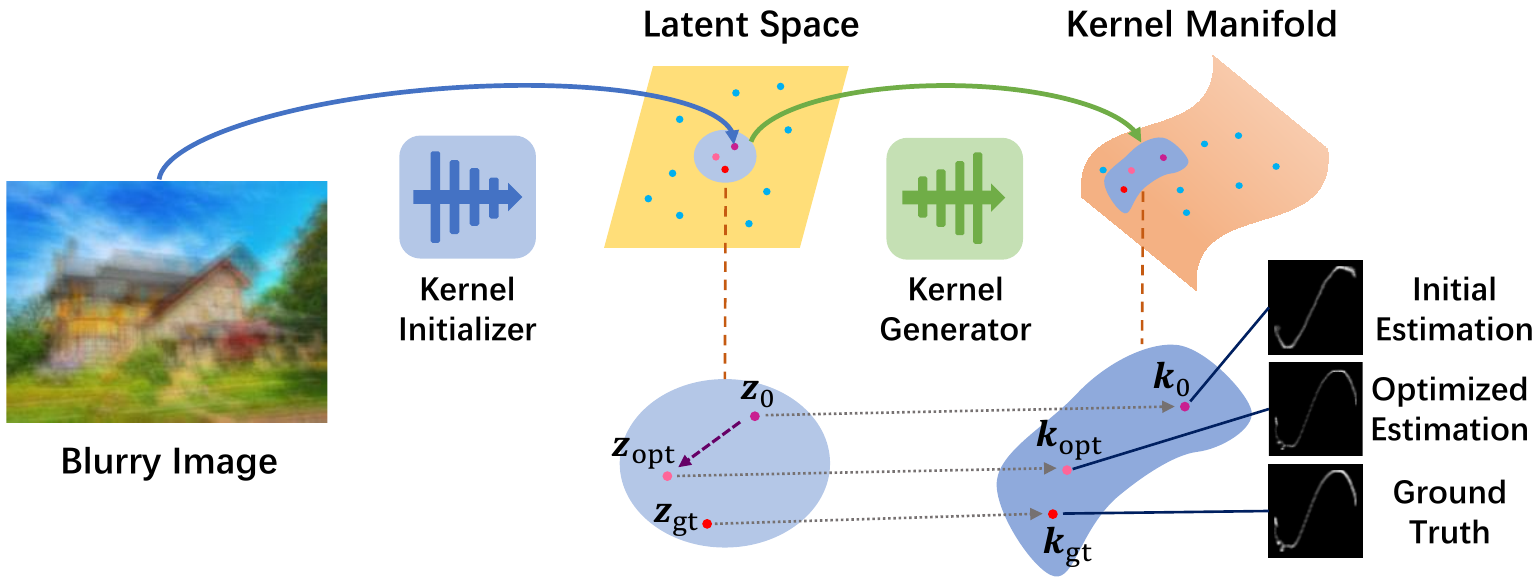}\\
    \vspace{-3mm}
    \caption{Illustration of the latent encoding-based kernel initialization and fine-tuning mechanism in our BID framework. The blur kernel is first initialized from the blurry image by a pre-trained kernel initializer via the latent code $\z_0$, which is expected to be close to the code of the ground truth kernel, $\z_{\mathrm{gt}}$. The corresponding kernels $\k_0$ and $\k_{\mathrm{gt}}$ in the kernel manifold is also expected to be close. Then the optimization is performed within the latent space, such that $\z_0$ is fine-tuned to $\z_{\mathrm{opt}}$, and the final estimated kernel $\k_{\mathrm{opt}}$ is closer to the ground truth $\k_{\mathrm{gt}}$. See Sec. \ref{sec:method} for more details.}\label{fig_idea}
    \vspace{-6mm}
\end{figure}

To alleviate the aforementioned issues of DIP-based methods, we proposed a new approach to better characterize and initialize the blur kernel with deep generative prior (DGP)~\cite{pan2021exploiting, chan2021glean, yang2021gan, wang2021towards}. Specifically, focusing on the BID of motion blurring, we train a kernel generator using generative adversarial network (GAN)~\cite{goodfellow2014generative} as kernel prior. Owing to GAN's powerful fitting capability on complex distributions, the learned generator is expected to well characterize the blur kernels in a low-dimensional latent space. Then a mapping from the blurry image to the latent code of the kernel generator is trained, acting as a kernel initializer for the subsequent BID task. Intuitively, such a kernel initializer is easier to learn as the latent kernel space is more compact than that of the original kernel. To solve the problem of BID, we first obtain a coarse initialization for the blur kernel by the initializer from the blurry image, and then jointly fine-tune the DIP for the image and the DGP for the blur kernel. The mechanism of the latent encoding-based kernel initialization and fine-tuning is illustrated in Fig.~\ref{fig_idea}. Since the kernel generator sufficiently fits the kernel distribution and the kernel initializer provides a better initial estimation, the overall BID performance can be significantly improved with faster convergence, especially for large blur kernels.

Our main contributions are summarized as follows:
\begin{itemize}
    \item We use GAN to learn the DGP for blur kernels, which can better depict the kernel structures and thus offer a more effective and compact kernel prior.
    \item We propose to learn a kernel initializer that maps from the blurry image to the latent code of the corresponding kernel. Attributed to the compactness of the latent kernel space, the proposed initializer can provide a more accurate kernel initialization for the subsequent BID process.	
    \item By harnessing the designed kernel prior and initializer, our BID method achieves state-of-the-art (SotA) performance on the challenging benchmark dataset provided by Lai et al. \cite{lai2016comparative}, especially with large blur kernels.
\end{itemize}


\section{Related work}\label{sec:related_work}

\subsection{Blind image deconvolution}
BID methods can be roughly divided into two main categories, namely optimization-based methods and deep learning-based methods.

\noindent {\bf Optimization-based BID methods.} Due to its high ill-posedness, the solution spaces of both the clean image and blur kernel should be properly constrained to make the BID problem solvable. To this aim, traditional optimization-based methods mainly focus on designing appropriate priors for clean images and blur kernels, and plenty of studies have been conducted. 
For example, total variation \cite{chan1998total}, hyper-Laplacian prior \cite{krishnan2009fast}, $l_1/l_2$-norm \cite{krishnan2011blind}, and transform-specific $l_1$-norm \cite{cai2009blind, xu2013unnatural} were considered to model the image prior in the gradient domain, and patch-based priors \cite{michaeli2014blind, sun2013edge}, low-rank prior \cite{ren2016image} and dark/bright channel prior \cite{pan2016blind, yan2017image} were directly applied in the image domain. As for blur kernels, the non-negative and normalization constraints result in a basic regularization. In addition, focusing on the properties of kernel, sparsity prior \cite{pan2017deblurring} and spectral prior \cite{liu2014blind} have been proposed. To get an ideal solution, tricks such as delayed normalization \cite{perrone2014total} and multi-scale implementation \cite{zuo2016learning} for blur kernels are also suggested. 
However, these methods highly rely on handcrafted priors, which may not accurately characterize the intrinsic properties of the clean images and blur kernels, and thus are less competitive in this deep learning era.

\noindent {\bf Deep learning-based BID methods.}
Having witnessed great success in a wide range of applications, deep learning techniques have also been adopted for the BID task. Early methods attempted to replace certain components of the traditional optimization-based methods with DNNs, by virtue of their flexibility. For example, some researchers introduced DNNs as kernel predictors \cite{sun2015learning, chakrabarti2016neural, gong2017motion, yue2024deepvariationa}. As the computing resources increase, it is then more popular to directly learn a mapping from the blurred image to its clean counterpart in a fully supervised way, with paired training data and well-designed DNNs \cite{bmvc15_hradis_cnn_deblur, nah2017deep, tao2018scale, kupyn2018deblurgan, kupyn2019deblurgan, pami2020_pan_physics_gan, tip2020_cai_dcp_nn, zamir2021multi, zamir2022restormer}. Though achieved SotA performance on several benchmark datasets, such as GoPro \cite{nah2017deep} and RealBlur \cite{eccv2020_rim_realblur}, these methods may encounter the limitations in generalization to images with large complex blur kernels that are not simulated in the pre-defined training sets. 

Recently, focusing on motion blurring, a new type of deep learning methods for the BID problem that leverages the deep priors (will be reviewed in the next subsection), has attracted increasing attention. Along this line, Ren et al. \cite{ren2020neural} proposed the SelfDeblur method, which adopted DIP to parameterize both the clean image and blur kernel for the first time. After that, multiple variants and extensions, mainly paying attention to the image prior, have been proposed for further improvement of performance \cite{huo2023blind, li2023self}. These methods achieved promising results in certain cases, especially that they outperform fully supervised deep learning methods on the challenging benchmark by Lai et al. \cite{lai2016comparative}. However, the properties of the blur kernel were not explicitly explored, and the performance tends to be unstable for large kernels due to the non-convex optimization. In contrast to these studies that used DIP with untrained DNNs, Asim et al. \cite{asim2020blind} proposed to pre-train generators for both the clean images and blur kernels as DGPs, and then fine-tune the generators when doing BID. 

\subsection{Deep prior for image processing}
Due to their flexibility, DNNs have been used to characterize the image priors in recent years. There are mainly two kinds of such deep priors, namely DIP and DGP, which are briefly reviewed in the following.

\noindent{\bf Deep image prior.} DIP was originally proposed by Ulyanov et al. \cite{ulyanov2018deep} who trained a DNN to approximate the single target (maybe up to a transformation) with random noise as its input. Since there are complex structures and operators involved, the DNN is expected to well depict the manifold of images, playing a similar role as the prior or regularizer in traditional image processing methods. After being raised, DIP has attracted remarkable attention and been applied to various image processing tasks (other than BID mentioned before), including natural image denoising \cite{ulyanov2018deep}, super-resolution \cite{DIP-FKP,sr_yue}, inpainting \cite{ulyanov2018deep}, image decomposition \cite{DoubleDIP}, low-light enhancement \cite{RetinexDIP}, PET image reconstruction \cite{DIP_PET} and hyperspectral image denoising \cite{S2DIP}.

\noindent{\bf Deep generative prior.} Attributed to their powerful generation capabilities, the trained deep generators, such as GANs~\cite{goodfellow2014generative}, can be seen as approximators to the distributions of images, hoping to provide reasonable priors for images. Such priors can be referred to as DGPs \cite{pan2021exploiting}. After a pre-training stage for the generator, DGP can be used as an estimator to approximate the target image via fine-tuning like DIP. There are two common ways to fine-tune DGP. The first is to fix the generator and optimize the input. 
For example, Menon et al. \cite{menon2020pulse} used this idea for photo upsampling. This strategy is also closely related to GAN-inversion \cite{xia2022gan}. The second is to jointly fine-tune the latent code and the parameters of the pre-trained generator like \cite{pan2021exploiting}, which may better locate the optimal estimation within the image manifold. 

In addition to directly modeling the images, DGP was also used to characterize the degradation operators, such as blur kernels. For example, Asim et al. \cite{asim2020blind} proposed to pre-train a variational auto-encoder (VAE) \cite{iclr2014_kingma_vae} as DGP for motion blur kernels for the BID task; and Liang et al. \cite{DIP-FKP} used the normalizing flow (NF) \cite{DinhNICENonlinear2015, DinhDensityEstimation2017, KingmaGlowGenerative2018} as the DGP for Gaussian blur kernels, applying to blind super-resolution. These studies highlight the effectiveness of DGP for blur kernels, and partially inspire our work.

\section{Proposed method}\label{sec:method}
\subsection{DIP-based BID revisit}
Before presenting our BID method, we first revisit DIP-based BID approaches, or more specifically SelfDeblur \cite{ren2020neural}, and discuss the issues we try to address.

In SelfDeblur~\cite{ren2020neural}, the clean image $\x$ in Eq.~\eqref{eq_model} was approximated by a convolutional network equipped with encoder-decoder architecture, and the blur kernel $\k$ was estimated by a fully-connected network with one hidden layer. The resulting optimization can then be formulated as
\vspace{-1mm}
\begin{equation}\label{eq_dip_bid_basic}
    \underset{\theta_k,\theta_x}{\min}\left\|G_k(\z_k;\theta_k)\otimes G_x(\z_x;\theta_x)-\y\right\|^2,
    \vspace{-2mm}
\end{equation}
where $G_k(\cdot;\theta_k)$ and $G_x(\cdot;\theta_x)$ are networks for the kernel and image, respectively, and $\z_k$ and $\z_x$ are input random noise. Note that we ignore the additional constraints imposed on the blur kernel and clean image, positing that these constraints can be readily satisfied through thoughtful network design. Thus, the problem of estimating the blur kernel $\k$ and the clean image $\x$ is reduced to the optimization of parameters $\theta_k^{\ast}$ and $\theta_x^{\ast}$, respectively.

\begin{figure}[t]
    \begin{minipage}{0.65\textwidth}
        \vspace{-3mm}
        \centering
        \includegraphics[width=0.88\textwidth]{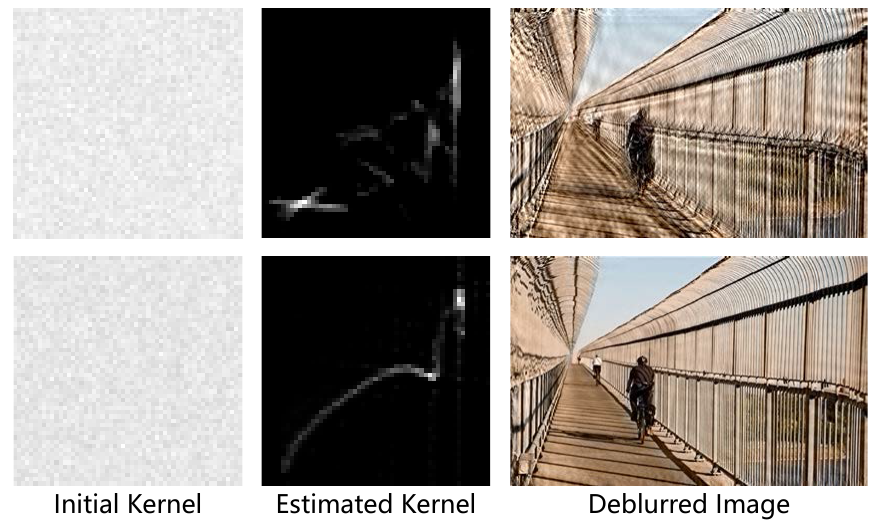}
    \end{minipage}
    \begin{minipage}{0.3\textwidth}
        \centering
        \vspace{-1mm}
        \caption{Illustration of the initialization effect of the DIP-based BID. The two rows correspond to two independent runs of SelfDeblur \cite{ren2020neural}. From left to right: the randomly initialized kernel, the finally estimated kernel, and the deblurred image.}\label{fig_intro}
    \end{minipage}
    \vspace{-8mm} 
\end{figure}

Owing to the powerful capabilities of DNNs, SelfDeblur obtained promising results on the BID task with motion blurring. Notably, it established a new SotA performance on the challenging benchmark introduced by Lai et al. \cite{lai2016comparative}. Huo et al. \cite{huo2023blind} further improved SelfDeblur by incorporating a more sophisticated DIP within the variational Bayesian framework. However, these DIP-based BID methods did not fully explore the statistical structures or priors for blur kernels, thereby limiting their performance. A more important concern arises from the non-convexity of the optimization, where the randomness of $\z_k$ causes unstable solutions, particularly when the blur kernel is large. Fig.~\ref{fig_intro} shows a typical example with two different random runs. Despite the initializations following the same distribution, the estimated blur kernels can be diverse, consequently leading to significantly different deblurring results.

The aforementioned issues redirect our focus toward the blur kernels. Therefore, this study explores more precise statistical structures for blur kernels with DGP, followed by the development of a kernel initializer in the latent space. 

\subsection{Overview of the proposed method}
Our proposed method consists of two stages. The first is the pre-training stage, aiming to learn a \emph{kernel generator} and a \emph{kernel initializer}. The second stage is to solve the deconvolution problem following the paradigm of the DIP-based BID method, while with the aid of the learned generator and initializer.

In the pre-training stage, we train a kernel generator $G_k(\cdot;\theta_k^{\ast})$ within the GAN~\cite{goodfellow2014generative} framework, where $\theta_k^{\ast}$ denotes the generator parameters after training. Once learned, it can generate a blur kernel from the randomly sampled noise $\z_k$. This means the generator indeed models the distribution of the blur kernels and thus characterizes their statistical structures. Such a learned kernel generator can be referred to as the DGP of blur kernels, with $\z_k$ serving as the kernel latent code. Next, with the generator $G_k(\cdot;\theta_k^{\ast})$ fixed, we train an encoder $E(\cdot;\theta_E^{\ast})$ that maps a blurry image $\y$ to the corresponding kernel latent code through the technique of GAN-inversion \cite{xia2022gan}. The learned encoder allows for a relatively accurate prediction for the blur kernel based on its latent code, and thus acts as an effective initializer for the kernel DGP.

\begin{figure}[t]
    \centering
    \includegraphics[width=0.93\columnwidth]{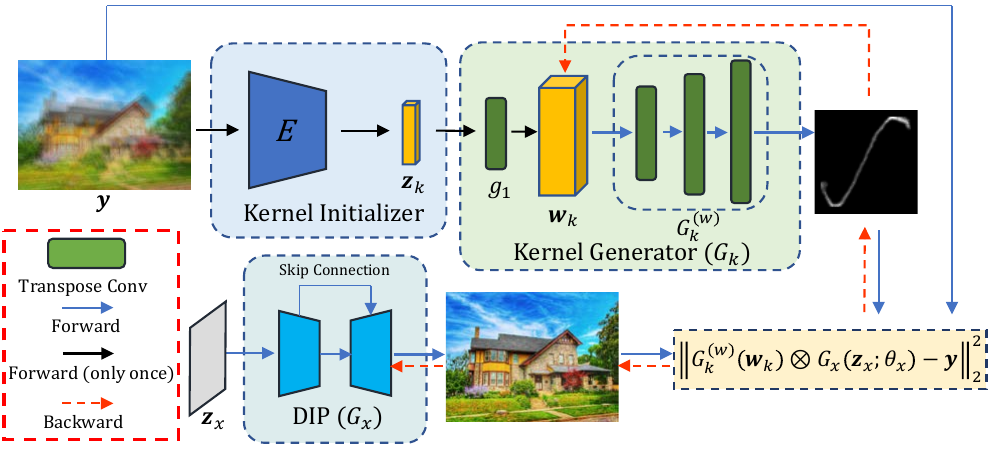}
    \vspace{-2mm}
    \caption{Overview of the proposed BID process, with the pre-trained kernel generator and initializer, of our proposed method.}\label{fig_framework}
    \vspace{-6mm}
\end{figure}

In the second stage, we first predict the kernel latent code corresponding to the input blurry image $\y$ using the pre-trained kernel initializer, which provides an initial value for the kernel DGP in the latent space. Then we jointly optimize the parameters of DIP for the clean image and the latent code (or more precisely its mapped features as discussed in Sec. \ref{sec:bid_process}) of the DGP for the blur kernel.
The overview of the proposed BID process is shown in Fig. \ref{fig_framework}.

In Sec. \ref{sec:kg_learning} and Sec. \ref{sec:ki_learning}, we provide details for learning the kernel generator and kernel initializer, respectively. Then in Sec. \ref{sec:bid_process}, we discuss how to apply the pre-trained generator and initializer to the BID process.


\subsection{Kernel generator learning}\label{sec:kg_learning}
\begin{figure}[t]
    \begin{minipage}{0.58\textwidth}
        \vspace{-4mm}
        \centering
        \includegraphics[width=0.98\textwidth]{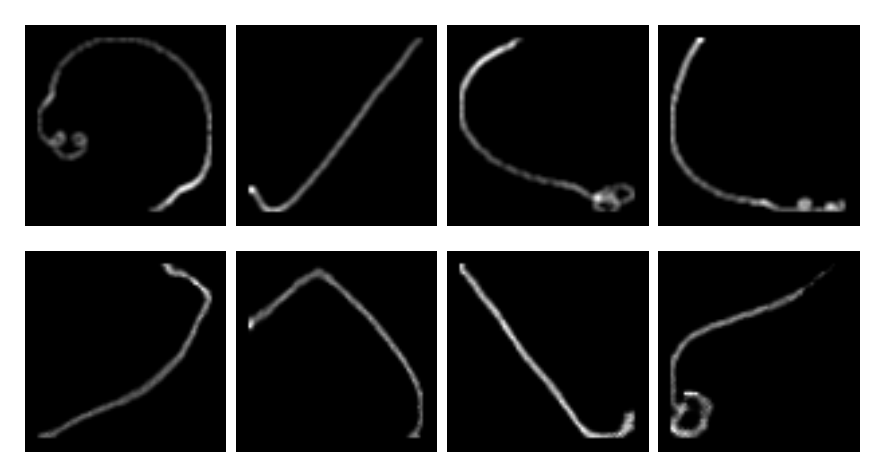}
    \end{minipage}
    \begin{minipage}{0.35\textwidth}
        \caption{Comparison of blur kernels synthesized according to physical model and generated by our pre-trained generator. Top row: blur kernels synthesized according to \cite{lai2016comparative}. Bottom row: blur kernels generated by the pre-trained kernel generator.}\label{fig_3.3}
    \end{minipage}
    \vspace{-10mm}
\end{figure}

To learn the kernel generator, a large amount of motion blur kernels are synthesized to simulate their distribution based on the physical generation mechanism proposed in \cite{kupyn2018deblurgan} or \cite{lai2016comparative}. Then we train the kernel generator $G_k(\cdot;\theta_k^{\ast})$ with these synthesized blur kernels. Note that our kernel generator is made up of several convolutional layers inspired by DCGAN \cite{radford2015unsupervised}. Though can be seen as a straightforward application of GAN, this step is crucial for our method. On the one hand, the learned generator depicts the statistical structures of blur kernels, serving as a kernel DGP in the BID process; on the other hand, the learned kernel generator also forms the milestone for learning the kernel initializer in a low-dimensional latent space, which will be discussed in Sec. \ref{sec:ki_learning}. Fig. \ref{fig_3.3} shows several blur kernels generated by $G_k(\cdot;\theta_k^{\ast})$ and the physical model according to \cite{lai2016comparative}. As can be observed, the blur kernels generated by our learned generator are morphologically very similar to that synthesized by the physical model, indicating its capability to characterize the underlying manifold of the blur kernels.

\vspace{0.5mm}
\noindent\emph{\textbf{Remark}}. It should be mentioned there are some studies also considering learning DGP for blur kernels while with various generative models. For example, Asim et al. \cite{asim2020blind} used the VAE for the BID task, and Liang et al. \cite{DIP-FKP} adopted the NF for the blind image super-resolution. In this work, we prefer GAN due to the following reasons. First, compared with the vanilla VAE that tends to generate blurry samples \cite{Bredell2023ExplicitlyMT}, GAN can provide sharper ones. Second, in contrast to NF, GAN imposes no constraints on the invertibility between the latent code and the generated sample, thereby facilitating easier learning of the kernel manifold in a lower-dimensional latent space. This aspect is particularly advantageous when considering motion blur kernels, whose sizes can be relatively large (up to $75\times75$ in the Lai dataset \cite{lai2016comparative}).


\subsection{Kernel initializer learning}\label{sec:ki_learning}
As discussed before, kernel initialization is important for the overall performance in the BID task since the optimization is highly non-convex. Therefore, we propose to initialize the blur kernel in the latent space using an encoder $E(\cdot;\theta_E^{\ast})$ that maps the blurry image $\y$ to the latent code $\z$ of the learned kernel DGP.

A direct way to train the encoder is to solve the following optimization:
\vspace{-2mm}
\begin{equation}\label{eq_encoder1}
    \theta_E^{\ast} = \arg\mathop{\min}_{\theta_E} \sum\nolimits_n \ell(G_k(E(\y_n;\theta_E);\theta_k^{\ast}),\k_n),
    \vspace{-3mm}
\end{equation}
where $\y_n$ denotes the $n$-th blurry image, $\k_n$ the corresponding blur kernel, $\theta_k^{\ast}$ is the parameters of the pre-trained kernel generator, and $\ell(\cdot,\cdot)$ is some loss function. This optimization can be seen as training a predictor that directly estimates the blur kernel from the blurry image, which is not easy due to the relatively complex structures of blur kernels. On the other hand, if we can find the latent code $\z_n$ of $\k_n$, and learn to predict $\z_n$, the task could be simpler, since the latent space is expected to be much more compact than the original kernel space. Therefore, we propose to use a collaborative learning strategy motivated by \cite{guan2020collaborative}, to guide the encoder training within the latent space. Specifically, we first formulate the following optimization to train the encoder:
\vspace{-2mm}
\begin{equation}\label{eq_encoder2}
    \begin{split}
        &\underset{\theta_E}{\min}\!\sum_n\!\Big\{\|G_k(E(\y_n;\!\theta_E);\!\theta_k^{\ast})\!-\!G_k(\z_n;\!\theta_k^{\ast})\|_1\!+\!\lambda\|E(\y_n;\!\theta_E)\!-\!\z_n\|_2^2\Big\},\\
        &s.t.~~\z_n=\arg\underset{\z}{\min}~\|G_k(\z;\theta_k^{\ast})-\k_n\|_1,
    \end{split}
\end{equation}
where $\lambda$ is a tuning parameter set to $0.1$ in this work. For convenience, we denote the objective functions of the outer and inner optimizations as $\mathcal{L}_E(\theta_E)$ and $\mathcal{L}_z^n(\z)$, respectively, in the following.



\begin{algorithm}[tb]
    \caption{\footnotesize Kernel initializer learning}
    \label{algo_ki}
    \begin{algorithmic}[1]
        \scriptsize
        \Require Pre-trained kernel generator $G_k(\cdot;\theta_k^{\ast})$, blurry image-kernel pairs $\{\y_n,\k_n\}$, step size $\epsilon$
        \State Initialize $\theta_E^{(0)}$
        \For {$t=1,\cdots,T$}
        \State For each $n$:
        \State \hspace{3mm} $\z^{(0)}=E\big(\y_n;\theta_E^{(t-1)}\big)$
        \State  \hspace{3mm} {\bf for} $s=1,\cdots,S$ {\bf do} $\z^{(s)}=\z^{(s-1)}-\epsilon\nabla_{\z}\mathcal{L}_z^{n}(\z)|_{\z=\z^{(s-1)}}$ {\bf end for}
        \State  \hspace{3mm} $\z_n^{(t)}=\z^{(S)}$
        \State $\theta^{(0)}=\theta_E^{(t-1)}$
        \State {\bf for} $l=1,\cdots,L$ {\bf do} $\theta^{(l)}=\theta^{(l-1)}-\epsilon\nabla_\theta\mathcal{L}_E(\theta)|_{\theta=\theta^{(l-1)}}$ {\bf end for}
        \State $\theta_E^{(t)}=\theta^{(L)}$
        \EndFor
        \Ensure Kernel initializer $E(\cdot;\theta_E^{\ast})$, where $\theta_E^{\ast}=\theta_E^{(T)}$
    \end{algorithmic}
    \normalsize
\end{algorithm}

The optimization of Eq.~\eqref{eq_encoder2} involves a sub-task of solving $\z_n$ that corresponds to a GAN-inversion problem~\cite{xia2022gan}. 
Following \cite{guan2020collaborative}, we alternatively optimize $\z_n$ and $\theta_E$ in each iteration. Specifically, $\z_n$ is firstly initialized as $E(\y_n;\theta_E)$, and then updated by several gradient descent steps with respect to the lower-level objective $\mathcal{L}_z^n(\z)$. Based on the updated $\z_n$, $\theta_E$ is subsequently optimized according to the upper-level objective $\mathcal{L}_E(\theta_E)$, also by gradient descent steps. The whole procedure is summarized in Algorithm \ref{algo_ki}.

\begin{figure}[t]
    \vspace{-2mm}
    \begin{minipage}{0.63\textwidth}
        \centering
        \vspace{-4mm}
        \includegraphics[width=0.9\textwidth]{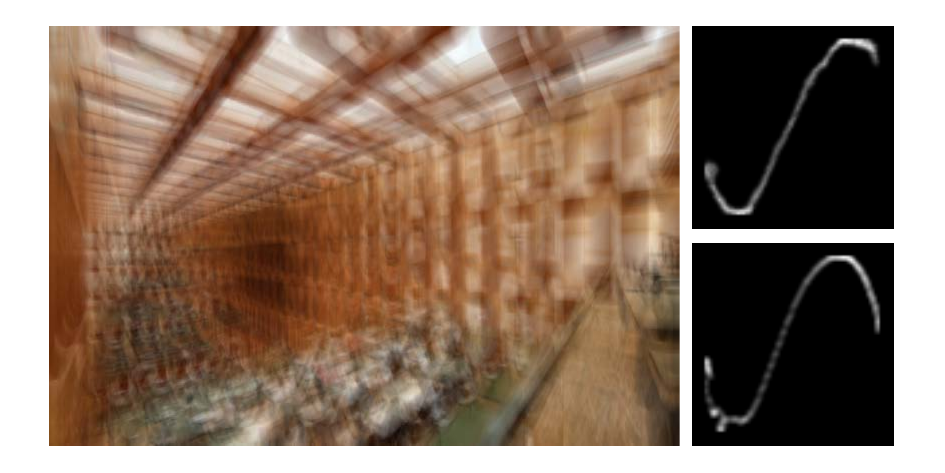}
    \end{minipage}
    \begin{minipage}{0.3\textwidth}
        \caption{Illustration of the estimating ability of the pre-trained kernel initializer. Left: the blurry image. Top-right: the estimated blur kernel by the kernel initializer. Bottom-right: the ground-truth blur kernel.}\label{fig_3.4}
    \end{minipage}
    \vspace{-10mm}
\end{figure}

Once trained, encoder $E(\cdot;\theta_E^{\ast})$ can be regarded as an effective kernel initializer, providing a promising prediction of the blur kernel from the blurry image. Fig. \ref{fig_3.4} shows an example of the predicted kernel by the learned kernel initializer, in comparison with the ground truth. It can be observed that, though not perfect, the prediction is indeed very close to the ground truth, and thus is expected to be a good starting point for further processing. 

\subsection{BID Process}\label{sec:bid_process}
With the pre-trained kernel generator $G_k(\cdot;\theta_k^{\ast})$ and initializer $E(\cdot;\theta_E^{\ast})$, we can formulate the BID problem as follows: 
\vspace{-1mm}
\begin{equation}\label{eq_bid_ours_naive}
    (\z_k^{\ast},\theta_x^{\ast})=\arg\!\underset{\z_k,\theta_x}{\min}\!\!\left\|G_k(\z_k;\theta_k^{\ast})\!\otimes\! G_x(\z_x;\theta_x)-\y\right\|^2,
    \vspace{-3mm}
\end{equation}
where $\z_x$ is a random vector, $\z_k$ is initialized by $\z_k^{(0)}=E(\y;\theta_E^{\ast})$. After optimization, the desired clean image and blur kernel can be accessed via $\hat{\x}=G_x(\z_x;\theta_x^{\ast})$ and $\hat{\k}=G_k(\z_k^{\ast};\theta_k^{\ast})$, respectively. In fact, such a naive implementation still cannot achieve satisfactory results as empirically illustrated in Sec.~\ref{sec:ab_study}, in particular when the blur kernel is large. This can be attributed to the relatively low dimensionality of $\z_k$, compared with the blur kernel, which enforces too strong constraints and makes the optimization more difficult. 

To address this issue, let's first recall one small yet effective trick in StyleGAN \cite{karras2019style}. It introduced a network to map the latent code $\z$ to a style vector $\w$, enabling more precise control of the generated style. Inspired by this trick, we propose to optimize the feature map of the first layer in the kernel generator $G_k(\cdot;\theta_k^{\ast})$, whose dimension is relatively higher, instead of the latent code $\z$. By denoting the first layer of $G_k(\cdot;\theta_k^{\ast})$ as $g_1(\cdot)$ and the reduced generator without $g_1(\cdot)$ as $G_k^{(w)}(\cdot)$, the BID problem of Eq.~\eqref{eq_bid_ours_naive} is reformulated as
\vspace{-3mm}
\begin{equation}\label{eq_bid_ours}
    (\w_k^{\ast},\theta_x^{\ast})=\arg\!\!\underset{\w_k,\theta_x}{\min}\!\!\left\|G_k^{(w)}(\w_k)\!\otimes\! G_x(\z_x;\theta_x)-\y\right\|^2,
    \vspace{-2mm}
\end{equation}
where $\w_k$ is initialized by $\w_k^{(0)}=g_1\left(E(\y;\theta_E^{\ast})\right)$. The whole BID process for solving Eq.~\eqref{eq_bid_ours} is summarized in Algorithm \ref{algo_bid}.
\begin{algorithm}[tb]
    \caption{\footnotesize BID process with pre-trained kernel generator and initializer}
    \label{algo_bid}
    \begin{algorithmic}[1]
        \scriptsize
        \Require Pre-trained kernel generator $G_k(\cdot;\theta_k^{\ast})=G_k^{(w)}(g_1(\cdot))$, pre-trained kernel initializer $E(\cdot;\theta_E^{\ast})$, blurry image $\y$, random vector $\z_x$, step size $\epsilon$
        \State Initialize $\w_k^{(0)}=g_1(E(\y;\theta_E^{\ast}))$, $\theta_x^{(0)}$, denote $\mathcal{L}(\w,\theta)=\left\|G_k^{(w)}(\w)\otimes G_x(\z_x;\theta)-\y\right\|^2$
        \For {$t=1,\cdots,T$}
        \State $\w_k^{(t)}=\w_k^{(t-1)}-\epsilon\nabla_{\w}\mathcal{L}(\w,\theta)|_{\w=\w_k^{(t-1)},\theta=\theta_x^{(t-1)}}$
        \State $\theta_x^{(t)}=\theta_x^{(t-1)}-\epsilon\nabla_{\theta}\mathcal{L}(\w,\theta)|_{\w=\w_k^{(t-1)},\theta=\theta_x^{(t-1)}}$
        \EndFor
        \State $\w_k^{\ast}=\w_k^{(T)}$, $\theta_x^{\ast}=\theta_x^{(T)}$
        \Ensure Estimated clean image $\hat{\x}=G_x(\z_x;\theta_x^{\ast})$, estimated blur kernel $\hat{\k}=G_k^{(w)}(\w_k^{\ast})$
    \end{algorithmic}
    \normalsize
\end{algorithm}
\setlength{\textfloatsep}{4mm}

It should be noted that the parameters of the generator network can also be fine-tuned in previous studies~\cite {pan2021exploiting}. However, due to the highly non-convex nature of the BID optimization, too many to-be-optimized parameters may hinder its performance. We thus only update $\w_k$ for the blur kernel in this work. Empirical results in Sec. \ref{sec:ab_study} substantiate the superiority of this strategy over updating the whole kernel generator.

\section{Experiments}\label{sec:experiments}

In this section, we present experimental results to demonstrate the effectiveness of the proposed method and ablation studies to analyze some key components in our framework. More results can be found in the appendix, together with discussions on the limitations of this work.

\subsection{Datasets and implementation details}

\noindent\textbf{Datasets}. We evaluate our proposed framework mainly on two datasets: one is synthesized by ourselves, and the other is the well-known BID benchmark constructed by Lai et al. \cite{lai2016comparative}. For the synthetic dataset, we randomly select 80 source images from MSCOCO \cite{MSCOCO} and apply the motion blur synthesized by \cite{kupyn2018deblurgan} on them. The second Lai dataset consists of 25 clean images and 4 large-size blur kernels, resulting in 100 testing blur images in total. Beyond the two synthetic datasets, we also test our method real blurry images collected by Lai et al. \cite{lai2016comparative}.

\vspace{0.5mm}
\noindent\textbf{Implementation details}. As discussed in Sec. \ref{sec:method}, our whole framework consists of a pre-training stage and a BID process. In the pre-training stage, we first train the kernel generator $G_k(\cdot;\theta_k^{\ast})$ with DCGAN \cite{radford2015unsupervised}, using kernels synthesized following the way in \cite{kupyn2018deblurgan} or \cite{lai2016comparative}. Then we freeze the weights of $G_k(\cdot;\theta_k^{\ast})$ and train the kernel initializer $E(\cdot;\theta_E^{\ast})$ that is a ResNet-18 \cite{he2016deep}. To train the kernel initializer, we randomly crop $256\times256$ patches from the sources in OpenImages \cite{kuznetsova2020open} and convolve them with blur kernels to get blurry images. The Adam optimizer \cite{kingma2014adam} with its default configuration in PyTorch \cite{paszke2019pytorch} is adopted. The initial learning rate is 1e-4 and reduced to 1e-5 as the loss becomes stable and finally to 1e-6. In the BID process, we jointly estimate the blur kernel $\k$ and the clean image $\x$ by optimizing $\w_k$ and $\theta_x$ according to Eq. \eqref{eq_bid_ours}. The initial learning rate of $\w_k$ and $\theta_x$ are set as 5e-4 and 1e-2, respectively, and decayed following \cite{ren2020neural}. For the architecture of the DIP network $G_x(\cdot;\theta_x)$, we follow the settings in SelfDeblur \cite{ren2020neural}.

\subsection{Experimental results}
\noindent{\bf Results on our synthetic dataset}. We first verify the effectiveness of the proposed method on the synthetic dataset by ourselves. 
Nine comparison methods are considered, including two traditional model-based ones (Pan et al. \cite{pan2016blind}, Dong et al. \cite{dong2017blind}), four supervised deep learning ones (Tao et al. \cite{tao2018scale}, Kupyn et al. \cite{kupyn2019deblurgan}, Kaufman and Fattal \cite{kaufman2020deblurring}, Zamir et al. \cite{zamir2021multi}), and three DIP-based ones (Ren et al. \cite{ren2020neural}, Huo et al. \cite{huo2023blind}, Li et al. \cite{li2023self}). We follow the default settings in their papers for these methods or tune them by ourselves for the best performance.

\begin{table}[t]
    \centering
    \centering
    \caption{Quantitative comparisons of various methods on our synthetic dataset.}
    \label{tab_syn}
    \vspace{-3mm}
    \scriptsize
        \begin{tabular}{@{}C{4.0cm}@{}|@{}C{1.3cm}@{}|@{}C{1.3cm}@{}|
                        @{}C{2.8cm}@{}|@{}C{1.3cm}@{}|@{}C{1.3cm}@{}}
            \Xhline{0.8pt}
            Method &  PSNR & SSIM & Method &  PSNR & SSIM\\
            \hline
            Pan et al. \cite{pan2016blind} & 24.09 & 0.808 & Zamir et al. \cite{zamir2021multi} & 22.27 & 0.652\\
            Dong et al. \cite{dong2017blind} & 23.85 & 0.799 & Ren et al. \cite{ren2020neural} & 25.48 & 0.765 \\
            Tao et al. \cite{tao2018scale} & 23.21 & 0.699 & Huo et al. \cite{huo2023blind} & 28.13 & 0.898\\
            Kupyn et al. \cite{kupyn2019deblurgan} & 23.46 & 0.716 & 	Li et al. \cite{li2023self} & 24.49 & 0.744\\
            Kaufman and Fattal \cite{kaufman2020deblurring} & 25.42 & 0.814 & Ours & \bf{29.67} & \bf{0.928}\\				
         \Xhline{0.8pt}
        \end{tabular}
        \vspace{-2mm}
\end{table}
\begin{figure}[t]
    \centering
    \subfloat[Blurred]
    {\includegraphics[width=0.16\textwidth]{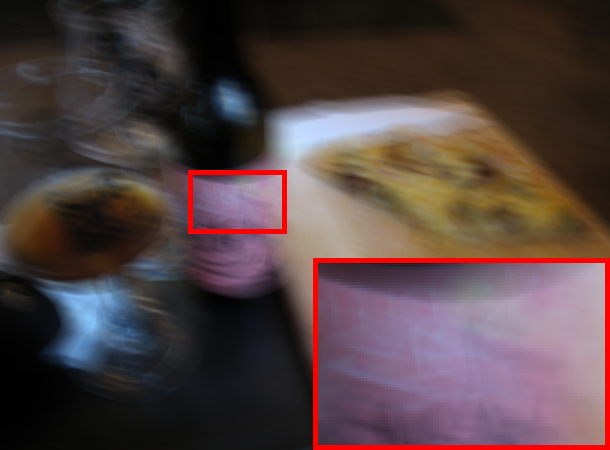}}\
    \subfloat[Pan \cite{pan2016blind}]
    {\includegraphics[width=0.16\textwidth]{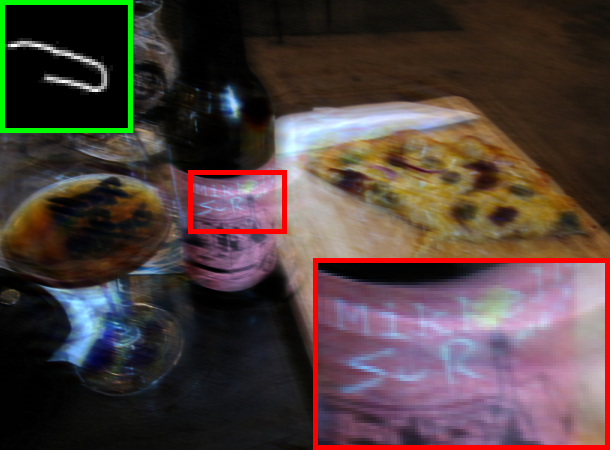}}\
    \subfloat[Dong \cite{dong2017blind}]
    {\includegraphics[width=0.16\textwidth]{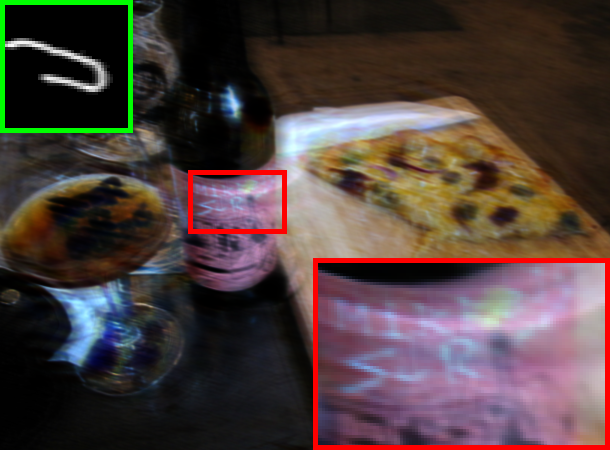}}\
    \subfloat[Tao \cite{tao2018scale}]
    {\includegraphics[width=0.16\textwidth]{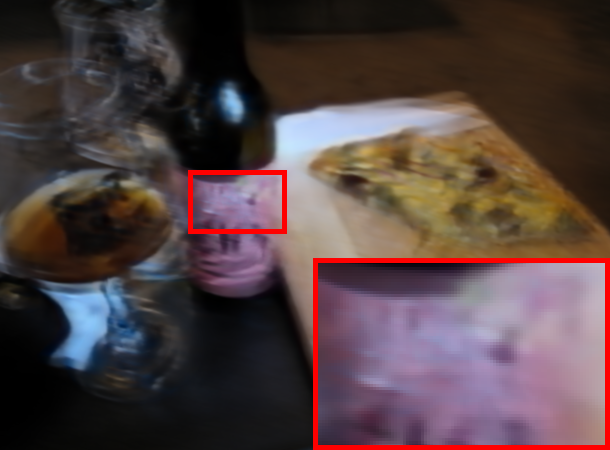}}\
    \subfloat[Kupyn \cite{kupyn2019deblurgan}]
    {\includegraphics[width=0.16\textwidth]{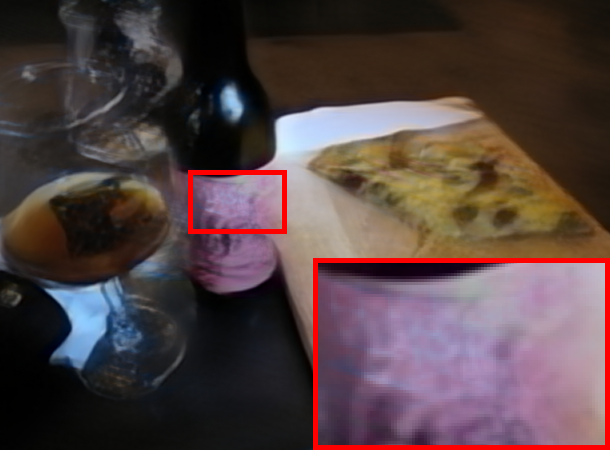}}\
    \subfloat[Kaufman \cite{kaufman2020deblurring}]
    {\includegraphics[width=0.16\textwidth]{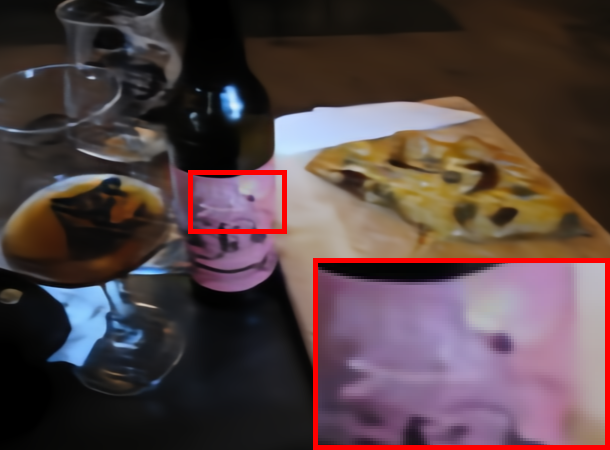}}\\
    \subfloat[Zamir \cite{zamir2022restormer}]
    {\includegraphics[width=0.16\textwidth]{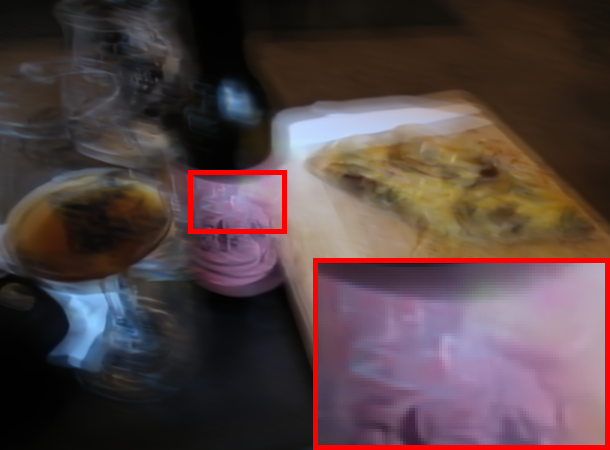}}\
    \subfloat[Ren \cite{ren2020neural}]
    {\includegraphics[width=0.16\textwidth]{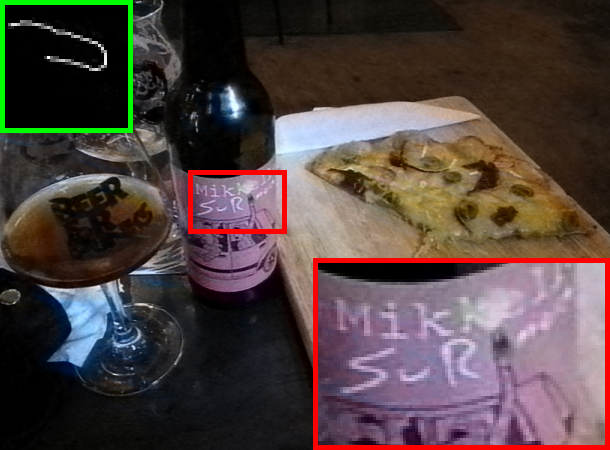}}\
    \subfloat[Huo \cite{huo2023blind}]
    {\includegraphics[width=0.16\textwidth]{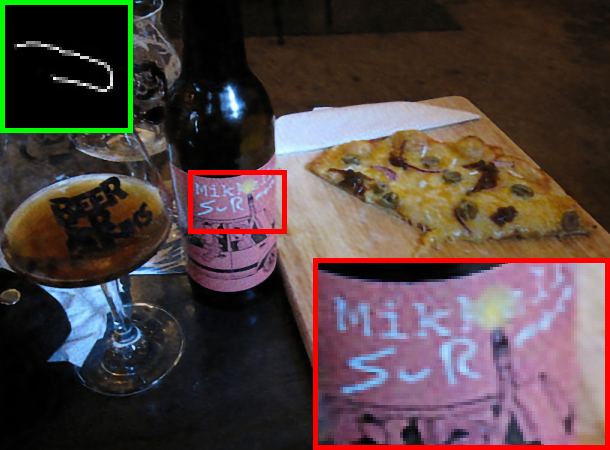}}\
    \subfloat[Li \cite{li2023self}]
    {\includegraphics[width=0.16\textwidth]{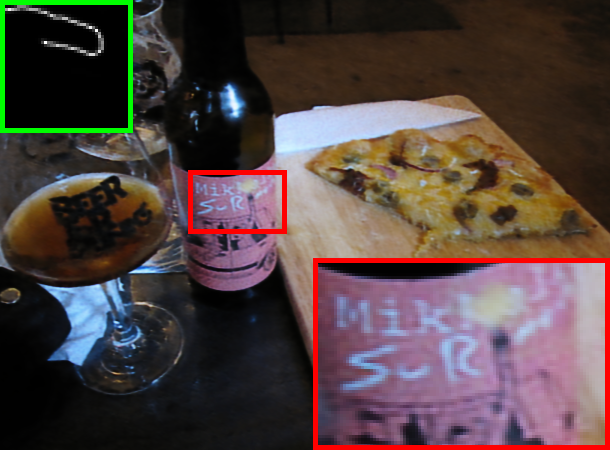}}\
    \subfloat[Ours]
    {\includegraphics[width=0.16\textwidth]{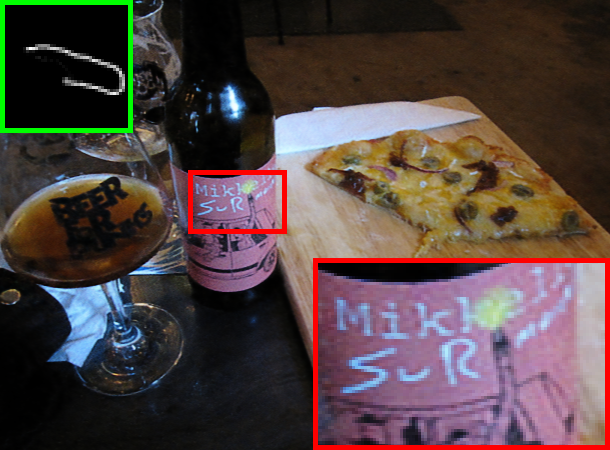}}\
    \subfloat[Ground truth]
    {\includegraphics[width=0.16\textwidth]{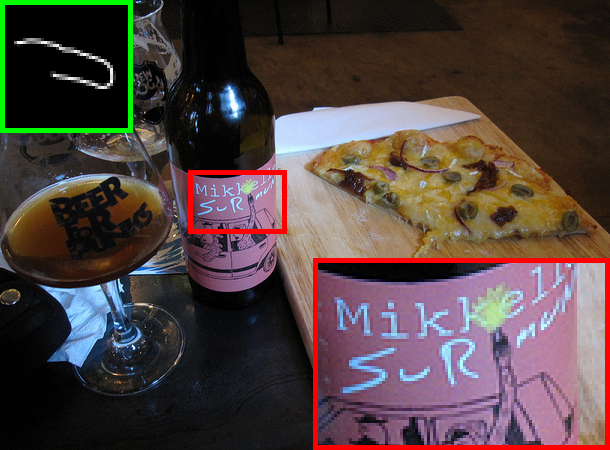}}
    \vspace{-3mm}
    \caption{Visual results on our synthetic dataset. The estimated blur kernel is placed on the top-left corner for each method if available.}\label{fig_syn}
    \vspace{-3mm}
\end{figure}

The quantitative results of all competing methods are summarized in Table \ref{tab_syn}. As can be seen from this table, the proposed method achieves the best performance in terms of both PSNR and SSIM, demonstrating its effectiveness. The superiority of our method can be visually observed from Fig.~\ref{fig_syn}. Specifically, the traditional model-based methods are able to obtain accurate estimations for the blur kernel, but the BID results are not very satisfactory due to the insufficiency of the handcrafted image priors. Most of the supervised deep learning methods generate over-smoothed outcomes, mainly attributed to the lack of exploitation of the physical blur model. Even though achieving relatively better results, the existing DIP-based methods still have some limitations. For example, Ren et al.'s method performs unsatisfactorily, and the structural details in the images by Huo et al. and Li et al.'s methods are not sharp enough. In contrast to these methods, our method can not only accurately estimate the blur kernel, but also recover more image details with fewer distortions.
\begin{table}[t]
    \caption{Quantitative results of various methods on the dataset by Lai et al. \cite{lai2016comparative}.}
    \label{tab_lai}
    \vspace{-7mm}
    \tiny
    \begin{center}
            \begin{tabular}{@{}C{2.8cm}@{}|
                            @{}C{1.5cm}@{}|@{}C{1.5cm}@{}|@{}C{1.5cm}@{}|
                            @{}C{1.5cm}@{}|@{}C{1.5cm}@{}|@{}C{1.5cm}@{}}
                \Xhline{0.8pt}
                Method & Manmade & Natural & People & Saturated & Text & Average\\
                \hline
                Cho and Lee \cite{cho2009fast} & 16.35/0.389 & 20.14/0.520 & 19.90/0.556 & 14.05/0.493 & 14.87/0.443 & 17.06/0.480 \\
                Krishnan et al. \cite{krishnan2011blind} & 15.73/0.408 & 19.44/0.526 & 21.50/0.647 & 14.09/0.507 & 15.40/0.485 & 17.23/0.514 \\
                Xu et al. \cite{xu2013unnatural} & 17.99/0.599 & 21.58/0.679 & 24,40/0.813 & 14.53/0.538 & 17.64/0.668 & 19.23/0.659 \\
                Perrone and Favaro \cite{perrone2014total} & 17.41/0.551 & 21.04/0.676 & 22.77/0.735 & 14.24/0.511 & 16.94/0.593 & 18.48/0.613 \\
                Pan et al. \cite{pan2016blind} & 18.25/0.561 & 22.15/0.672 & 22.72/0.710 & 16.01/0.591 & 16.69/0.530 & 19.16/0.613 \\
                Dong et al. \cite{dong2017blind} & 17.30/0.432 & 21.18/0.586 & 21.89/0.652 & 15.85/0.554 & 16.18/0.480 & 18.48/0.541  \\
                Tao et al. \cite{tao2018scale} & 16.77/0.342 & 19.90/0.459 & 21.41/0.616 &15.12/0.506 & 15.41/0.437 &17.72/0.472 \\
                Kupyn et al. \cite{kupyn2019deblurgan} & 17.05/0.367 &20.37/0.489& 21.70/0.642& 15.20/0.522& 15.86/0.476 &18.04/0.499 \\
                Kaufman and Fattal \cite{kaufman2020deblurring} & 19.36/0.650 & 23.14/0.756 & 26.78/0.873 & 16.40/0.654 & 17.92/0.683 & 20.72/0.723 \\
                Zamir et al. \cite{zamir2021multi} & 16.73/0.341 & 19.80/0.447 & 20.88/0.587 & 15.14/0.505 & 14.49/0.383 & 17.41/0.453 \\
                Ren et al. \cite{ren2020neural} & 19.00/0.617 &22.94/0.715 &26.43/0.781 &18.38/0.688 &22.82/0.776 &21.91/0.715 \\
                Huo et al. \cite{huo2023blind} & 22.51/0.820& 26.20/0.901 &31.38/0.953 &18.14/0.691 &27.72/0.933 &25.19/0.860 \\
                Li et al. \cite{li2023self} & 20.81/0.721 & 24.07/0.781 & 28.45/0.860 & 14.81/0.527 & 23.10/0.834 & 22.25/0.744\\
                Ours & \textbf{24.67/0.942} & \textbf{28.12/0.969} & \textbf{32.86/0.973} & \textbf{18.98/0.737} & \textbf{28.91/0.951} & \textbf{26.71/0.914}\\
                \Xhline{0.8pt}
            \end{tabular}
    \end{center}
    \vspace{-4mm}
\end{table}
\begin{figure}[t]
    \vspace{-4mm}
    \centering
    \subfloat[\tiny Blurred]
    {\includegraphics[width=0.115\textwidth]{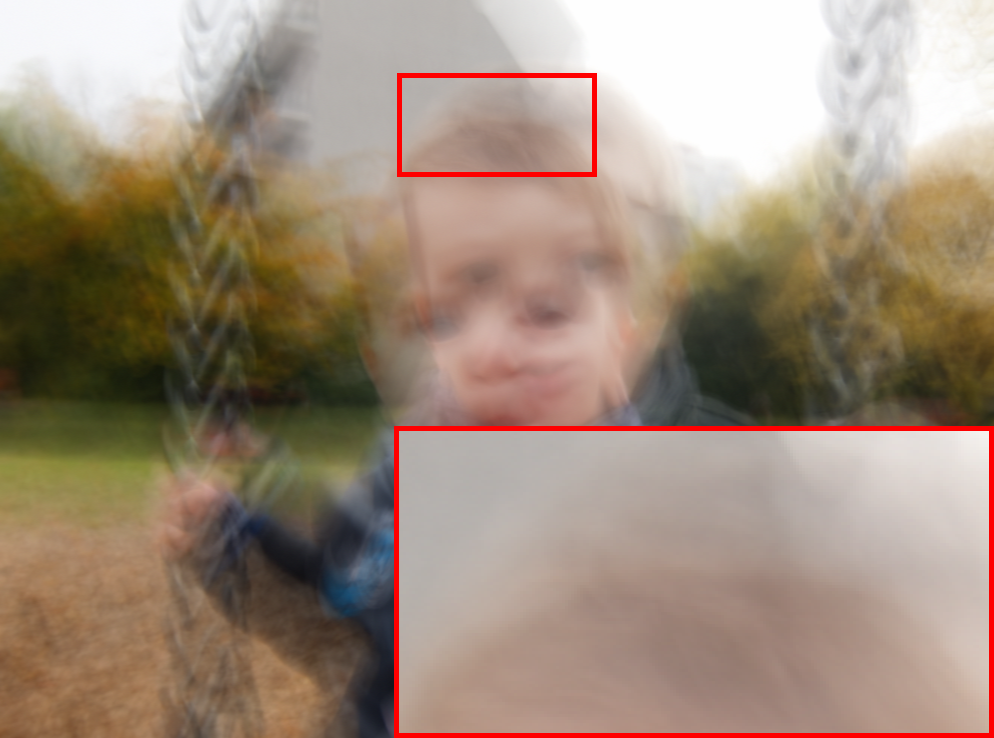}}\
    \subfloat[\tiny Cho \cite{cho2009fast}]
    {\includegraphics[width=0.115\textwidth]{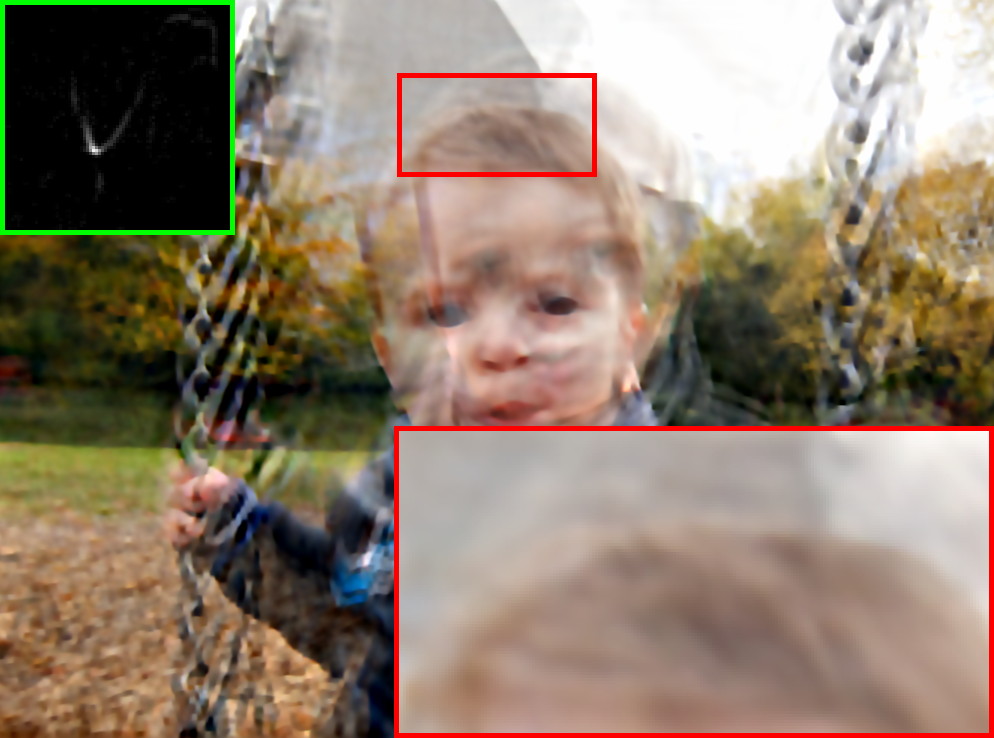}}\
    \subfloat[\tiny Krishnan \cite{krishnan2011blind}]
    {\includegraphics[width=0.115\textwidth]{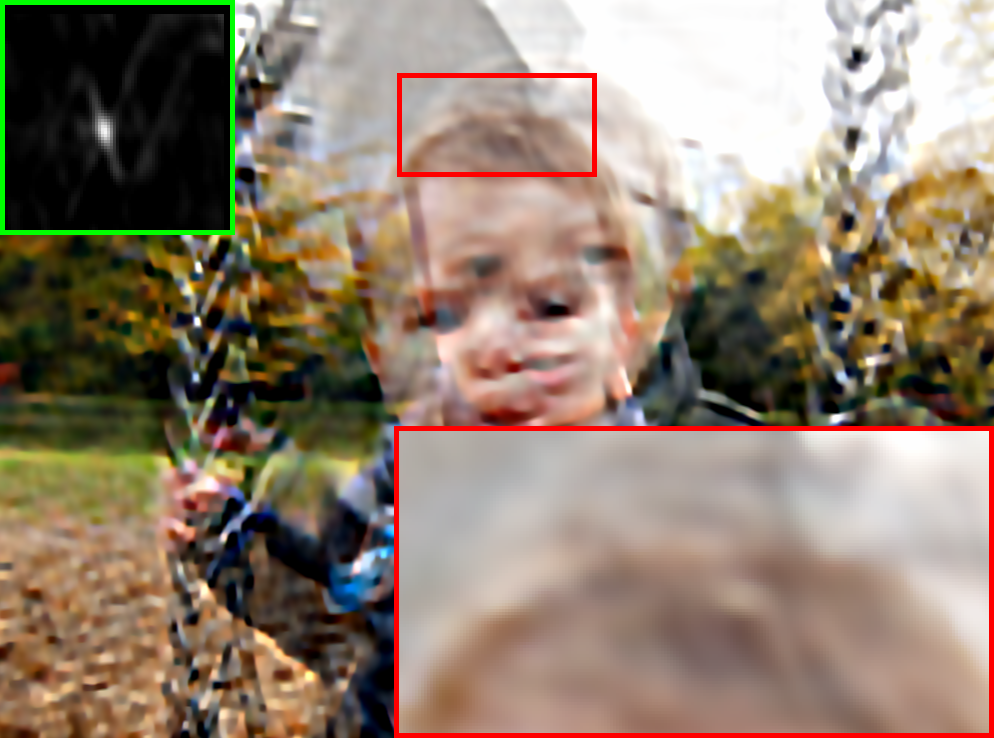}}\
    \subfloat[\tiny Xu \cite{xu2013unnatural}]
    {\includegraphics[width=0.115\textwidth]{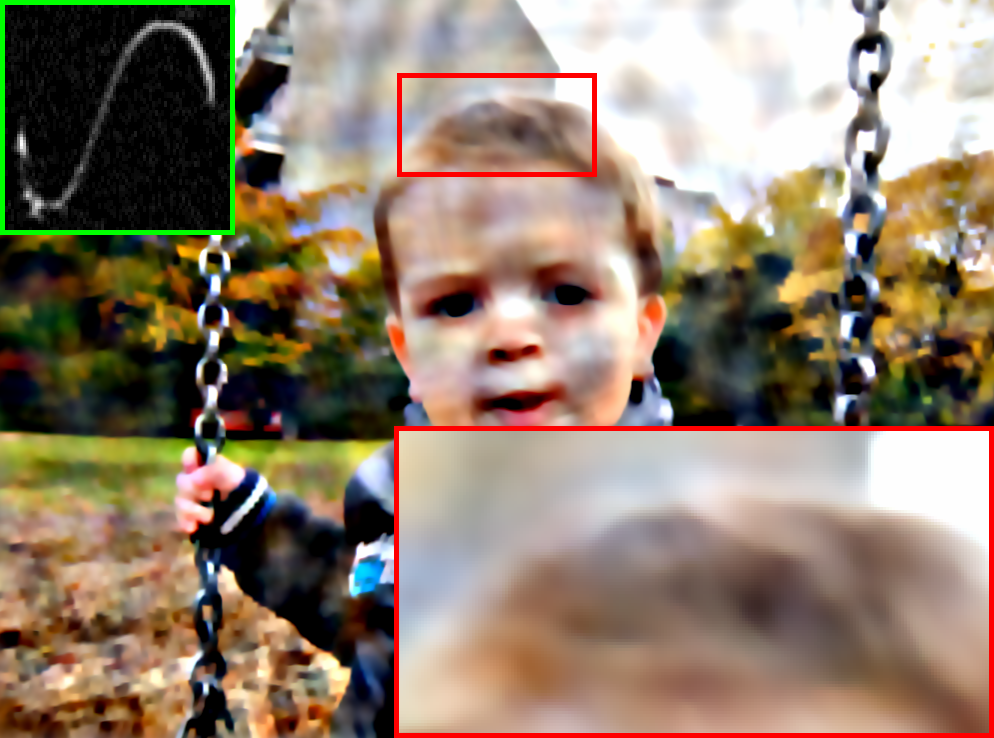}}\
    \subfloat[\tiny Perrone \cite{perrone2014total}]
    {\includegraphics[width=0.115\textwidth]{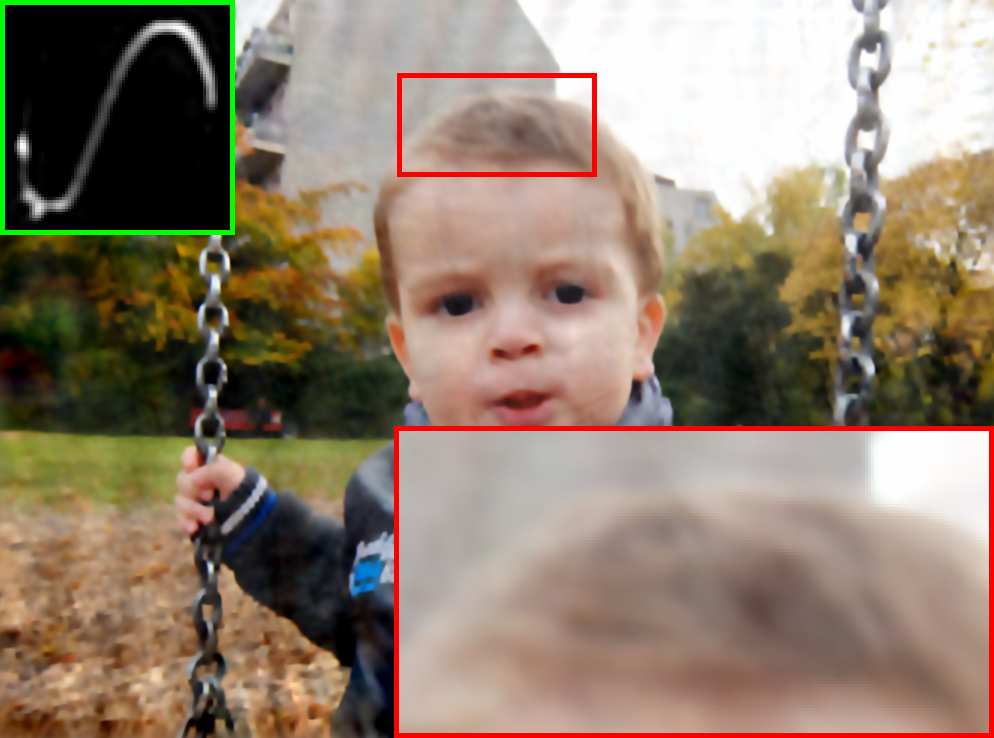}}\
    \subfloat[\tiny Pan \cite{pan2016blind}]
    {\includegraphics[width=0.115\textwidth]{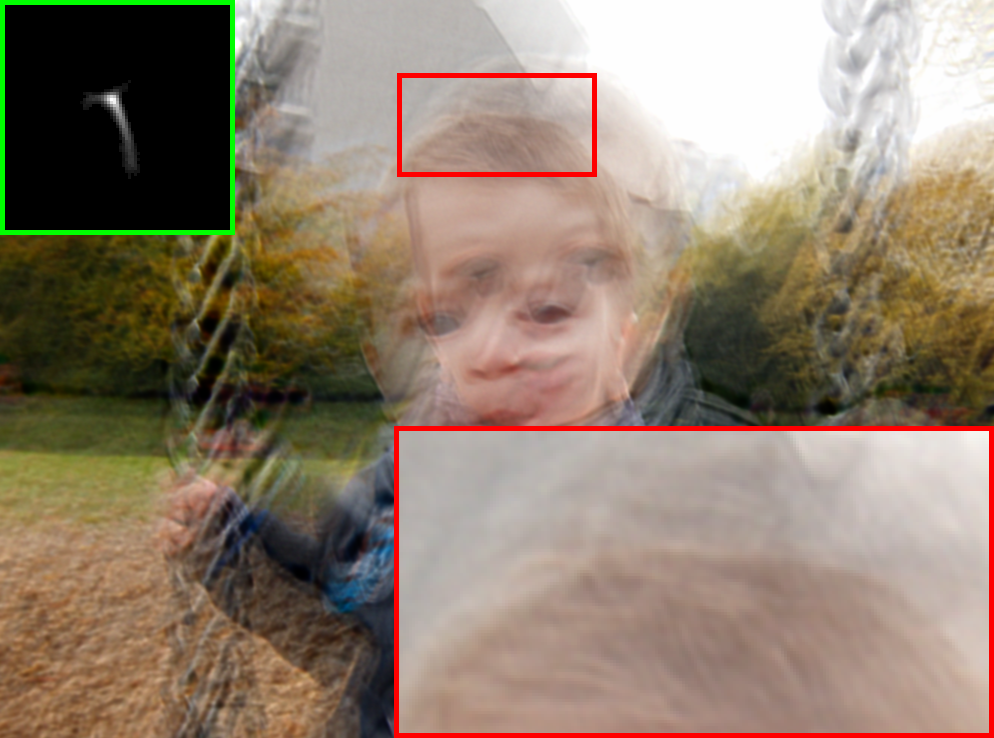}}\
    \subfloat[\tiny Dong \cite{dong2017blind}]
    {\includegraphics[width=0.115\textwidth]{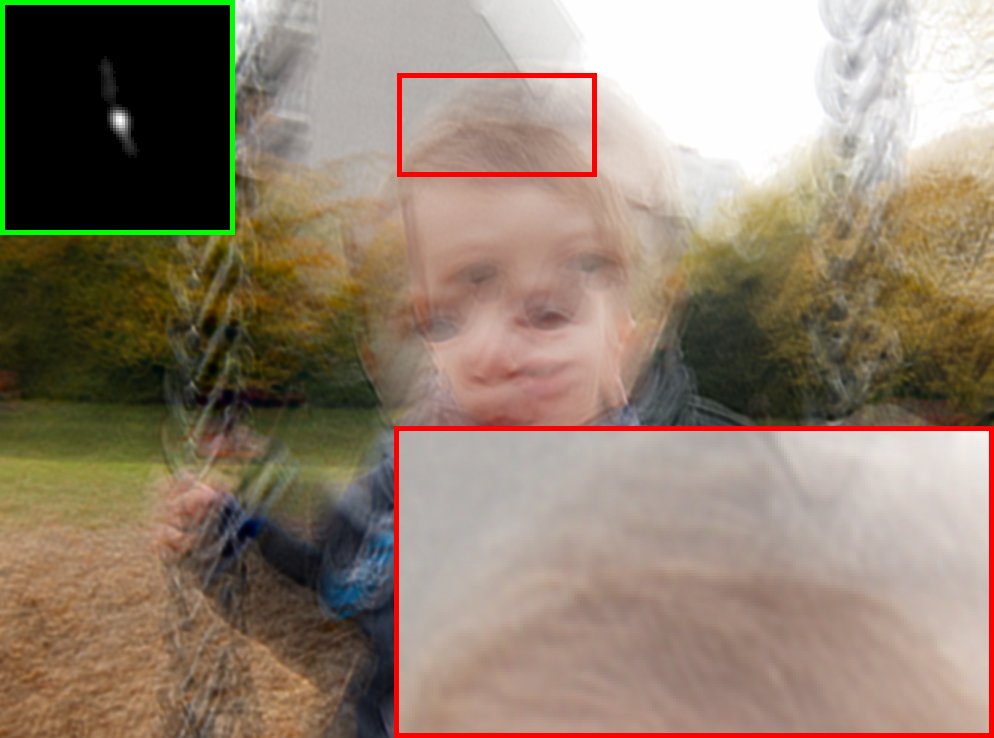}}\
    \subfloat[\tiny Tao \cite{tao2018scale}]
    {\includegraphics[width=0.115\textwidth]{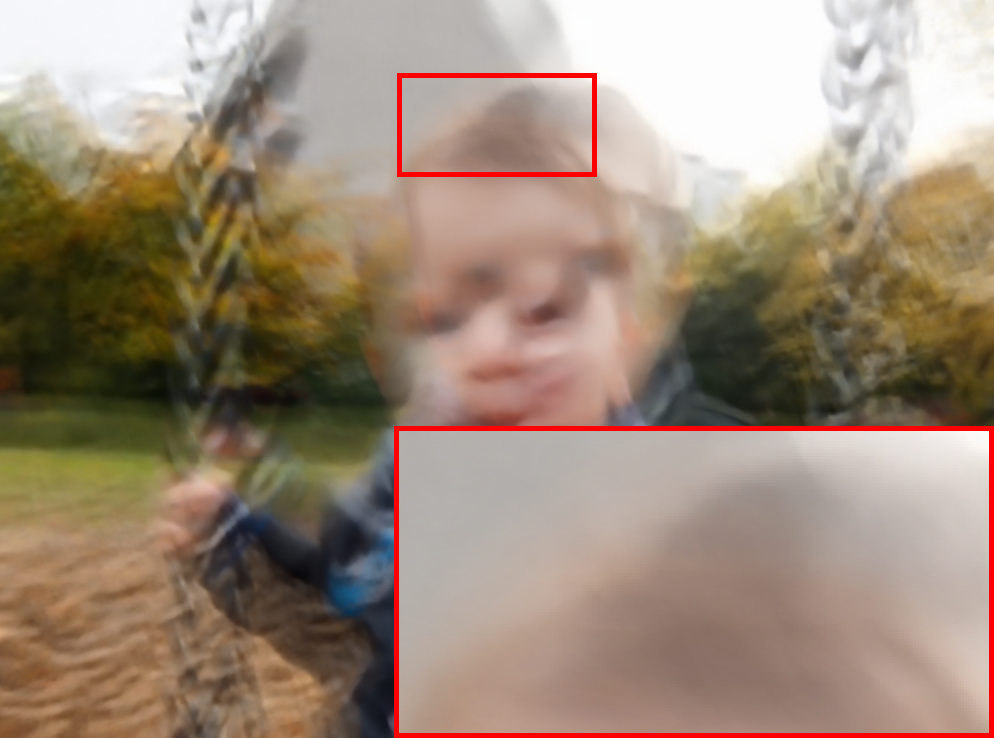}}\\
    \subfloat[\tiny Kupyn \cite{kupyn2019deblurgan}]
    {\includegraphics[width=0.115\textwidth]{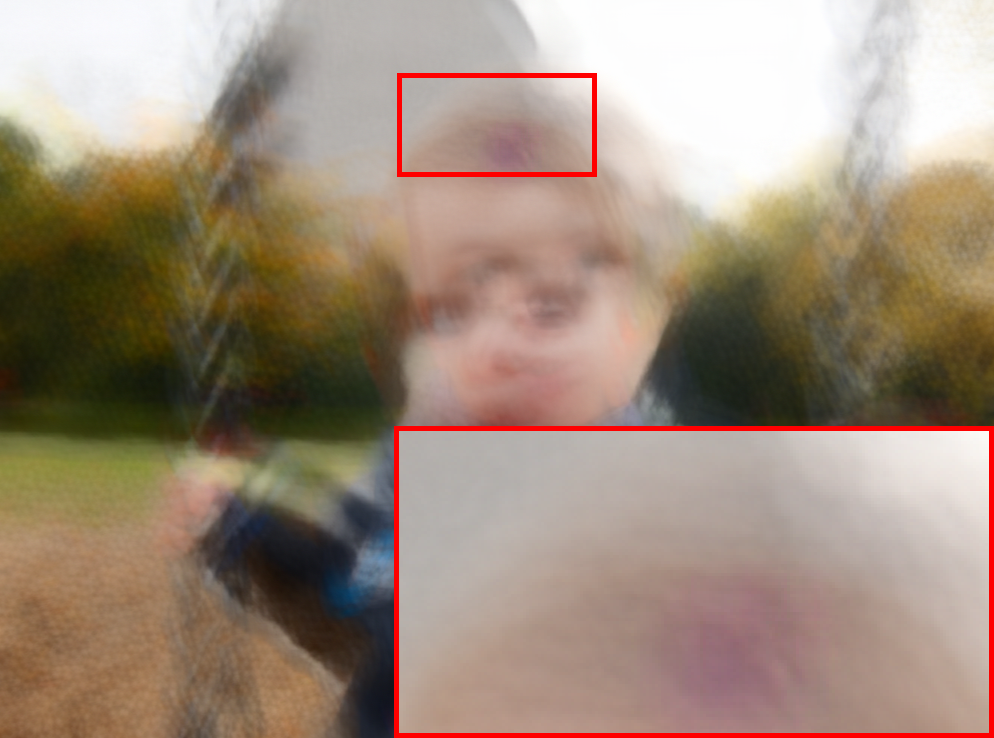}}\
    \subfloat[\tiny Kaufman \cite{kaufman2020deblurring}]
    {\includegraphics[width=0.115\textwidth]{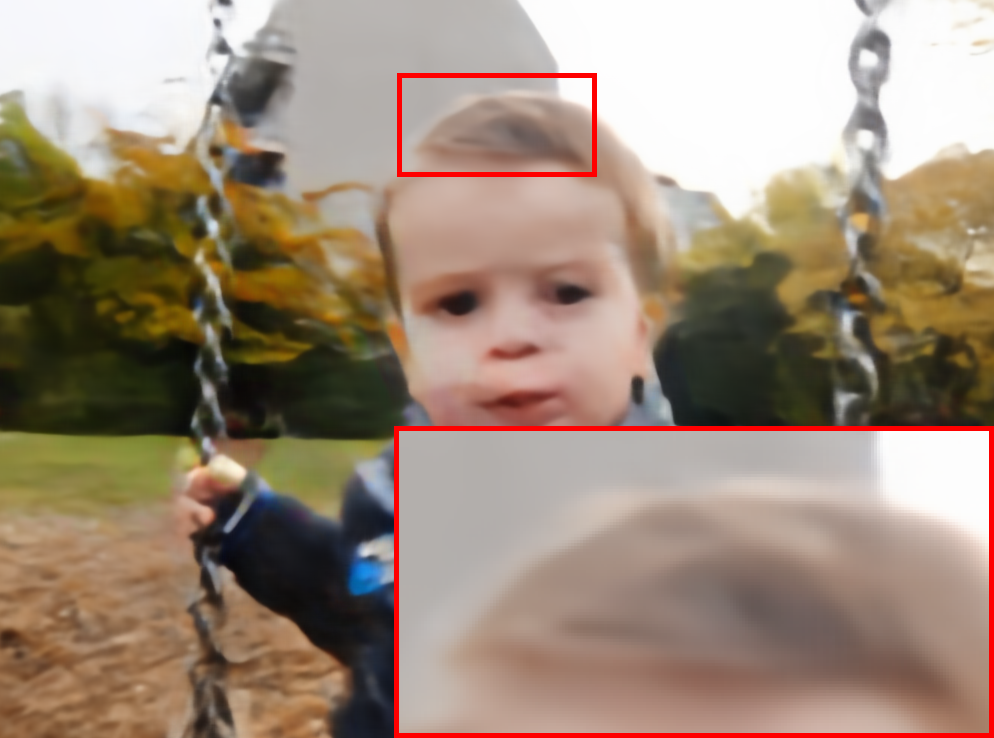}}\
    \subfloat[\tiny Zamir \cite{zamir2022restormer}]
    {\includegraphics[width=0.115\textwidth]{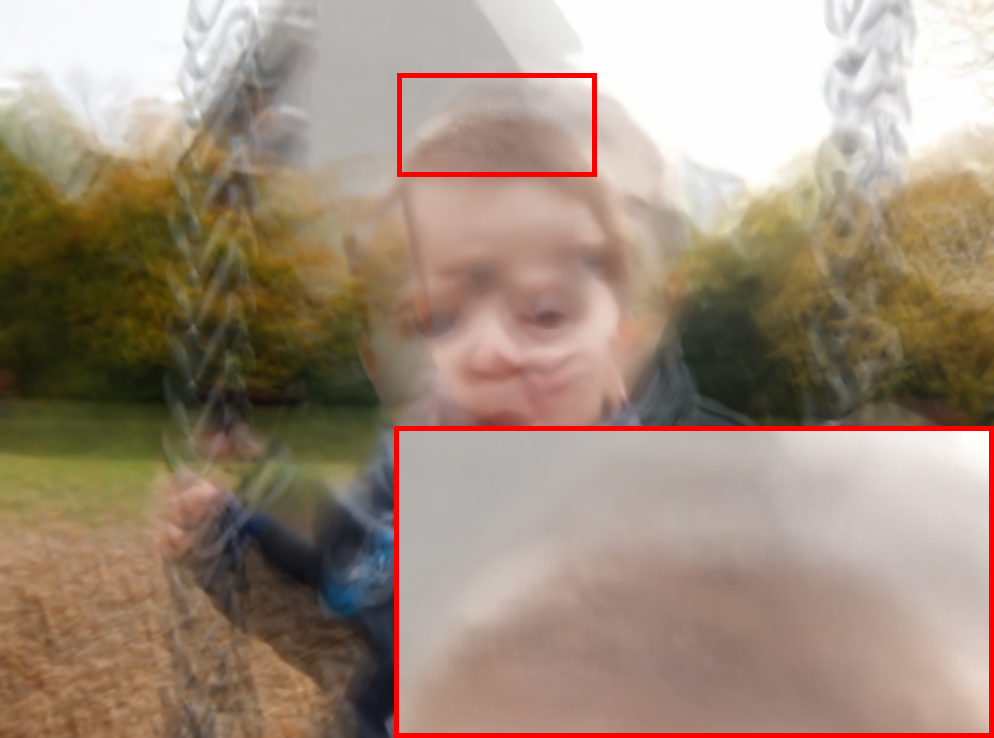}}\
    \subfloat[\tiny Ren \cite{ren2020neural}]
    {\includegraphics[width=0.115\textwidth]{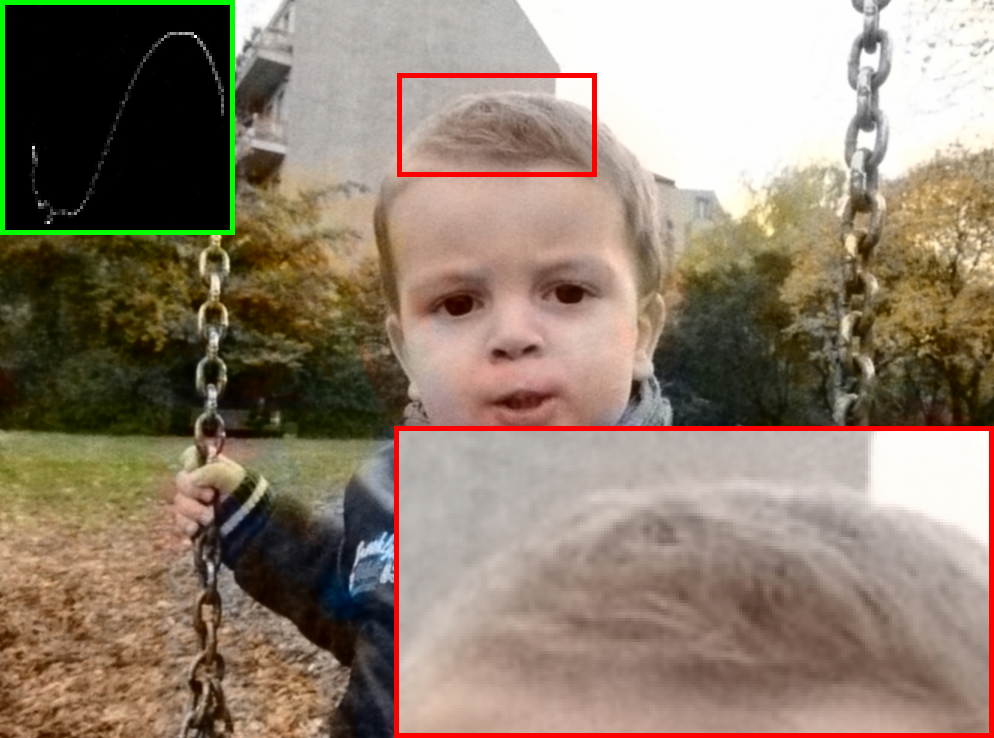}}\
    \subfloat[\tiny Huo \cite{huo2023blind}]
    {\includegraphics[width=0.115\textwidth]{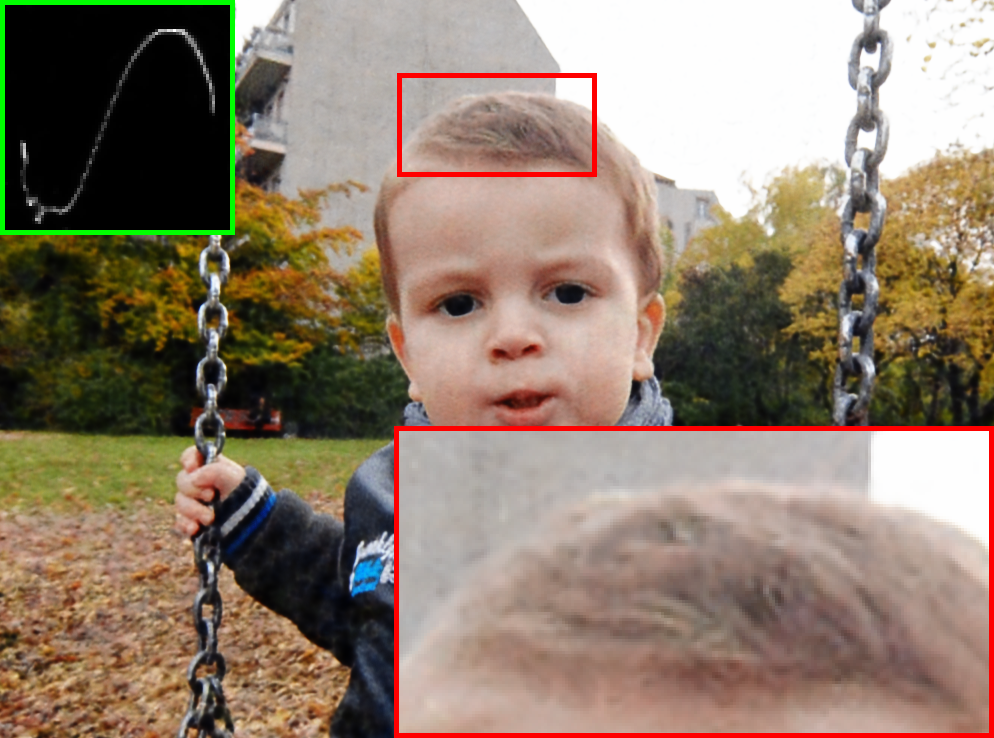}}\
    \subfloat[\tiny Li \cite{li2023self}]
    {\includegraphics[width=0.115\textwidth]{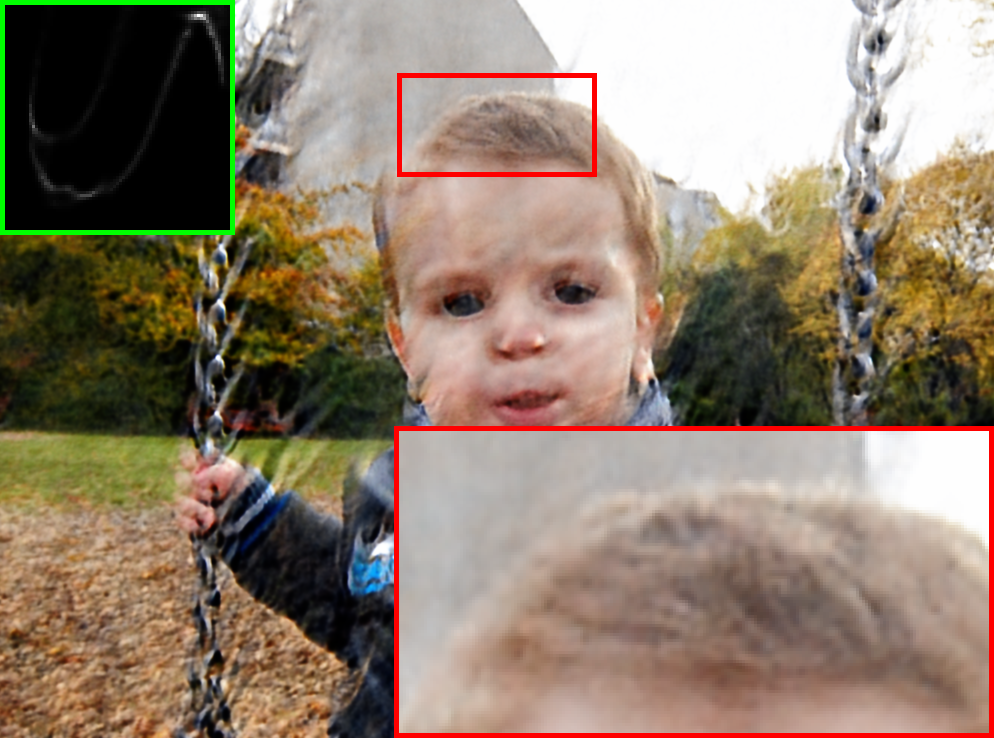}}\
    \subfloat[\tiny Ours]
    {\includegraphics[width=0.115\textwidth]{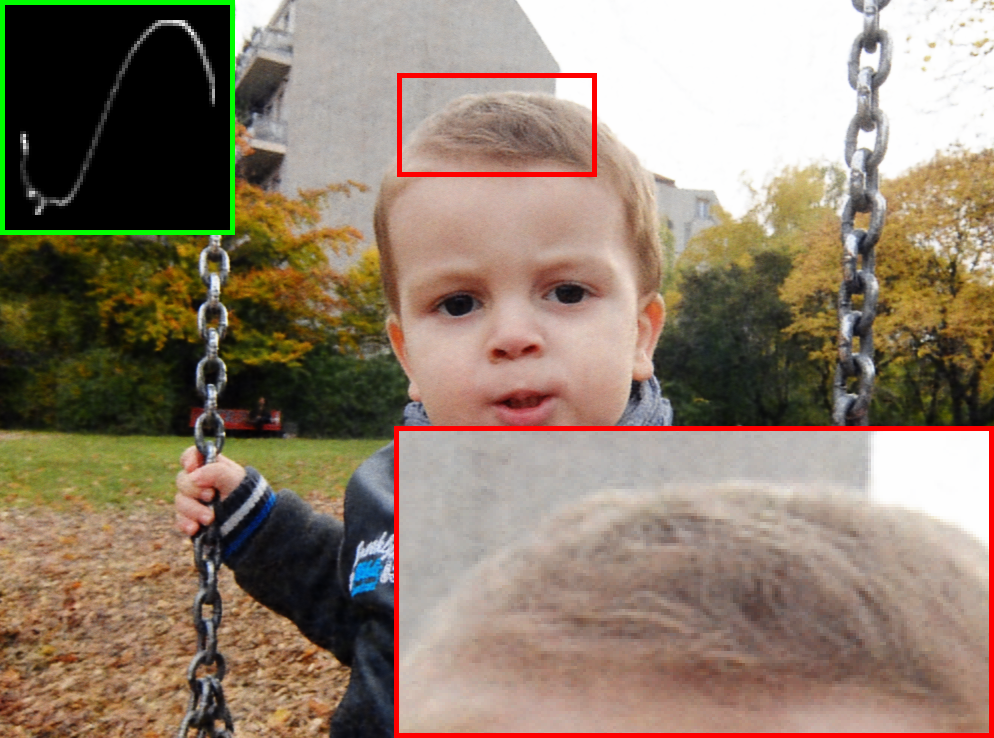}}\
    \subfloat[\tiny Ground truth]
    {\includegraphics[width=0.115\textwidth]{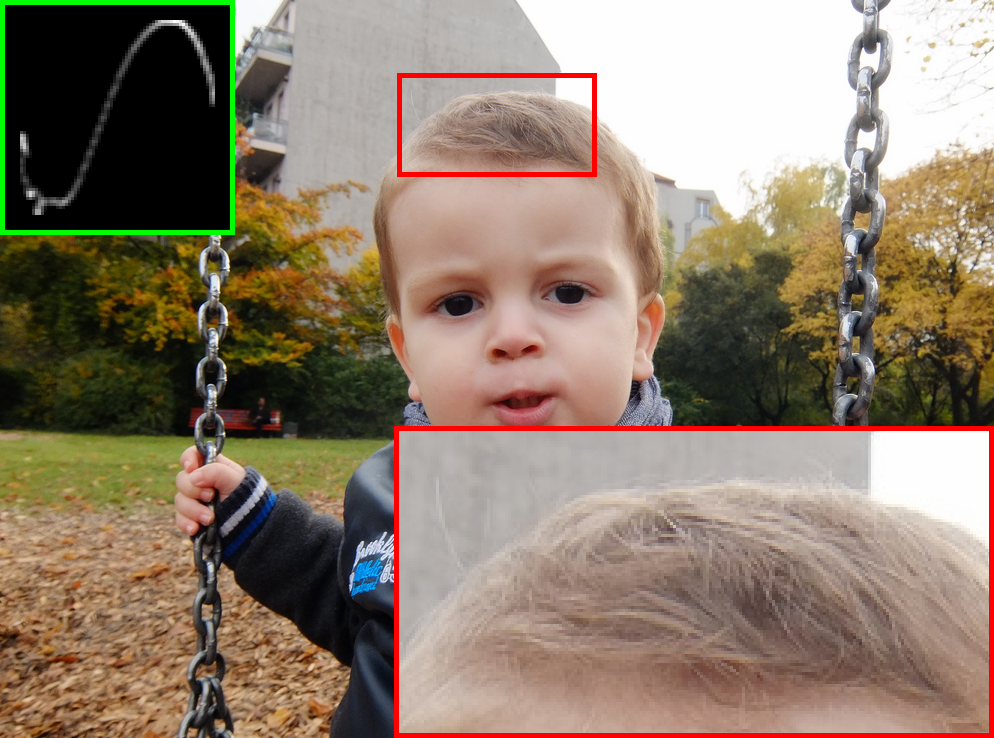}}
    \vspace{-2mm}
    \caption{Visual results on the dataset by Lai et al. \cite{lai2016comparative}. The estimated blur kernel is placed on the top-left corner of each method if available.}\label{fig_lai}
    \vspace{-3mm}
\end{figure}

\vspace{0.5mm}
\noindent{\bf Results on the dataset by Lai et al. \cite{lai2016comparative}}. We evaluate our proposed method on the challenging BID benchmark by Lai et al. \cite{lai2016comparative}. This dataset consists of 100 blurry images synthesized by applying 4 blur kernels, with sizes ranging from $31\times31$ to $75\times75$, to 25 clean images. These 25 clean images are categorized into 5 groups, namely \emph{Manmade}, \emph{Natural}, \emph{People}, \emph{Saturated} and \emph{Text}. 
We compare our method with 13 existing methods, including six traditional model-based methods (Cho and Li \cite{cho2009fast}, Krishnan et al. \cite{krishnan2011blind}, Xu et al. \cite{xu2013unnatural}, Perrone and Favaro \cite{perrone2014total}, Pan et al \cite{pan2016blind}, Dong et al. \cite{dong2017blind}), four supervised deep learning methods (Tao et al.\cite{tao2018scale}, Kupyn et al. \cite{kupyn2019deblurgan}, Kaufman and Fattal \cite{kaufman2020deblurring}, Zamir et al. \cite{zamir2021multi}), and three DIP-based methods (Ren et al. \cite{ren2020neural}, Huo et al. \cite{huo2023blind}, Li et al. \cite{li2023self}).

Table \ref{tab_lai} reports the quantitative results of all competing methods, in terms of PSNR and SSIM. We can see that our method significantly outperforms existing methods in all of the five categories. Notably, our method is the only one that achieves an SSIM value higher than 0.9 on average. The visual results on one typical example are shown in Fig.~\ref{fig_lai}. It can be observed that most of the traditional model-based methods fail to estimate the blur kernel due to the relatively larger kernel size in this dataset. The supervised deep learning methods still produce distorted results as before, because they do not carefully consider the physical blur model. In comparison, DIP-based methods, including ours, can accurately estimate the blur kernel and further achieve promising deblurring results. Among them, our method recovers more details, such as hairs, with reference to the ground truth image, showing its superiority.

\begin{figure}[t]
	\centering
	\subfloat[Blurred]
	{\includegraphics[width=0.14\textwidth]{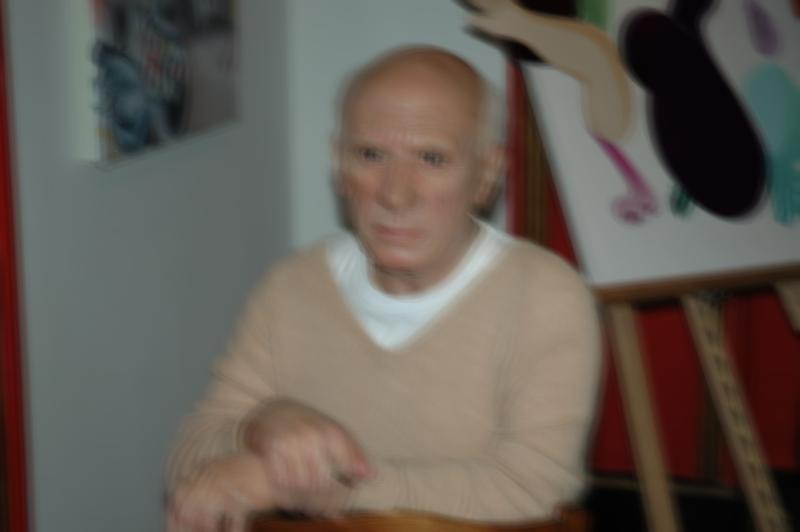}}\
	\subfloat[Pan \cite{pan2016blind}]
	{\includegraphics[width=0.14\textwidth]{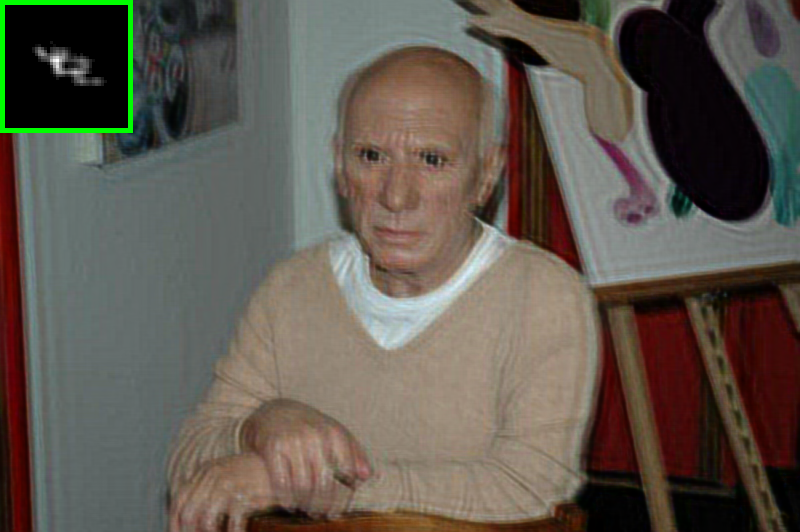}}\
	\subfloat[Dong \cite{dong2017blind}]
	{\includegraphics[width=0.14\textwidth]{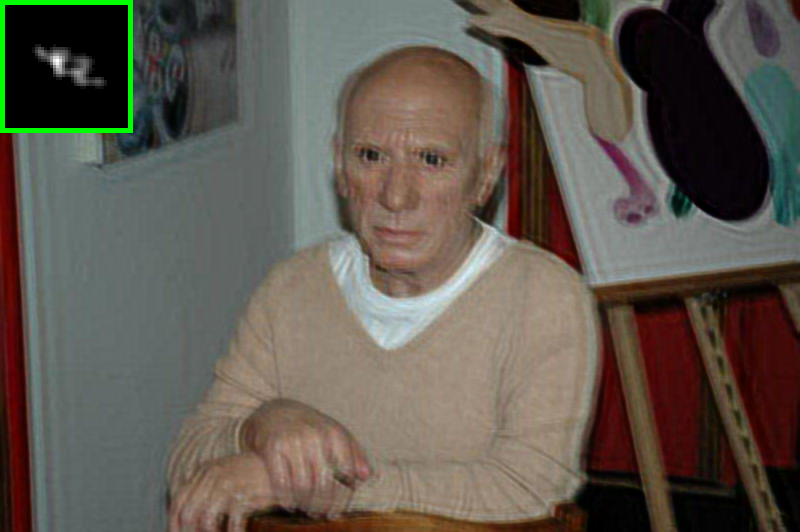}}\
	\subfloat[Tao \cite{tao2018scale}]
	{\includegraphics[width=0.14\textwidth]{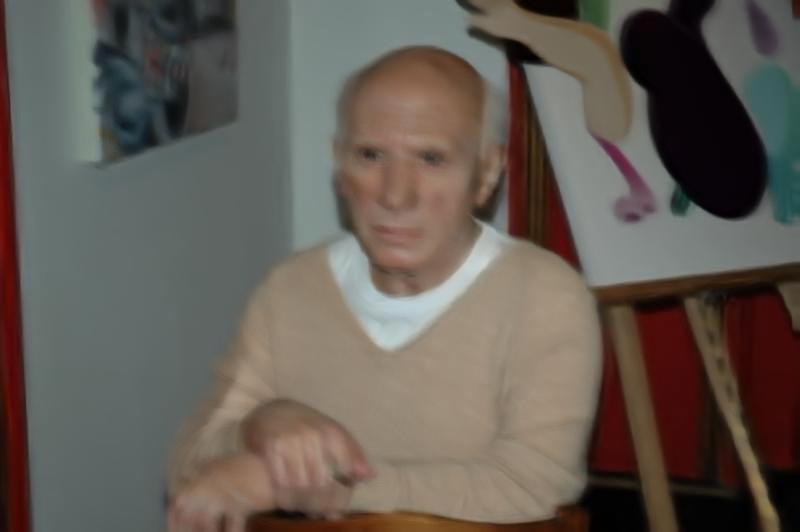}}\
	\subfloat[Kupyn \cite{kupyn2019deblurgan}]
	{\includegraphics[width=0.14\textwidth]{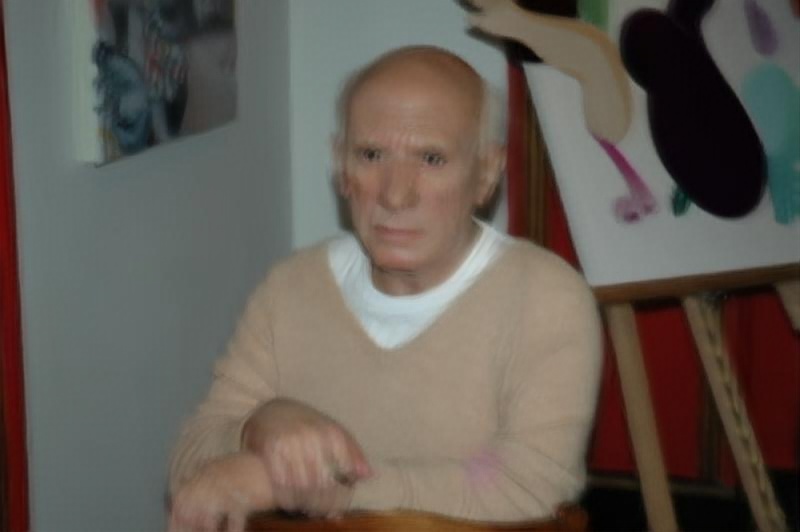}}\
	\subfloat[Kaufman \cite{kaufman2020deblurring}]
	{\includegraphics[width=0.14\textwidth]{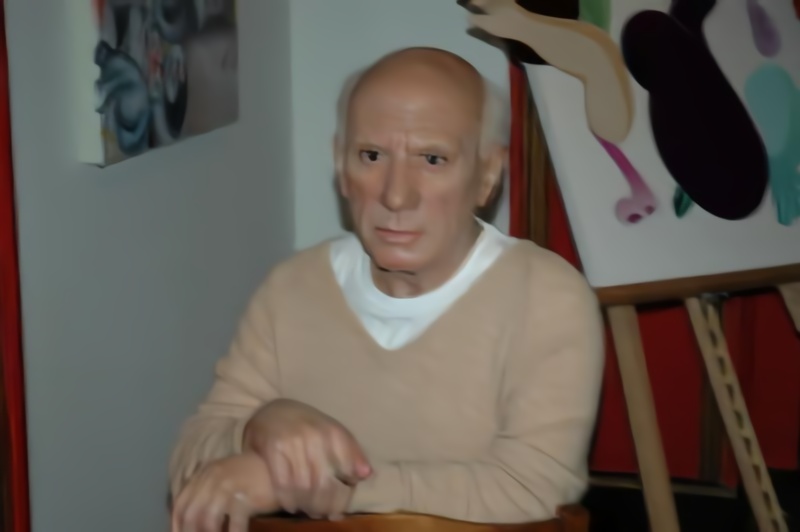}}\\
	\subfloat[Zamir \cite{zamir2022restormer}]
	{\includegraphics[width=0.14\textwidth]{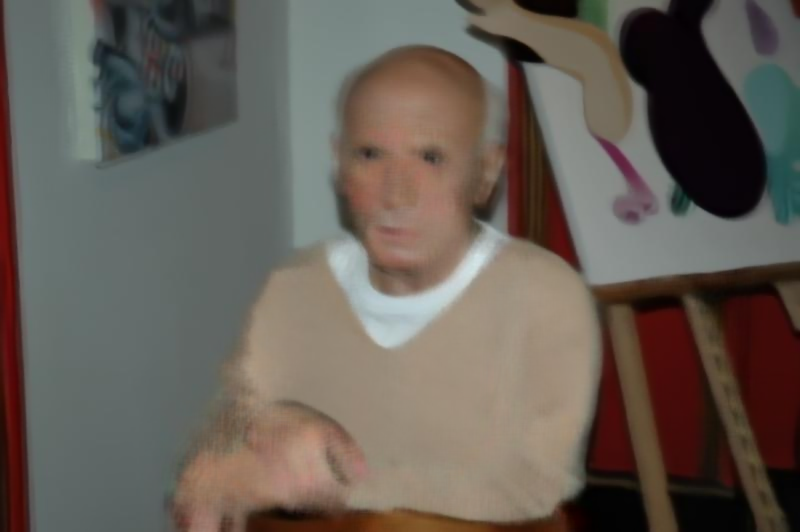}}\
	\subfloat[Ren \cite{ren2020neural}]
	{\includegraphics[width=0.14\textwidth]{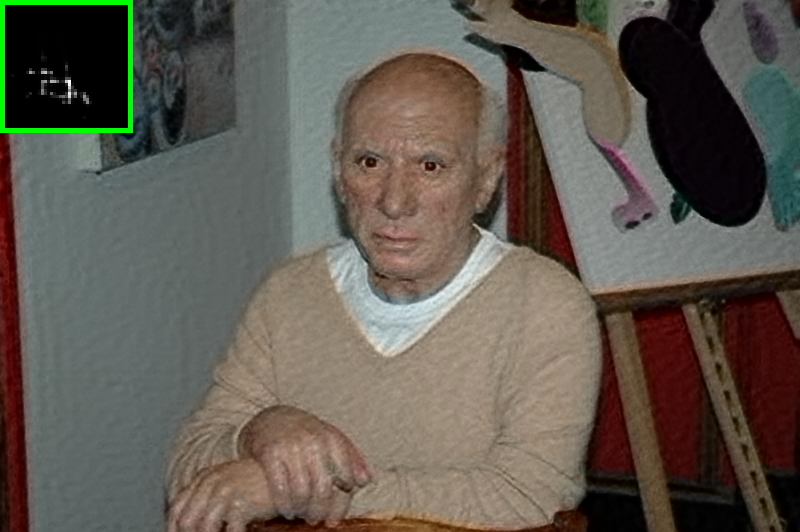}}\
	\subfloat[Huo \cite{huo2023blind}]
	{\includegraphics[width=0.14\textwidth]{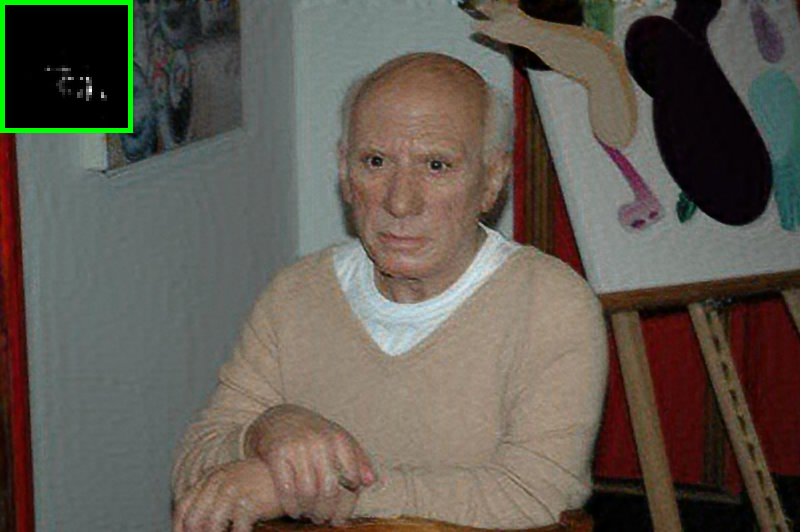}}\
	\subfloat[Li \cite{li2023self}]
	{\includegraphics[width=0.14\textwidth]{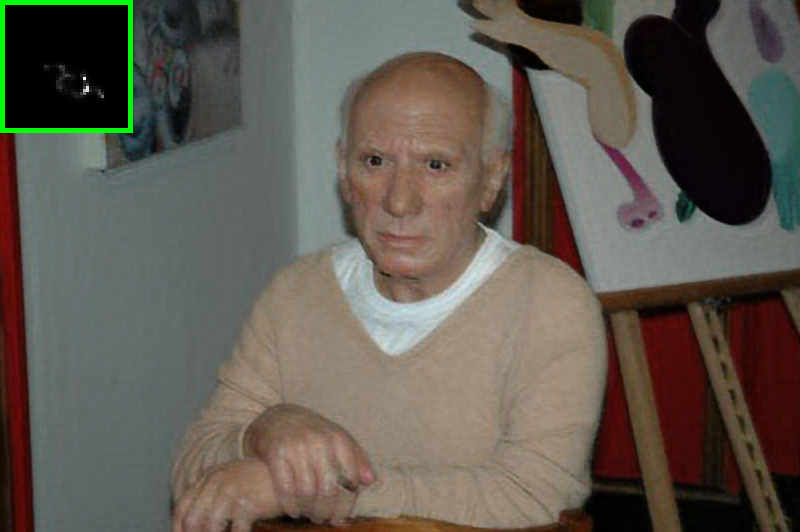}}\
	\subfloat[Ours]
	{\includegraphics[width=0.14\textwidth]{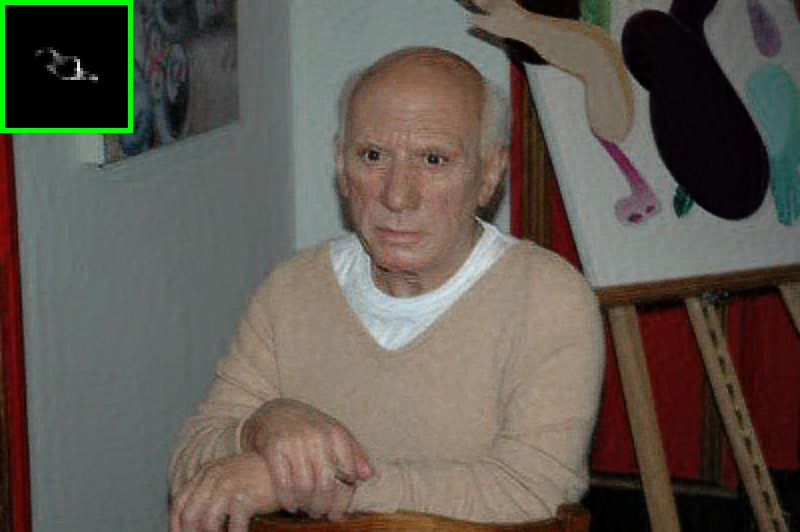}}
	\vspace{-3mm}
	\caption{Example results on the real blurry image provided by Lai et al. \cite{lai2016comparative}. The estimated blur kernel is placed on the top-left corner of each method if available.}
	\label{fig_lai_real}
\end{figure}

\vspace{0.5mm}
\noindent{\bf Example results on real blurry images}. We further apply comparison BID methods to real blurry images collected by Lai et al. \cite{lai2016comparative}, and show one example in Fig. \ref{fig_lai_real}. As can be seen, on this very challenging image, none of the methods generate perfect results, while our method provides a relatively promising one. 

\subsection{Ablation study}\label{sec:ab_study}
As shown in the previous experiments, the proposed method has demonstrated its effectiveness on various blurry images. In this section, we provide further analyses for better understanding its advantageous properties.

\begin{figure}[t]
    \vspace{-2mm}
    \begin{minipage}{0.5\textwidth}
        \centering
        \includegraphics[width=\columnwidth]{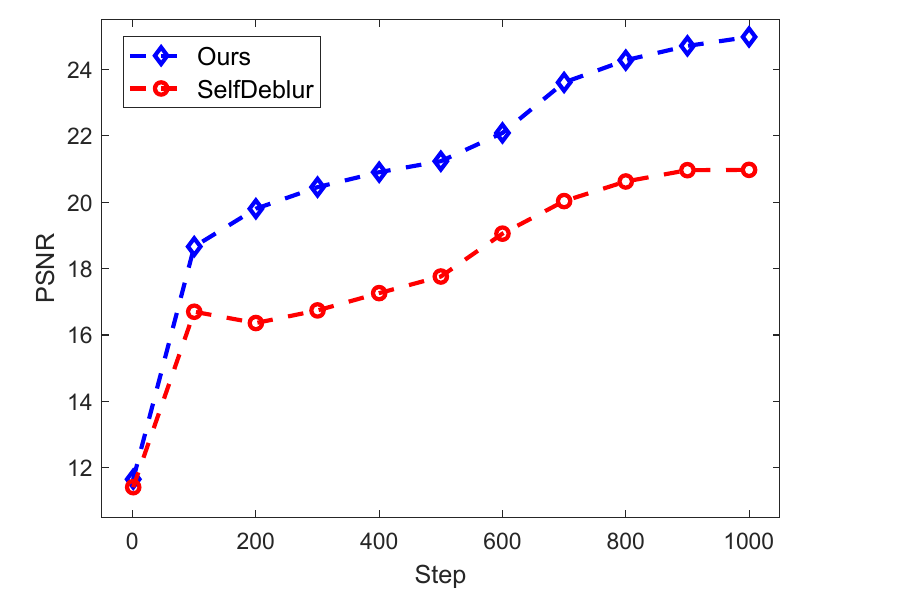}
    \end{minipage}
    \begin{minipage}{0.43\textwidth}
        \centering
        \vspace{-16mm}
        \caption{Performance comparison (PSNR) of the DIP-based BID method (SelfDeblur \cite{ren2020neural}) and our method in different iteration steps on the dataset by Lai et al. \cite{lai2016comparative}.}\label{fig_lai_step}
    \end{minipage}
    \vspace{-4mm}
\end{figure}

\vspace{0.5mm}
\noindent{\bf Convergence speed}. One major methodological difference between our method and existing DIP-based methods is that we can initialize the blur kernel with pre-trained models instead of random sampling. This property can not only result in more stable and better performance but also improve the convergence speed. To verify this point, we compare the PSNR values of our methods with that of Ren et al.'s \cite{ren2020neural} across various iterations on the Lai dataset in Fig.~\ref{fig_lai_step}. It can be seen that the performance of our method in the 400-th iteration is comparable with that of Ren et al.'s method in the 1000-th iteration. This faster convergence can be mostly attributed to the better initialization and characterization of the blur kernel, since both of them adopt the same DIP network for the image.

\vspace{0.5mm}
\noindent{\bf Optimization strategy for blur kernel}. As discussed in Sec. \ref{sec:bid_process}, during the BID process, we optimize the higher-dimensional feature map $\w_k$ instead of the original latent code $\z_k$ for the blur kernel. 
Now we empirically show the different behaviors of the two optimization strategies. As a reference, the results by optimizing the entire kernel generator are also reported, which has a larger search space for optimization. The detailed comparison results on the Lai dataset are summarized in Table \ref{tab_lai_kernel}. We can observe that optimizing the whole generator is not competitive due to the too-large search space. When the blur kernel is relatively small, the other two strategies have similar performance, while optimizing $\w_k$ is slightly better. As the blur kernel becomes larger, the performance by optimizing $\z_k$ significantly drops, while optimizing $\w_k$ still produces promising results. These observations demonstrate the reasonability of the proposed optimization strategy and also emphasize the importance of a proper search space. 

\vspace{0.5mm}
\noindent{\bf Effect of kernel initialization}. In our BID framework, a pre-trained encoder is employed to initialize the blur kernel. To show the necessity of such an initializer, we compare the proposed initialization strategy to two baselines. The first is the random initialization that randomly samples a latent code $\z$ for the blur kernel. The second is to initialize the kernel with all elements being the same, which can be seen as an ``average'' kernel, and then find the corresponding $\z_0$ by GAN-inversion. It should be noted that both of these two initialization strategies are independent of the blurry image. Table~\ref{tab_lai_init} lists the comparison results on the Lai dataset. We can see that the random initialization strategy produces the worst result since the randomly initialized kernel might largely deviate from the target without any constraint. The fixed ``average'' kernel performs better, and indeed already outperforms many other methods in Table~\ref{tab_lai}. That's because this strategy, in some sense, can be regarded as an ``average'' approximation to any blur kernel. Not surprisingly, the proposed initialization with a pre-trained encoder obtains the best performance as the initialized kernel has been very close to the ground-truth one, as shown in Fig.~\ref{fig_3.4}. These results substantiate the significant benefits brought by the pre-trained kernel initializer for the BID task.  

\begin{table}[t]
    \caption{Performance comparison (PSNR/SSIM) of different kernel optimization strategy on the dataset by Lai et al. \cite{lai2016comparative}.}
    \label{tab_lai_kernel}
    \vspace{-9mm}
    \scriptsize
    \begin{center}
            \begin{tabular}{@{}C{2.4cm}@{}|@{}C{1.95cm}@{}|@{}C{1.95cm}@{}|@{}C{1.95cm}@{}|@{}C{1.95cm}@{}|@{}C{1.95cm}@{}}
                \Xhline{0.8pt}
                Kernel size & 31 & 51 & 55 & 75 & Average \\
                \hline
                \textbf{$\w_k$} & 27.89/0.932 & 26.31/0.901 & 25.91/0.916 & 26.73/0.908 & 26.71/0.914\\
                \textbf{$\z_k$} & 27.11/0.927 & 23.46/0.830 & 22.75/0.825 & 22.43/0.763 & 23.94/0.836\\
                \textbf{$\z_k~\mathrm{and}~\theta_k$} & 24.38/0.830 & 22.33/0.716 & 24.45/0.819 & 24.17/0.814 & 23.83/0.795\\
                \Xhline{0.8pt}
            \end{tabular}
    \end{center}
    \vspace{-6mm}
\end{table}
\begin{table}[t]
    \caption{Performance comparison (PSNR/SSIM) of different kernel initialization strategies on the dataset by Lai et al. \cite{lai2016comparative}.}
    \label{tab_lai_init}
    \vspace{-7mm}
    \begin{center}
        \scriptsize
        \begin{tabular}{@{}C{2.4cm}@{}|@{}C{2.4cm}@{}|@{}C{2.8cm}@{}|@{}C{2.8cm}@{}}
            \Xhline{0.8pt}
            Baseline & Random & $G(\z_0;\theta_k^{\ast})$ & $G(E(\y;\theta_E^{\ast});\theta_k^{\ast})$\\
            \hline
            Metrics & 19.51/0.563 & 23.00/0.743 & 26.71/0.914\\
            \Xhline{0.8pt}
        \end{tabular}
    \end{center}
    \vspace{-5mm}
\end{table}

\section{Conclusion}\label{sec:conclusion}
In this paper, we have proposed a new framework for the BID task, by virtue of deep generative model. Within this framework, we first pre-train a kernel generator as a DGP for blur kernels and a kernel initializer that can offer a well-initialized kernel. Then during the BID process, the blur kernel is initialized in the latent feature space and jointly optimized with the DIP network for the final result. Comprehensive experiments have been conducted and demonstrated the effectiveness of the proposed method. The ablation study verifies the necessity of each component 
in our framework. 
In the future, we will try to generalize our method to non-uniform blur modeling, overcoming its current limitations.


\section*{Acknowledgments}
This work was supported in part by the NSFC Projects (No. 12226004, 62076196, 62331028, 62272375). The authors thank Prof. Jiangxin Dong for providing the code for generating the blur kernels with the type of Lai dataset.

%
%

\bibliographystyle{splncs04}
\bibliography{main}

\newpage
\appendix

\section{Model efficiency}
Here we compare the model efficiency of the proposed method with existing DIP-based BID methods, in terms of the number of parameters, floating point operations (FLOPs), and computational time. The evaluations were conducted using NVIDIA 3090 RTX GPU, and the results are summarized in Table \ref{tab_efficiency}. As can be seen from the table, all the DIP-based methods, including ours, have similar model efficiency. However, as is well-known, the computational complexity of DIP-based methods is generally much higher than that of the training-based deep learning ones, which limits their real applications and should be paid more attention in future research.

\begin{table}[h]
	\vspace{-4mm}
	\caption{Model efficiency comparison of DIP-based methods on the dataset by Lai \cite{lai2016comparative}.}\label{tab_efficiency}
	\vspace{-6mm}
	\begin{center}
		\scriptsize
		\begin{tabular}{@{}C{5.0cm}@{}|@{}C{1.4cm}@{}|@{}C{1.9cm}@{}|@{}C{2.0cm}@{}|@{}C{1.7cm}@{}}
			\Xhline{0.8pt}
			& Ours & Ren et al. \cite{ren2020neural} & Huo et al. \cite{huo2023blind} & Li et al. \cite{li2023self}\\
			\hline
			\#Params (DIP net + Kernel net) / M & 2.3 + 1.0 & 2.3 + 1.1 & 2.3 + 1.1 & 2.2 + 0.9\\
			FLOPs (per iteration) / G & 217.8 & 217.6 & 217.0 & 215.9\\
			Time (per iteration) / s & 0.40 & 0.37 & 0.46 & 0.37\\
			\Xhline{0.8pt}
		\end{tabular}
	\end{center}
	\vspace{-10mm}
\end{table}

\section{Results on our synthetic dataset and Lai benchmark}
\begin{figure}[t]
	\centering
	\subfloat[Blurred]
	{\includegraphics[width=0.23\textwidth]{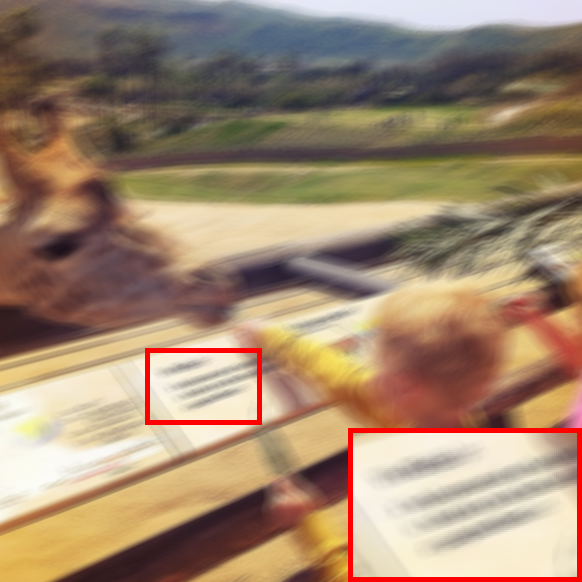}}\
	\subfloat[Pan \cite{pan2016blind}]
	{\includegraphics[width=0.23\textwidth]{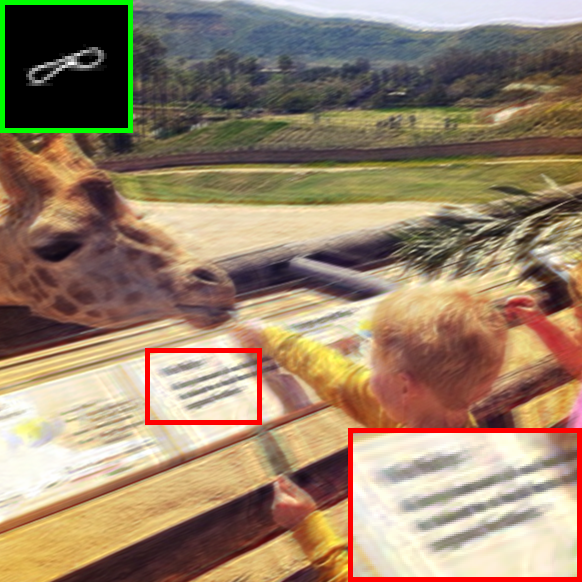}}\
	\subfloat[Dong \cite{dong2017blind}]
	{\includegraphics[width=0.23\textwidth]{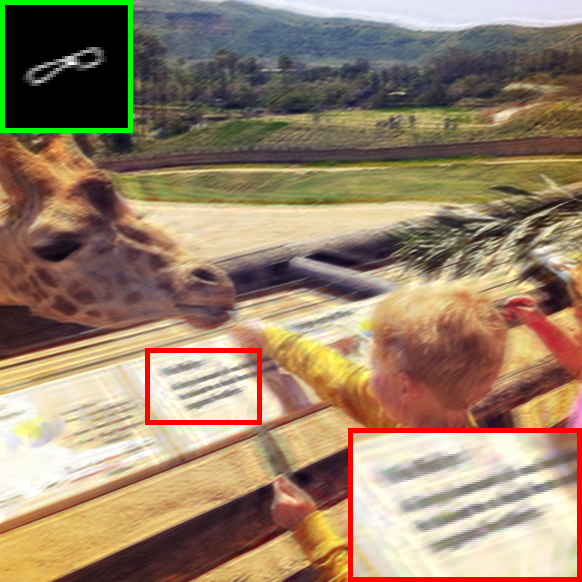}}\
	\subfloat[Tao \cite{tao2018scale}]
	{\includegraphics[width=0.23\textwidth]{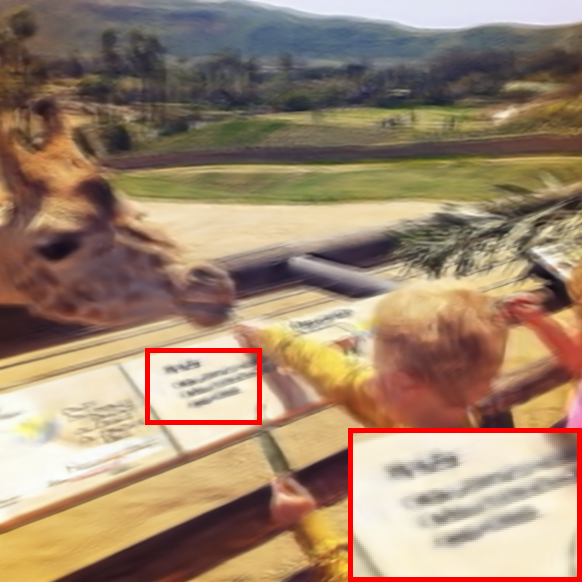}}
	\\
	\subfloat[Kupyn \cite{kupyn2019deblurgan}]
	{\includegraphics[width=0.23\textwidth]{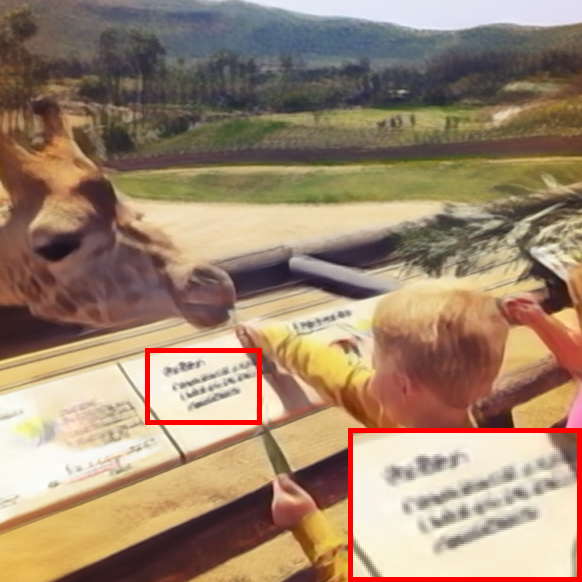}}\
	\subfloat[Kaufman \cite{kaufman2020deblurring}]
	{\includegraphics[width=0.23\textwidth]{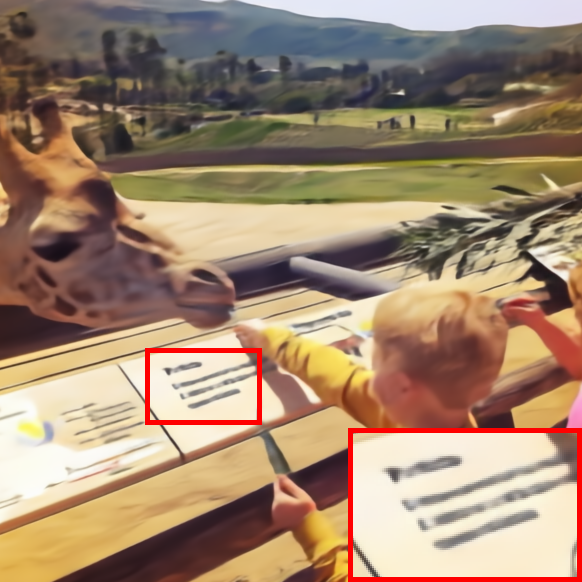}}\
	\subfloat[Zamir \cite{zamir2022restormer}]
	{\includegraphics[width=0.23\textwidth]{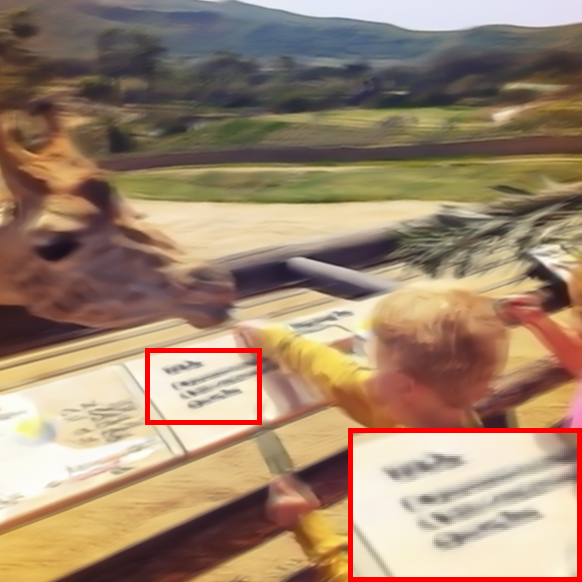}}\
	\subfloat[Ren \cite{ren2020neural}]
	{\includegraphics[width=0.23\textwidth]{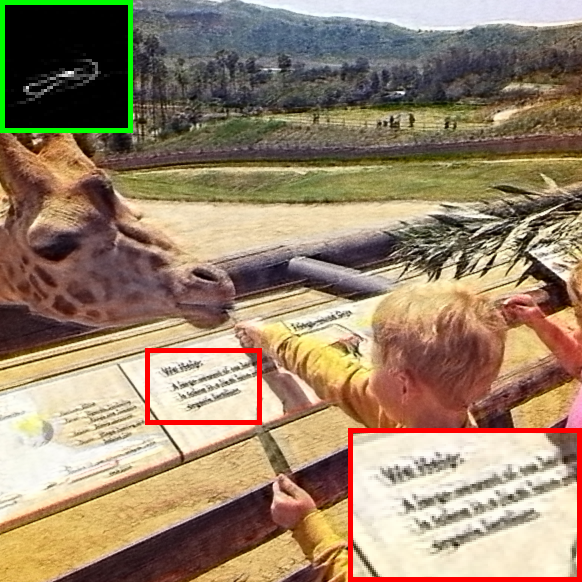}}
	\\
	\subfloat[Huo \cite{huo2023blind}]
	{\includegraphics[width=0.23\textwidth]{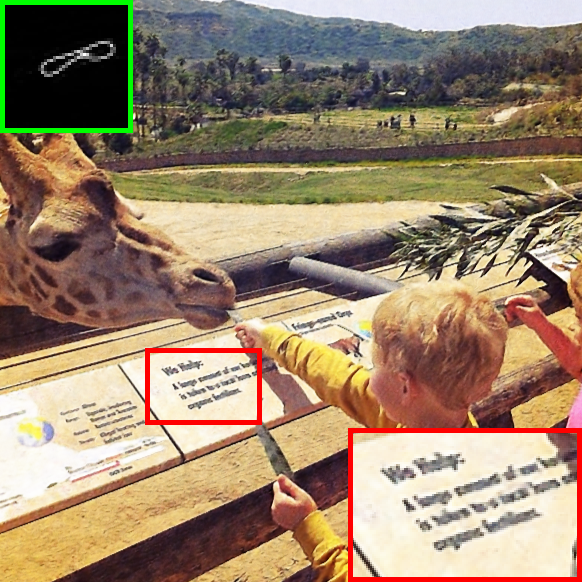}}\
	\subfloat[Li \cite{li2023self}]
	{\includegraphics[width=0.23\textwidth]{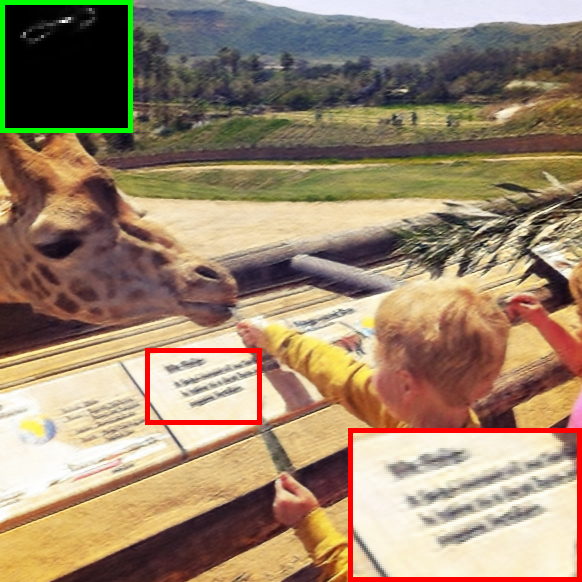}}\
	\subfloat[Ours]
	{\includegraphics[width=0.23\textwidth]{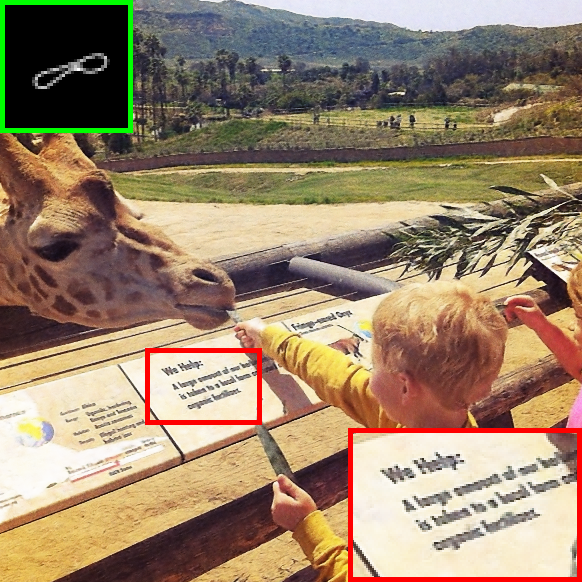}}\
	\subfloat[Ground truth]
	{\includegraphics[width=0.23\textwidth]{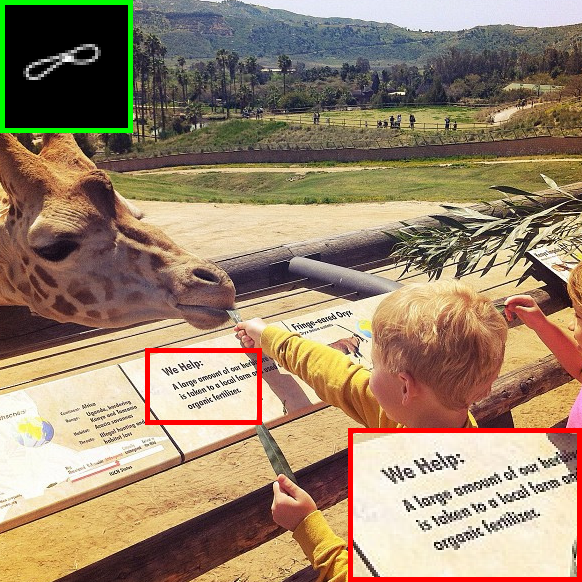}}
	\vspace{-3mm}
	\caption{Visual results on our synthetic dataset. The estimated blur kernel is placed on the top-left corner for each method if available.}\label{fig_syn2}
\end{figure}

\begin{figure}[t]
	\centering
	\subfloat[Blurred]
	{\includegraphics[width=0.23\textwidth]{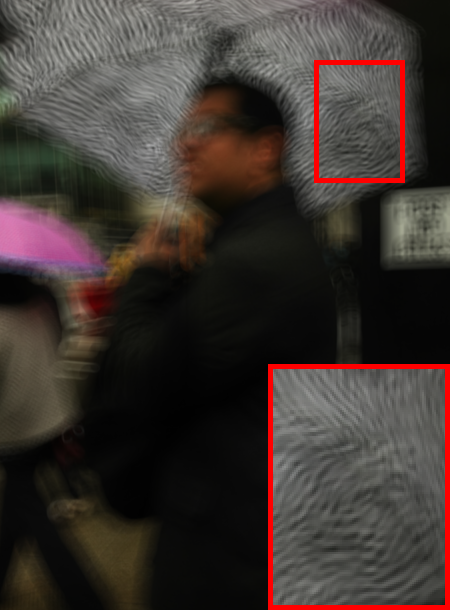}}\
	\subfloat[Pan \cite{pan2016blind}]
	{\includegraphics[width=0.23\textwidth]{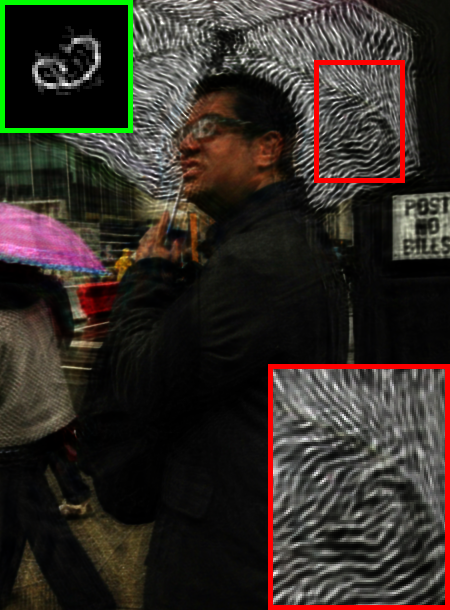}}\
	\subfloat[Dong \cite{dong2017blind}]
	{\includegraphics[width=0.23\textwidth]{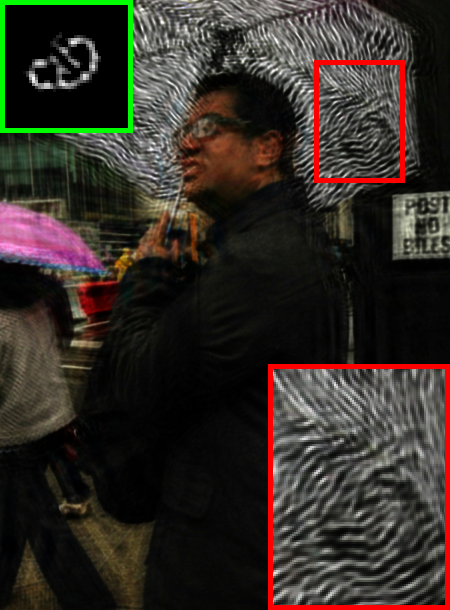}}\
	\subfloat[Tao \cite{tao2018scale}]
	{\includegraphics[width=0.23\textwidth]{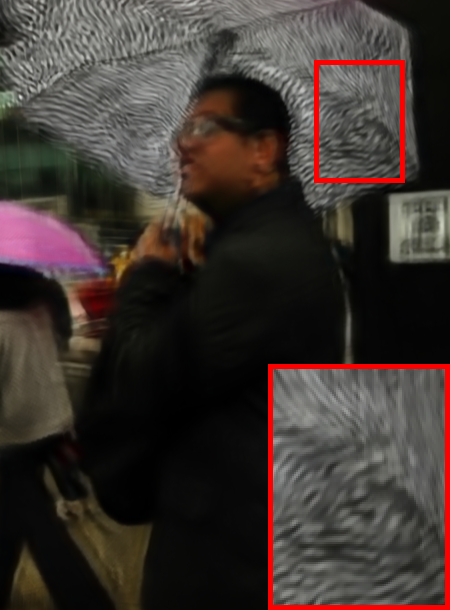}}
	\\
	\subfloat[Kupyn \cite{kupyn2019deblurgan}]
	{\includegraphics[width=0.23\textwidth]{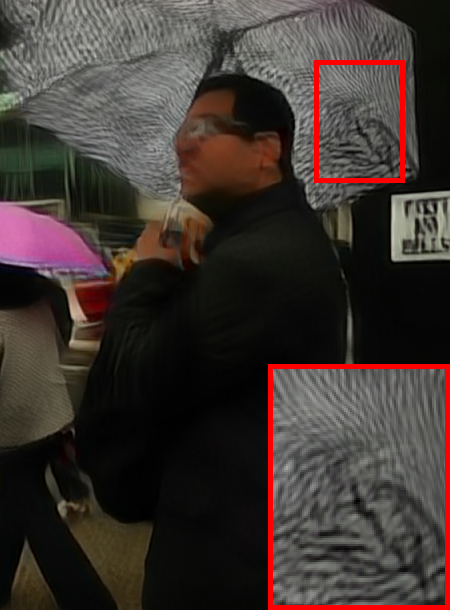}}\
	\subfloat[Kaufman \cite{kaufman2020deblurring}]
	{\includegraphics[width=0.23\textwidth]{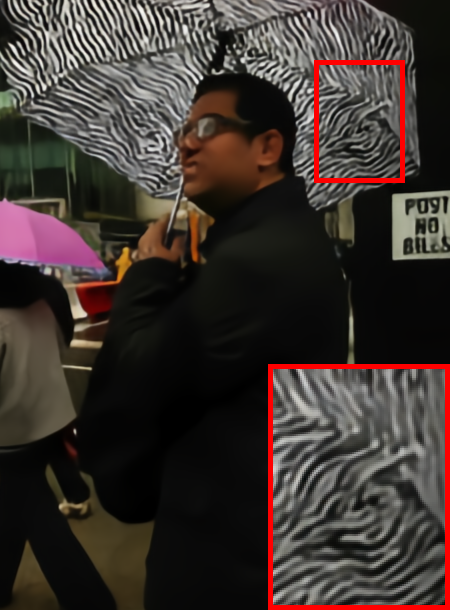}}\
	\subfloat[Zamir \cite{zamir2022restormer}]
	{\includegraphics[width=0.23\textwidth]{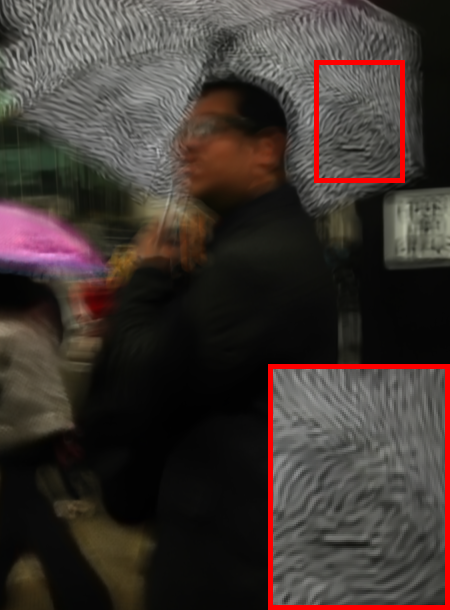}}\
	\subfloat[Ren \cite{ren2020neural}]
	{\includegraphics[width=0.23\textwidth]{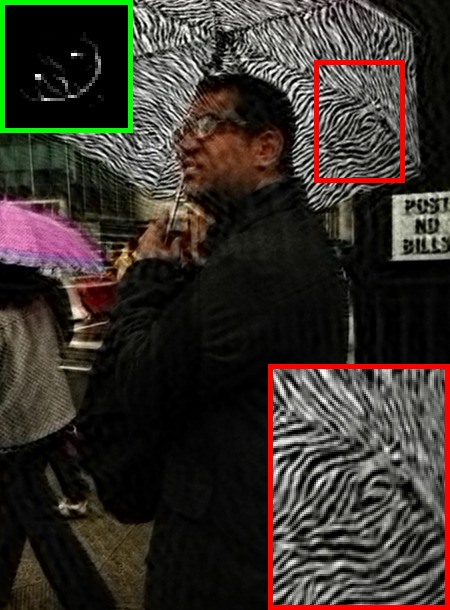}}
	\\
	\subfloat[Huo \cite{huo2023blind}]
	{\includegraphics[width=0.23\textwidth]{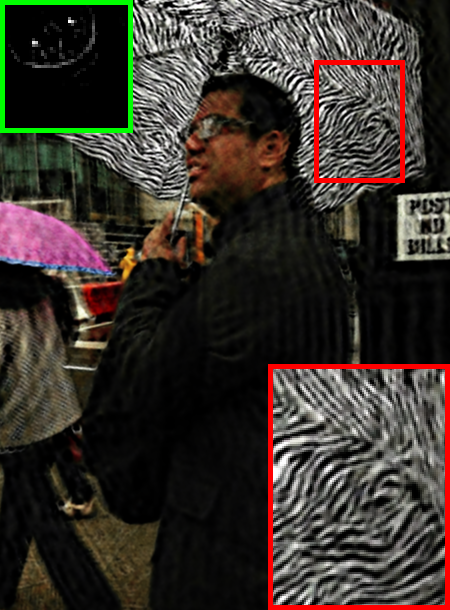}}\
	\subfloat[Li \cite{li2023self}]
	{\includegraphics[width=0.23\textwidth]{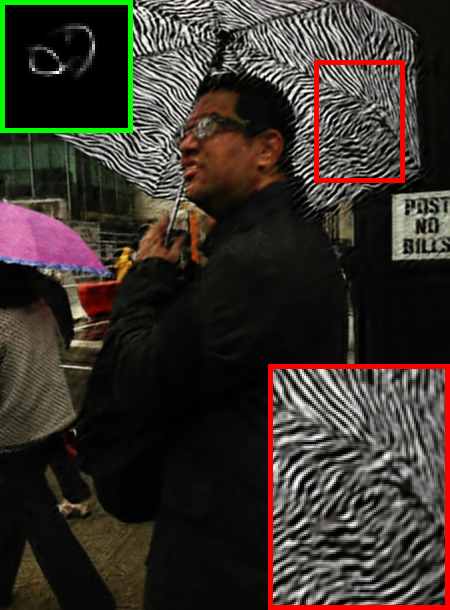}}\
	\subfloat[Ours]
	{\includegraphics[width=0.23\textwidth]{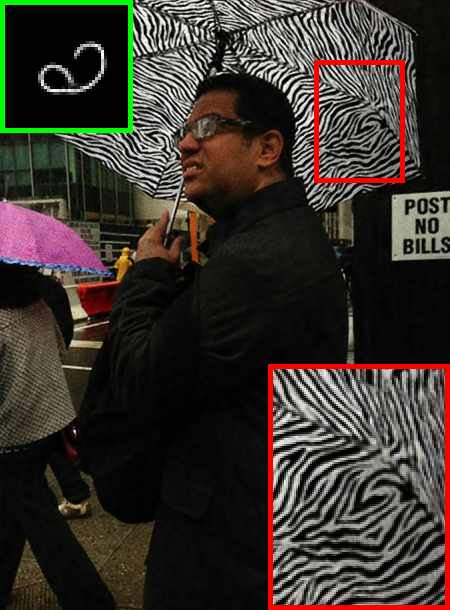}}\
	\subfloat[Ground truth]
	{\includegraphics[width=0.23\textwidth]{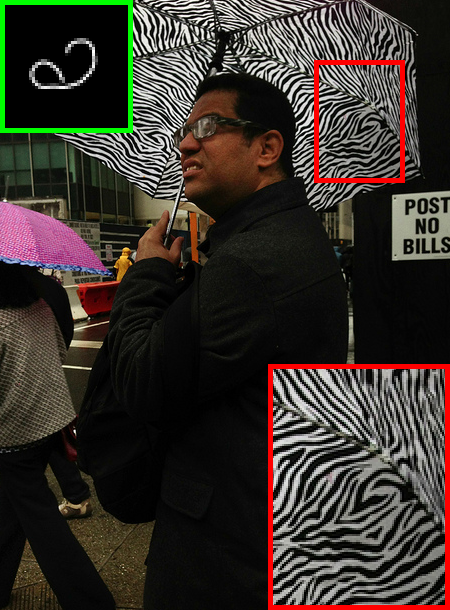}}
	\vspace{-3mm}
	\caption{Visual results on our synthetic dataset. The estimated blur kernel is placed on the top-left corner for each method if available.}\label{fig_syn3}
\end{figure}

\begin{figure}[h]
	\centering
	\subfloat[Blurred]
	{\includegraphics[width=0.23\textwidth]{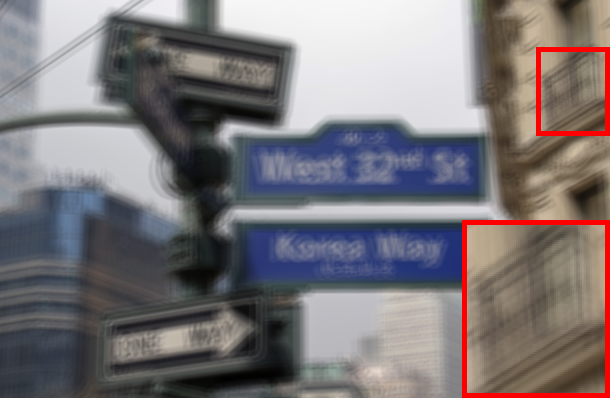}}\
	\subfloat[Pan \cite{pan2016blind}]
	{\includegraphics[width=0.23\textwidth]{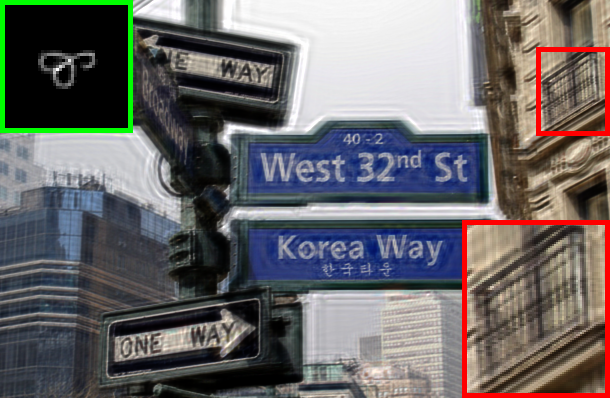}}\
	\subfloat[Dong \cite{dong2017blind}]
	{\includegraphics[width=0.23\textwidth]{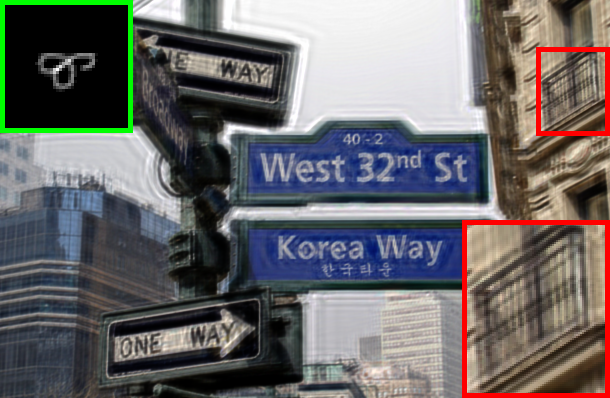}}\
	\subfloat[Tao \cite{tao2018scale}]
	{\includegraphics[width=0.23\textwidth]{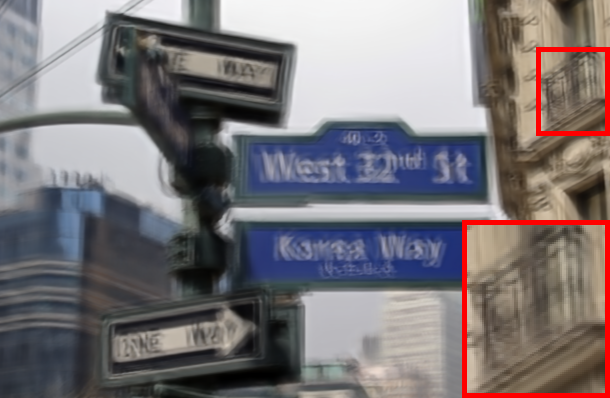}}
	\\
	\subfloat[Kupyn \cite{kupyn2019deblurgan}]
	{\includegraphics[width=0.23\textwidth]{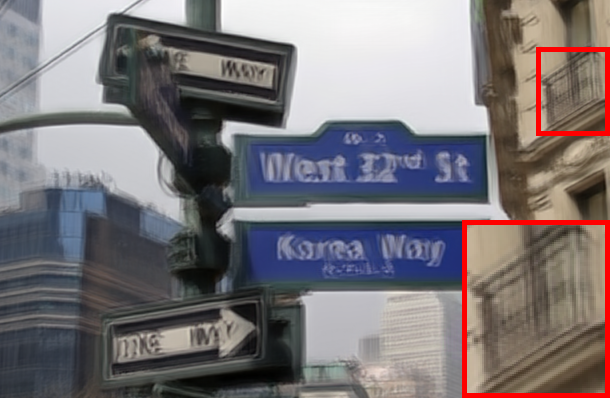}}\
	\subfloat[Kaufman \cite{kaufman2020deblurring}]
	{\includegraphics[width=0.23\textwidth]{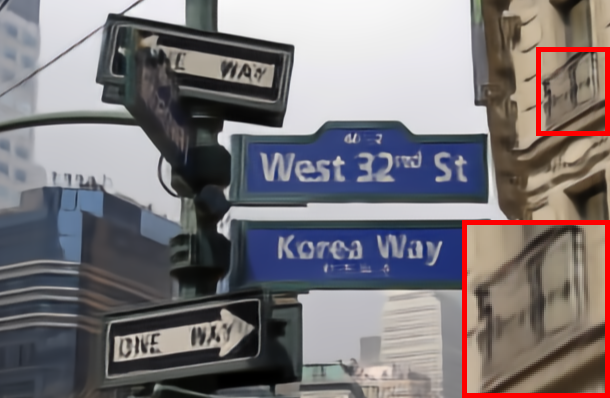}}\
	\subfloat[Zamir \cite{zamir2022restormer}]
	{\includegraphics[width=0.23\textwidth]{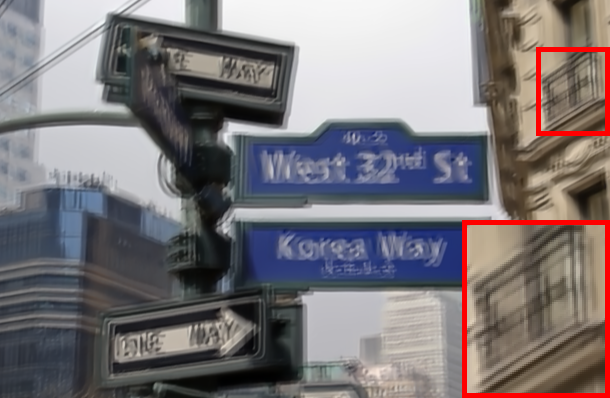}}\
	\subfloat[Ren \cite{ren2020neural}]
	{\includegraphics[width=0.23\textwidth]{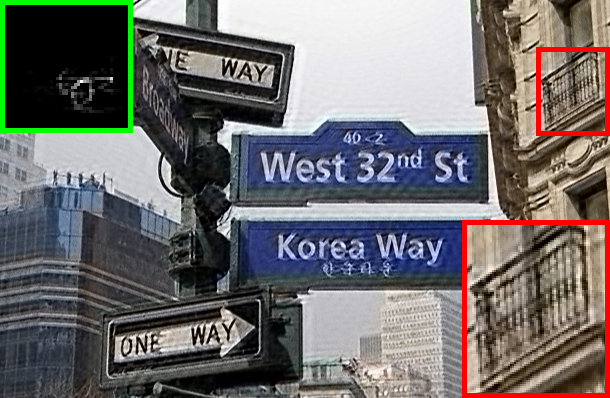}}
	\\
	\subfloat[Huo \cite{huo2023blind}]
	{\includegraphics[width=0.23\textwidth]{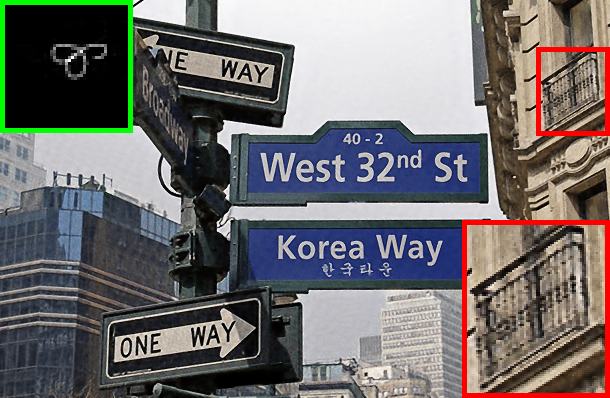}}\
	\subfloat[Li \cite{li2023self}]
	{\includegraphics[width=0.23\textwidth]{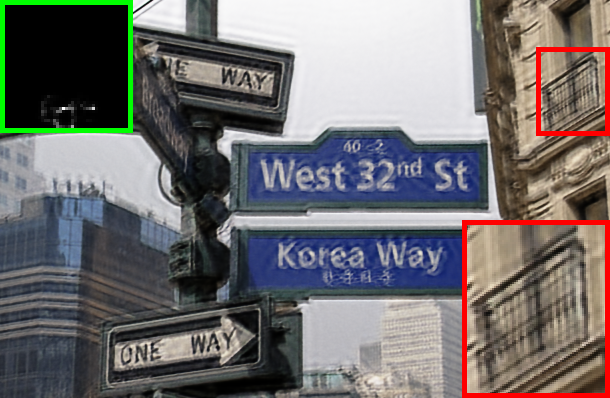}}\
	\subfloat[Ours]
	{\includegraphics[width=0.23\textwidth]{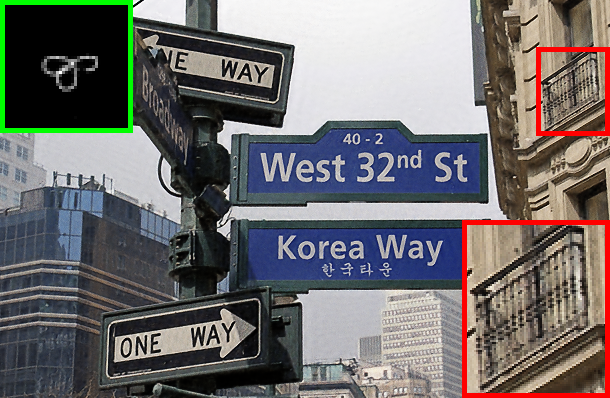}}\
	\subfloat[Ground truth]
	{\includegraphics[width=0.23\textwidth]{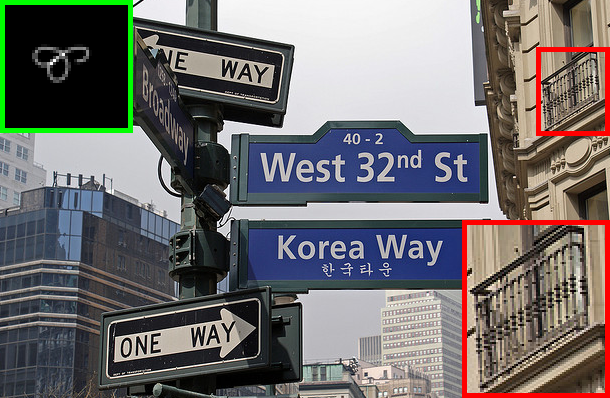}}
	\vspace{-3mm}
	\caption{Visual results on our synthetic dataset. The estimated blur kernel is placed on the top-left corner for each method if available.}\label{fig_syn4}
\end{figure}

\begin{figure}[!h]
	\centering
	\subfloat[Blurred]
	{\includegraphics[width=0.24\textwidth]{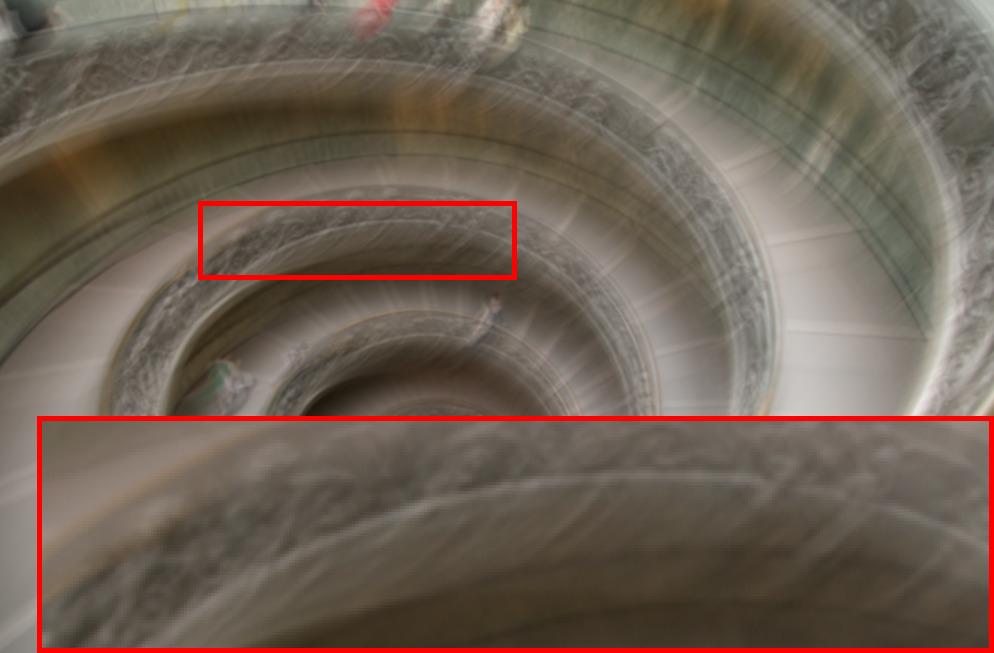}}\
	\subfloat[Cho \cite{cho2009fast}]
	{\includegraphics[width=0.24\textwidth]{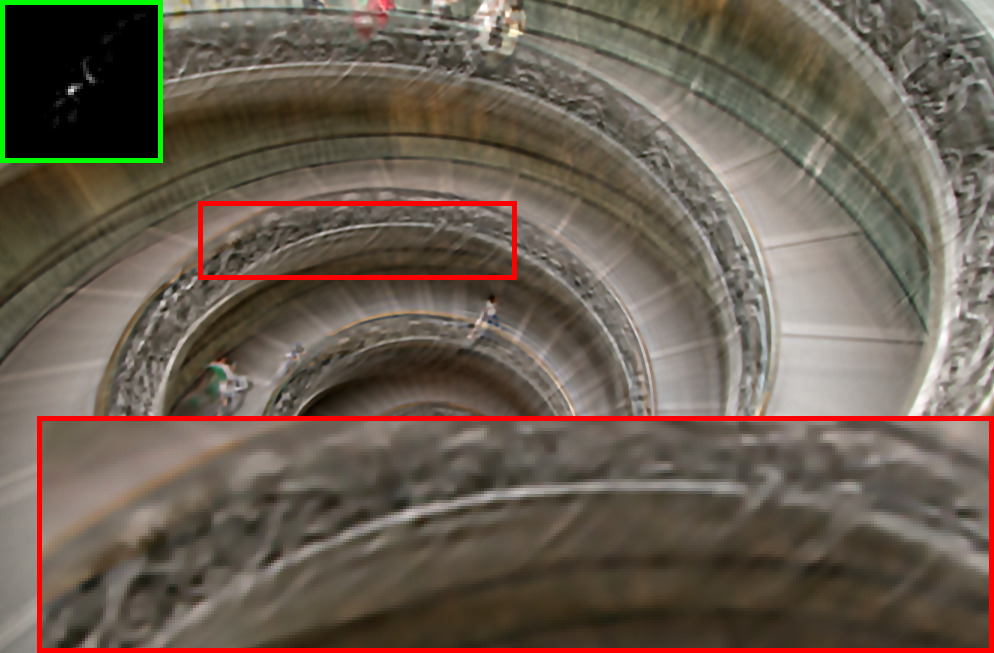}}\
	\subfloat[Krishnan \cite{krishnan2011blind}]
	{\includegraphics[width=0.24\textwidth]{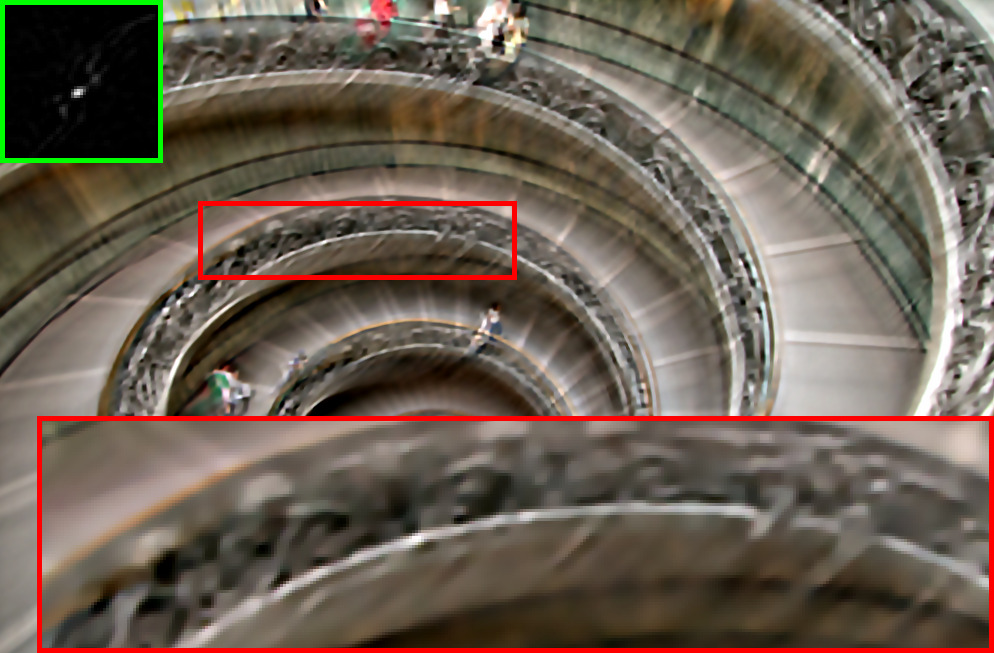}}\
	\subfloat[Xu \cite{xu2013unnatural}]
	{\includegraphics[width=0.24\textwidth]{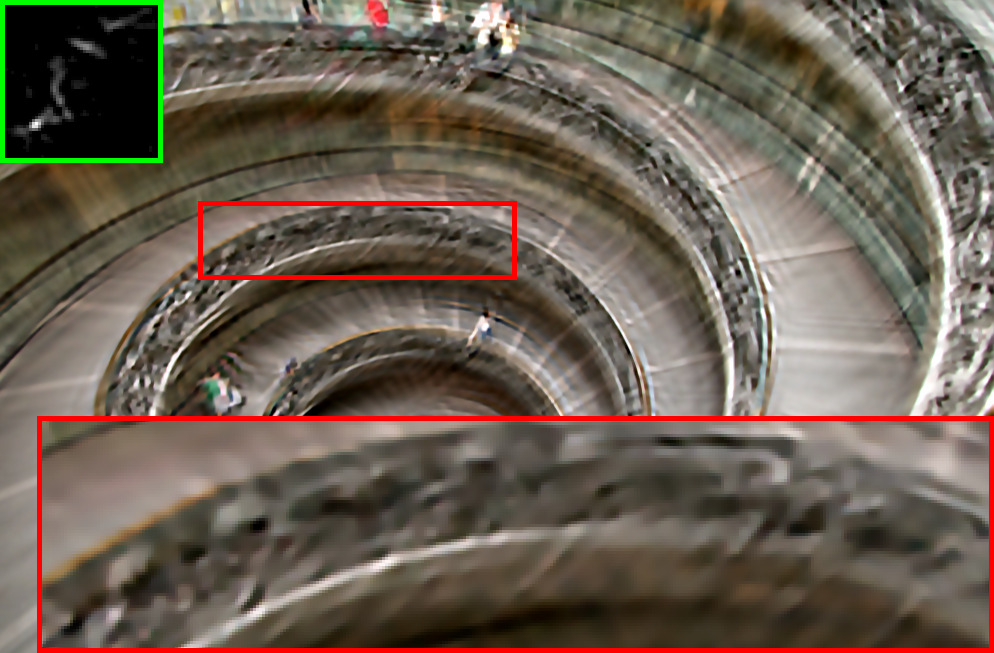}}
	\\
	\subfloat[Perrone \cite{perrone2014total}]
	{\includegraphics[width=0.24\textwidth]{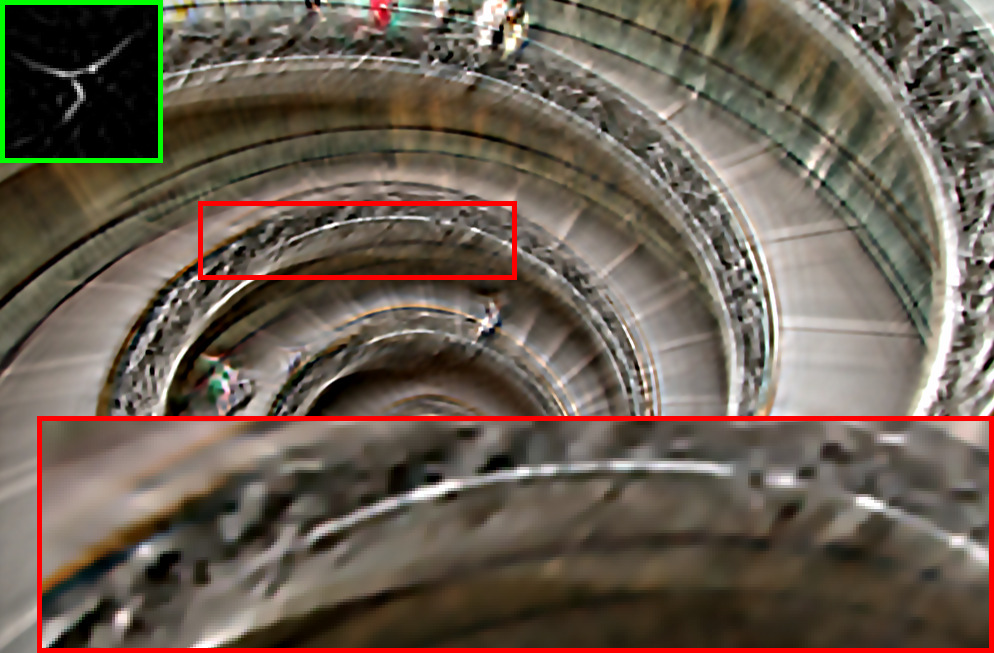}}\
	\subfloat[Pan \cite{pan2016blind}]
	{\includegraphics[width=0.24\textwidth]{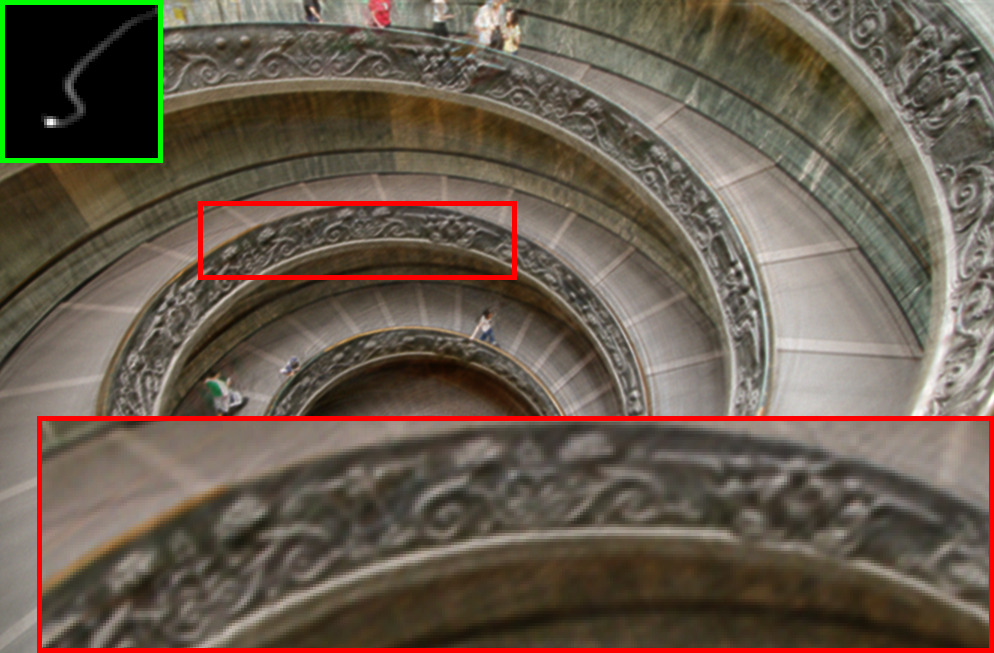}}\
	\subfloat[Dong \cite{dong2017blind}]
	{\includegraphics[width=0.24\textwidth]{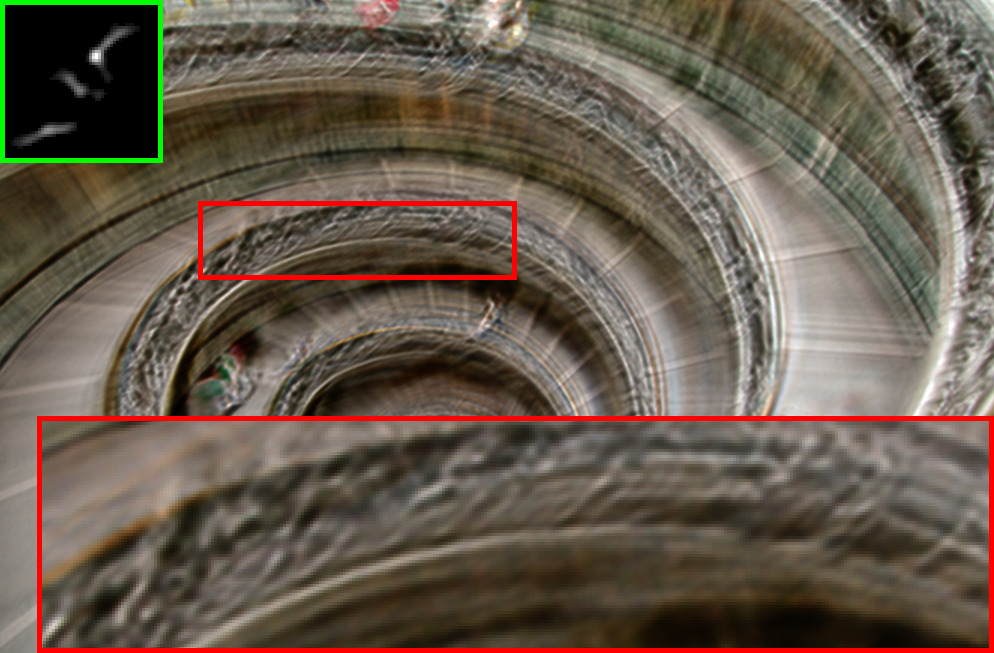}}\
	\subfloat[Tao \cite{tao2018scale}]
	{\includegraphics[width=0.24\textwidth]{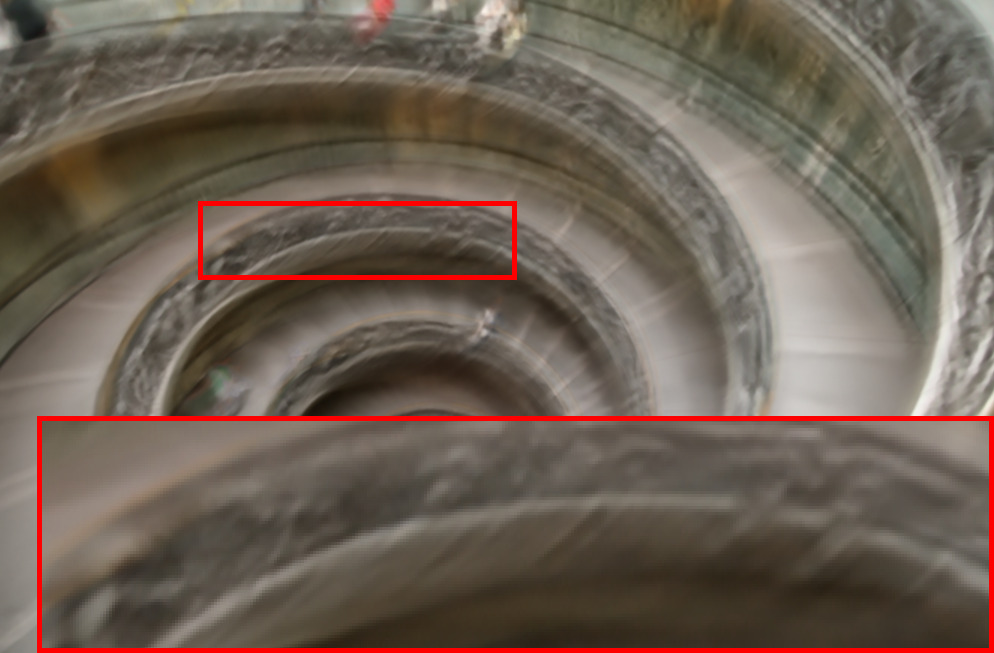}}
	\\
	\subfloat[Kupyn \cite{kupyn2019deblurgan}]
	{\includegraphics[width=0.24\textwidth]{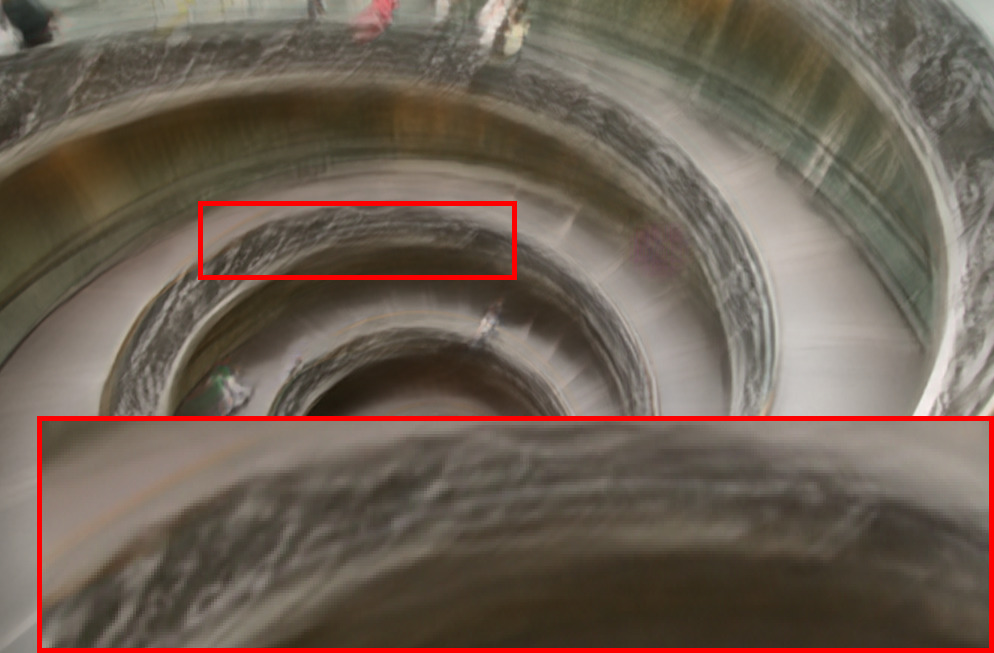}}\
	\subfloat[Kaufman \cite{kaufman2020deblurring}]
	{\includegraphics[width=0.24\textwidth]{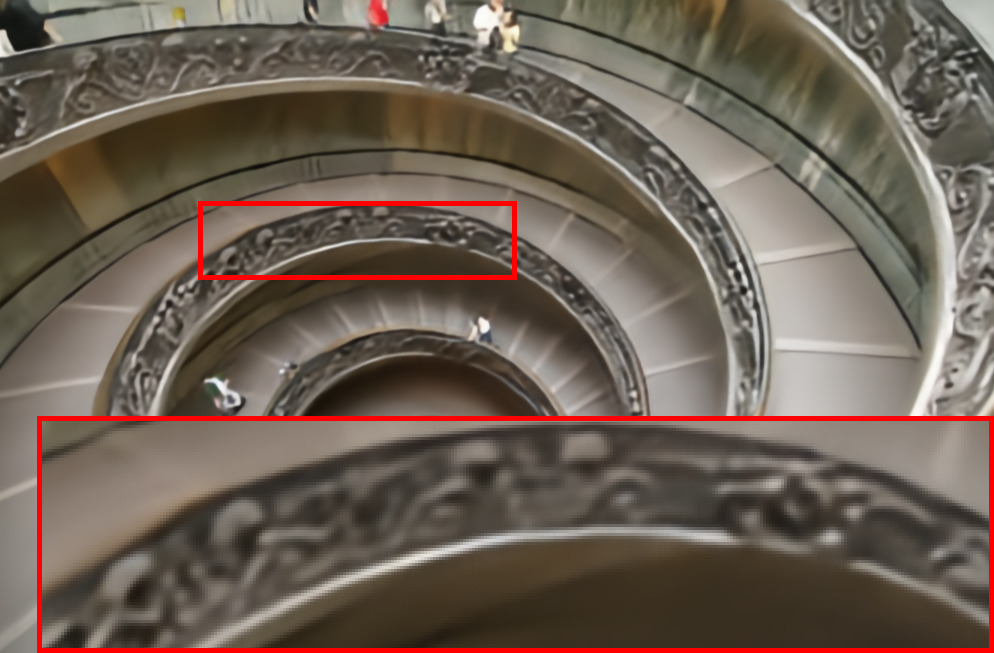}}\
	\subfloat[Zamir \cite{zamir2022restormer}]
	{\includegraphics[width=0.24\textwidth]{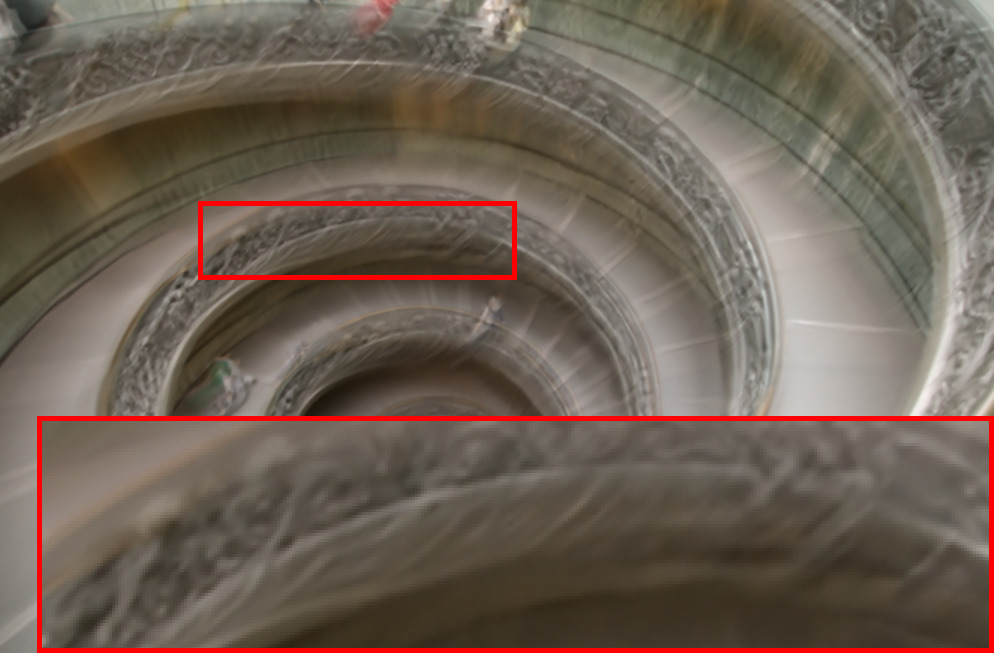}}\
	\subfloat[Ren \cite{ren2020neural}]
	{\includegraphics[width=0.24\textwidth]{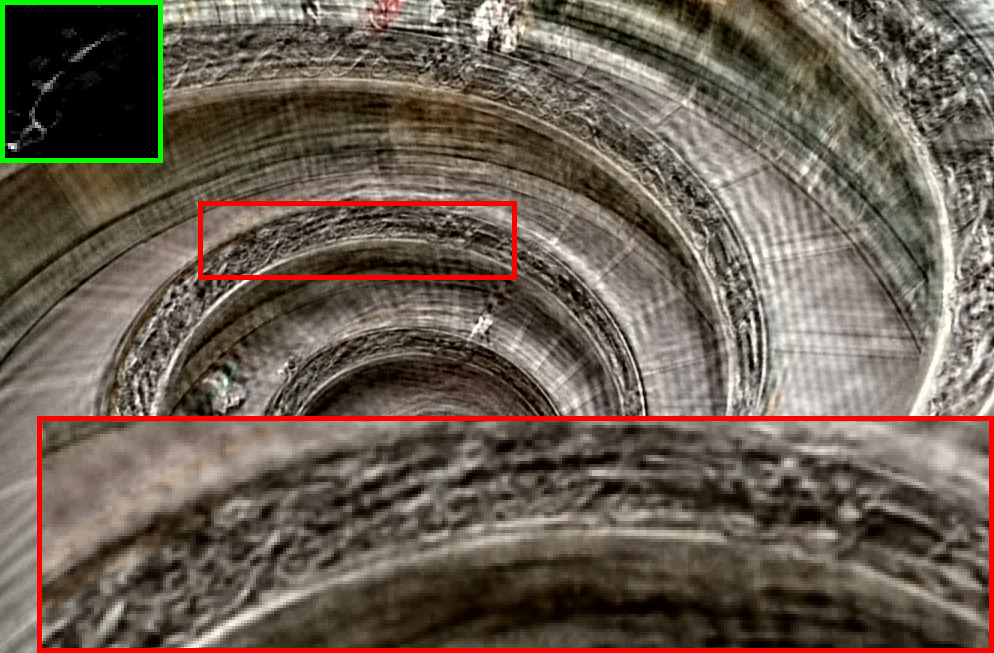}}
	\\
	\subfloat[Huo \cite{huo2023blind}]
	{\includegraphics[width=0.24\textwidth]{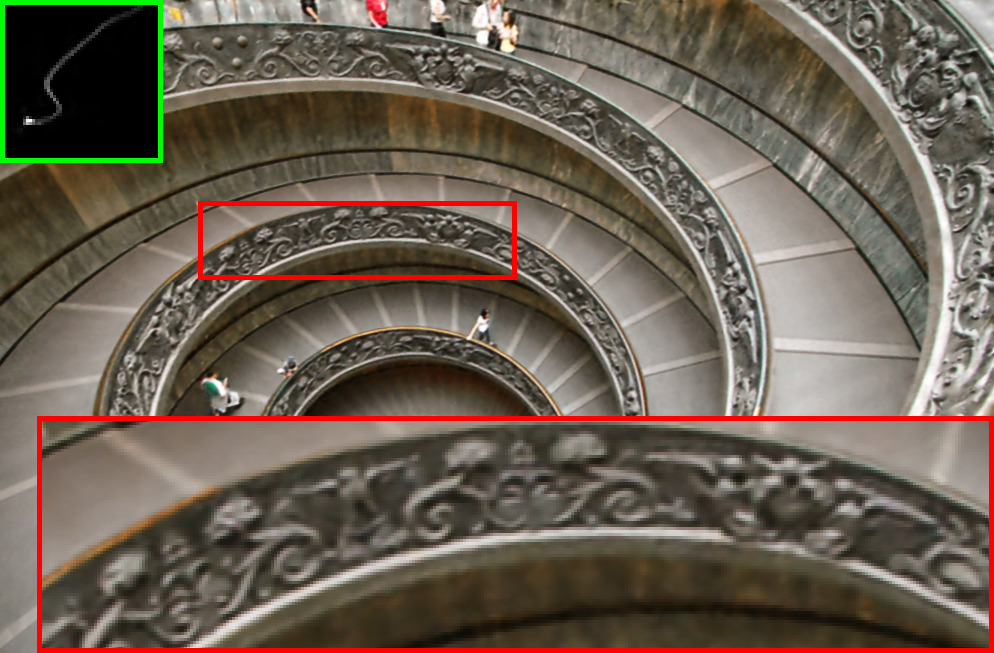}}\
	\subfloat[Li \cite{li2023self}]
	{\includegraphics[width=0.24\textwidth]{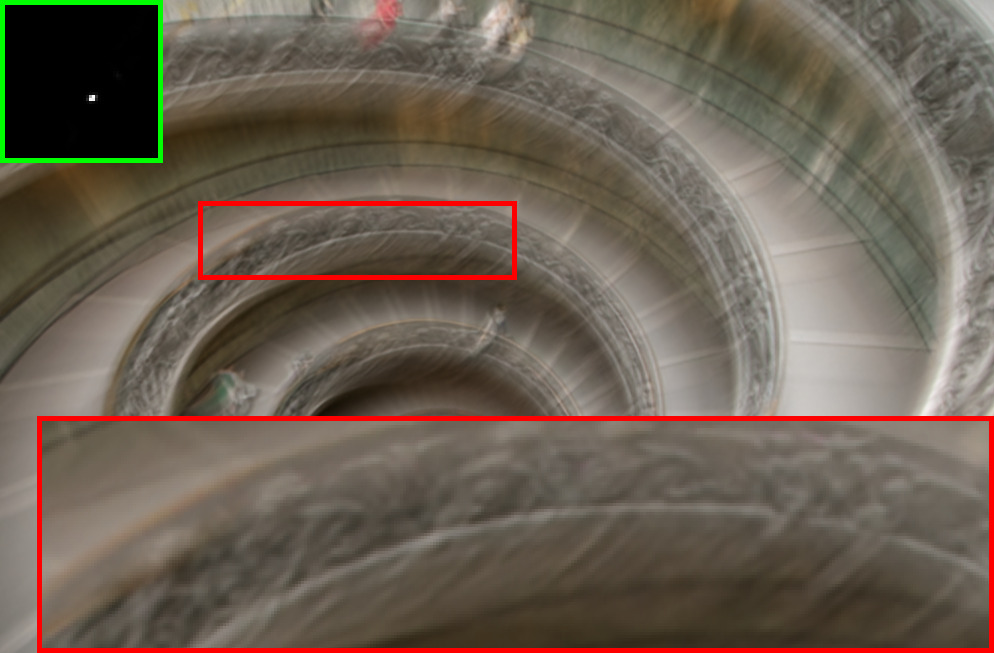}}\
	\subfloat[Ours]
	{\includegraphics[width=0.24\textwidth]{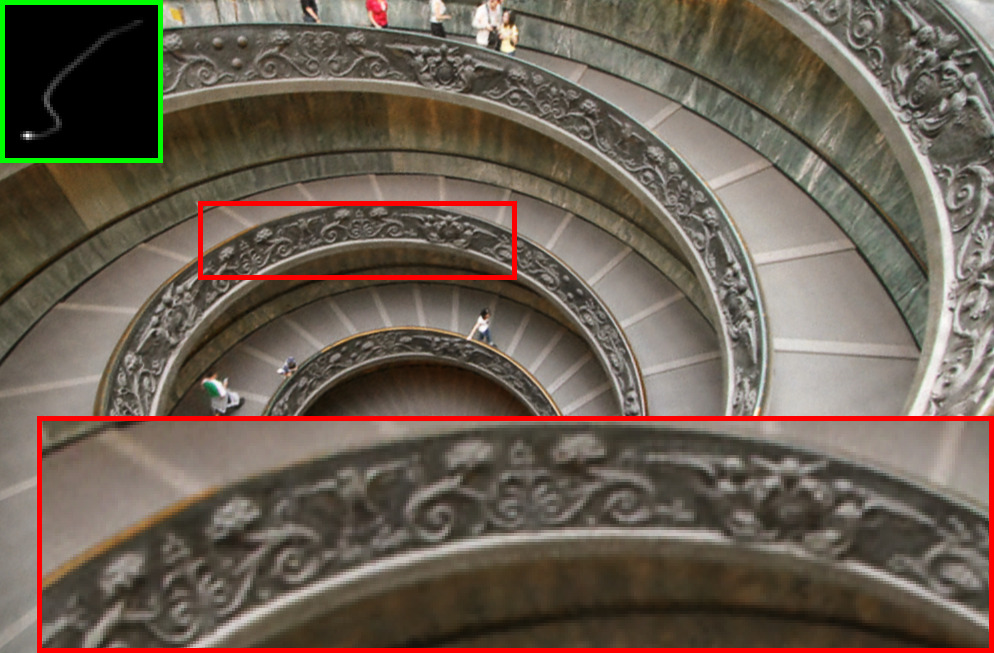}}\
	\subfloat[Ground truth]
	{\includegraphics[width=0.24\textwidth]{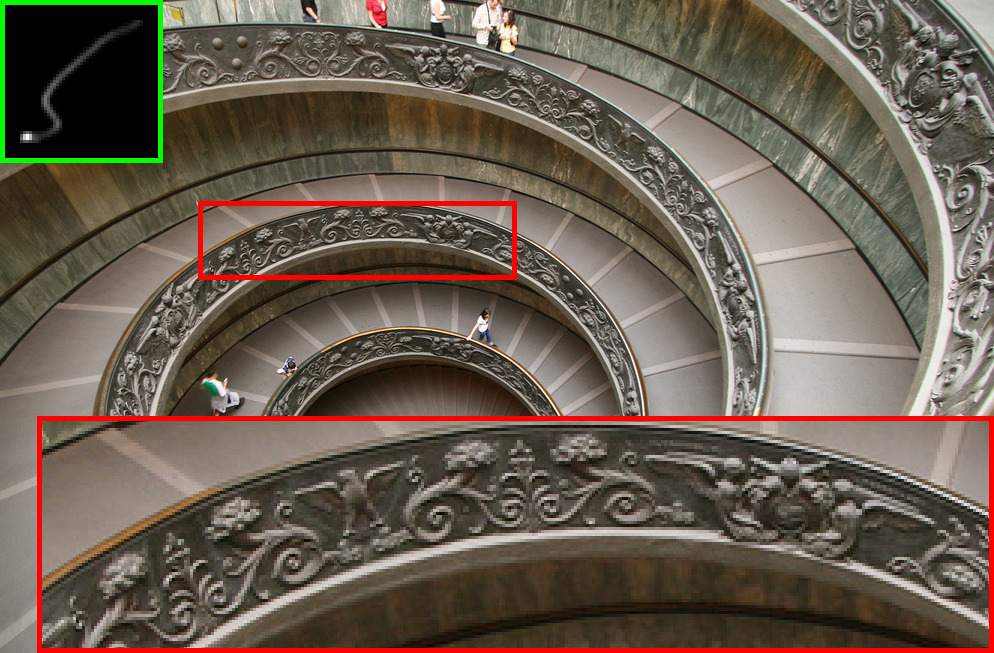}}
	\caption{Visual results on the dataset by Lai et al. \cite{lai2016comparative}. The estimated blur kernel is placed on the top-left corner of each method if available.}
	\label{fig_lai2}
\end{figure}

\begin{figure}[t]
	\centering
	\subfloat[Blurred]
	{\includegraphics[width=0.24\textwidth]{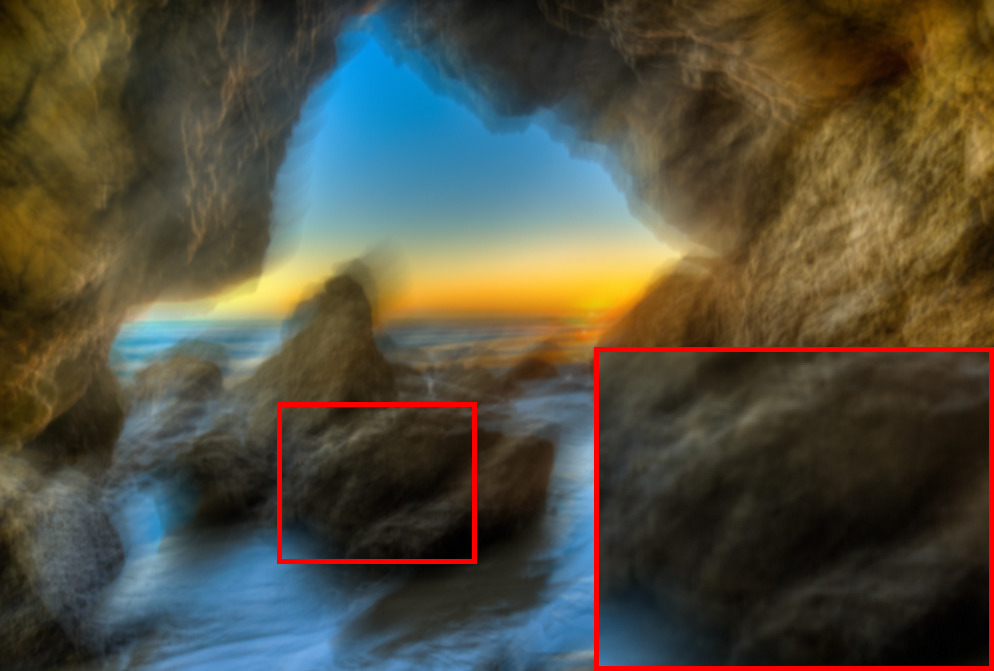}}\
	\subfloat[Cho \cite{cho2009fast}]
	{\includegraphics[width=0.24\textwidth]{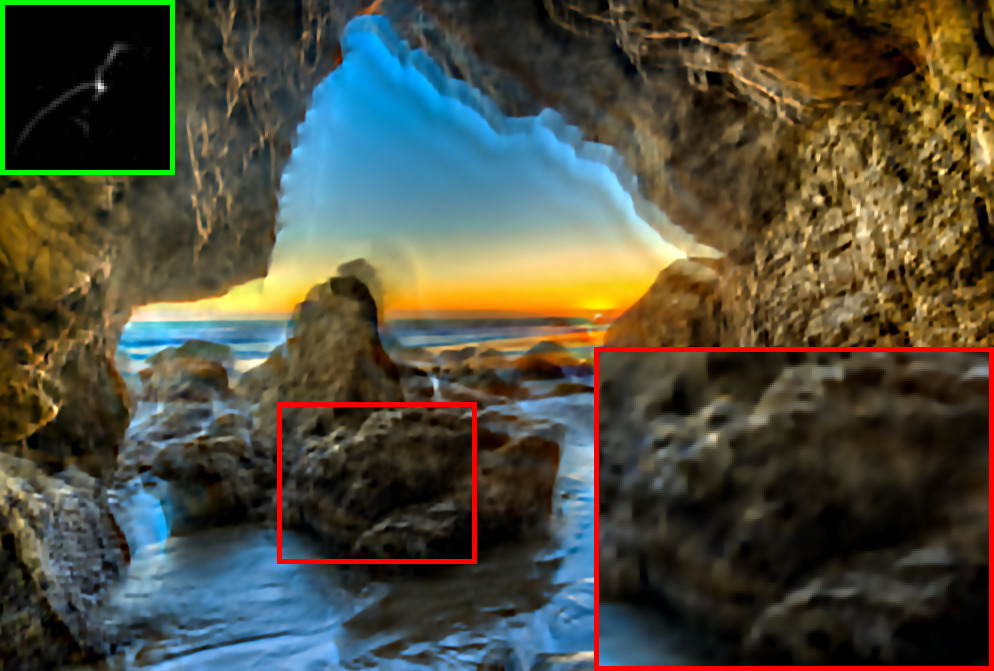}}\
	\subfloat[Krishnan \cite{krishnan2011blind}]
	{\includegraphics[width=0.24\textwidth]{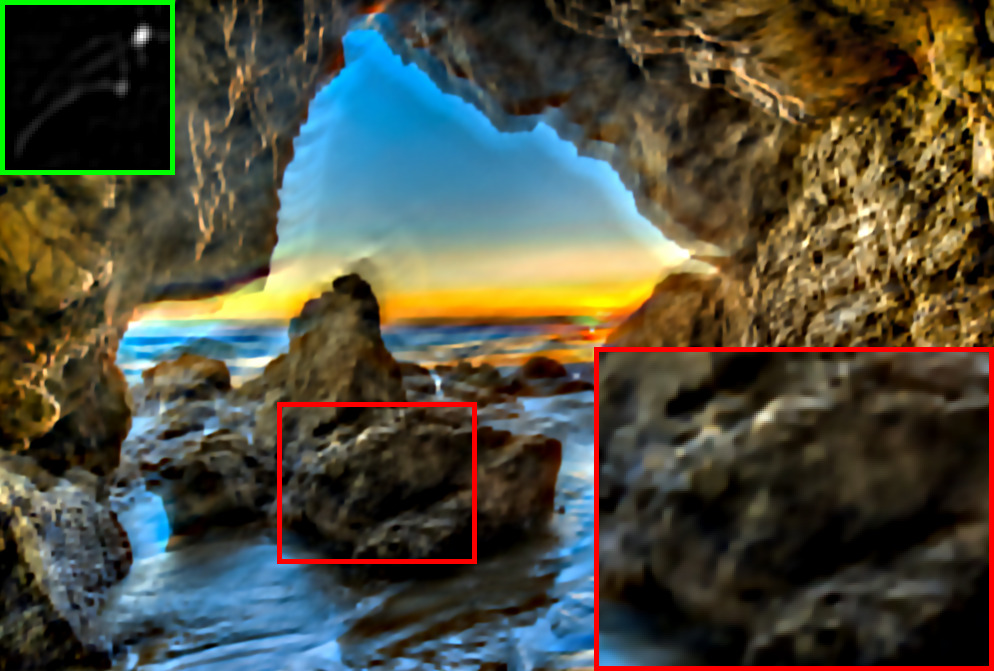}}\
	\subfloat[Xu \cite{xu2013unnatural}]
	{\includegraphics[width=0.24\textwidth]{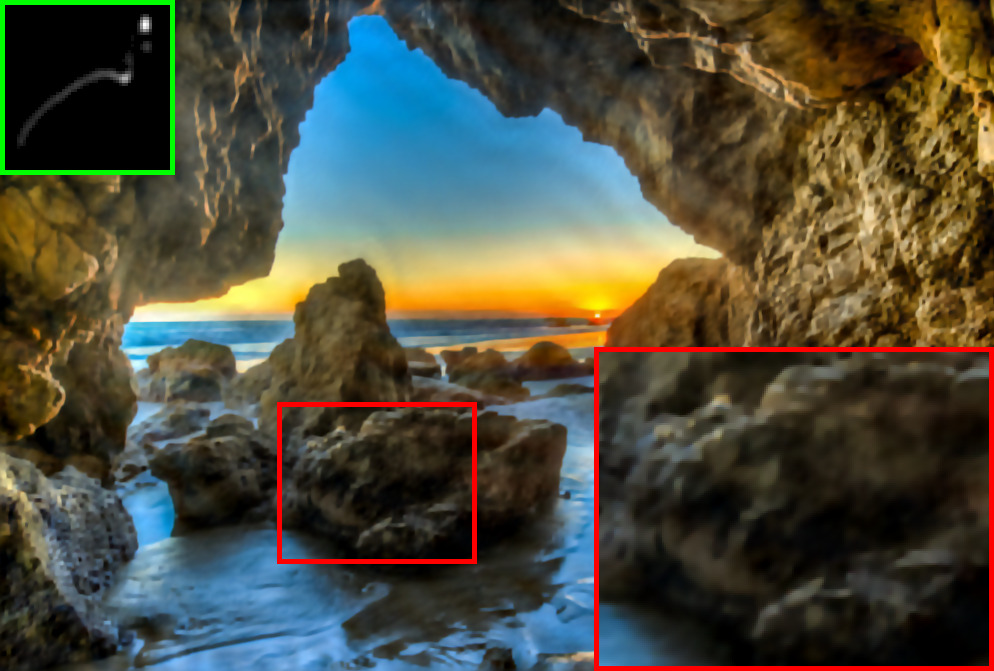}}
	\\
	\subfloat[Perrone \cite{perrone2014total}]
	{\includegraphics[width=0.24\textwidth]{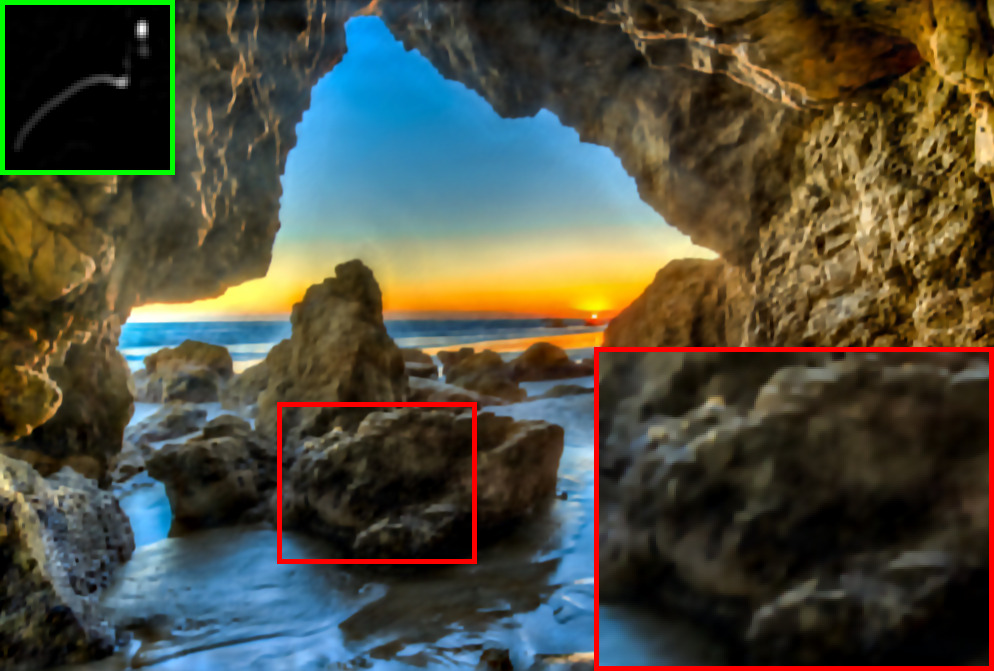}}\
	\subfloat[Pan \cite{pan2016blind}]
	{\includegraphics[width=0.24\textwidth]{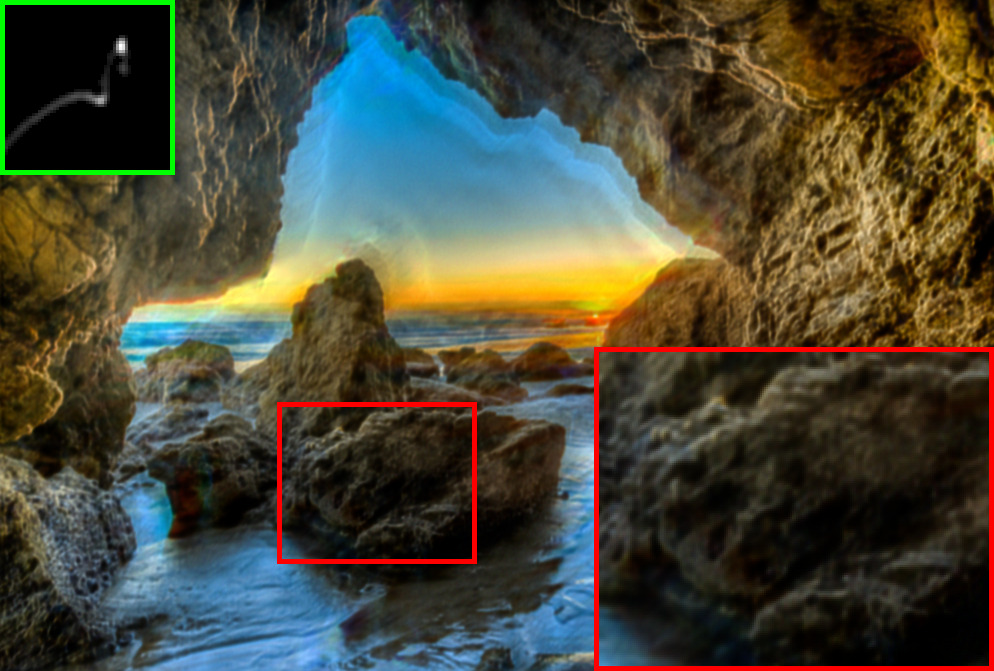}}\
	\subfloat[Dong \cite{dong2017blind}]
	{\includegraphics[width=0.24\textwidth]{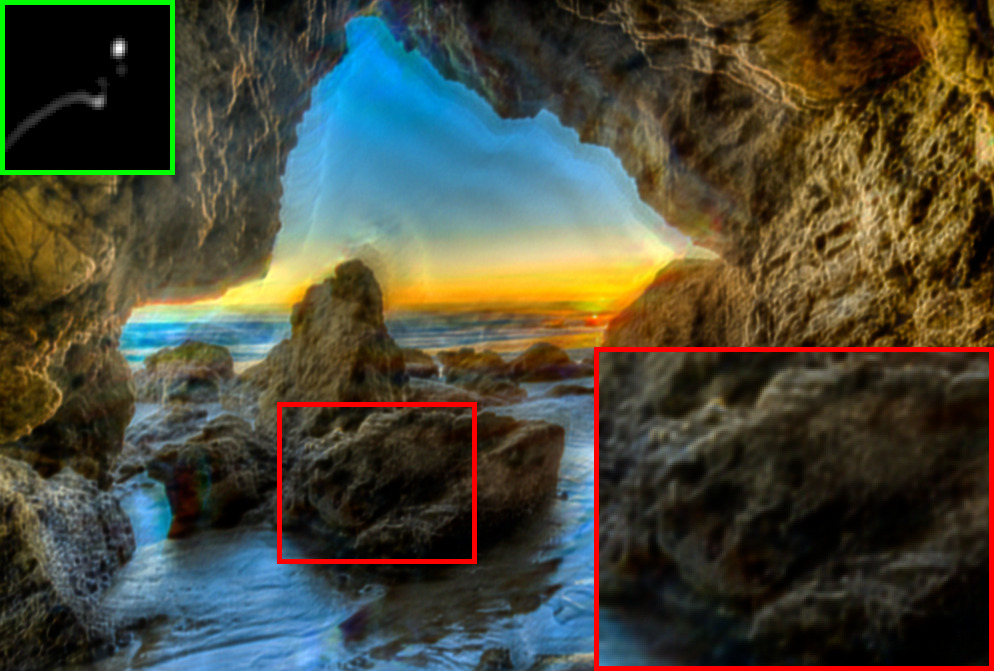}}\
	\subfloat[Tao \cite{tao2018scale}]
	{\includegraphics[width=0.24\textwidth]{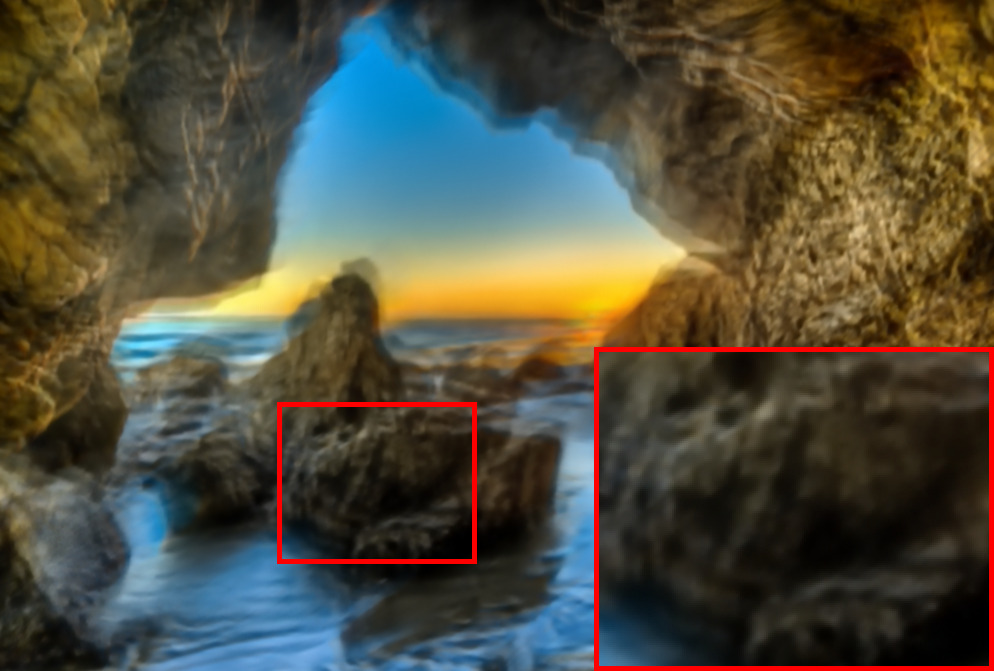}}
	\\
	\subfloat[Kupyn \cite{kupyn2019deblurgan}]
	{\includegraphics[width=0.24\textwidth]{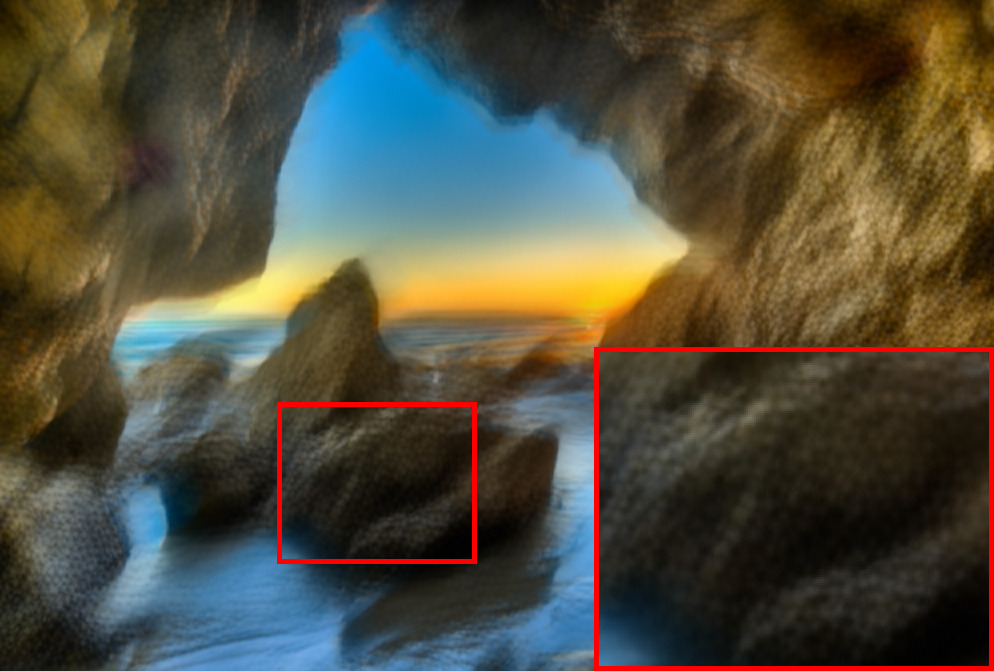}}\
	\subfloat[Kaufman \cite{kaufman2020deblurring}]
	{\includegraphics[width=0.24\textwidth]{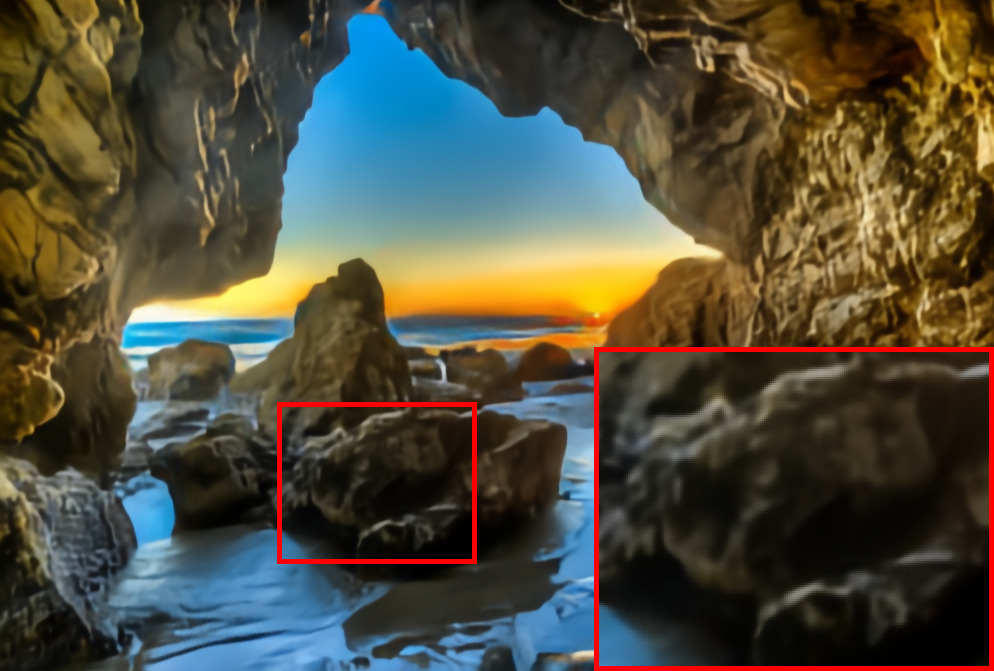}}\
	\subfloat[Zamir \cite{zamir2022restormer}]
	{\includegraphics[width=0.24\textwidth]{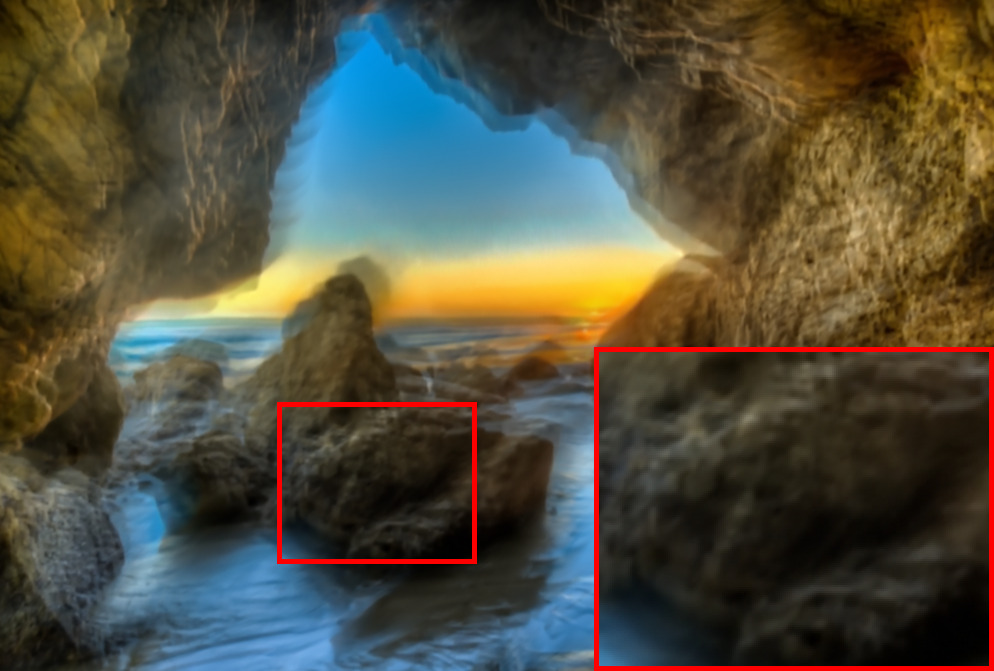}}\
	\subfloat[Ren \cite{ren2020neural}]
	{\includegraphics[width=0.24\textwidth]{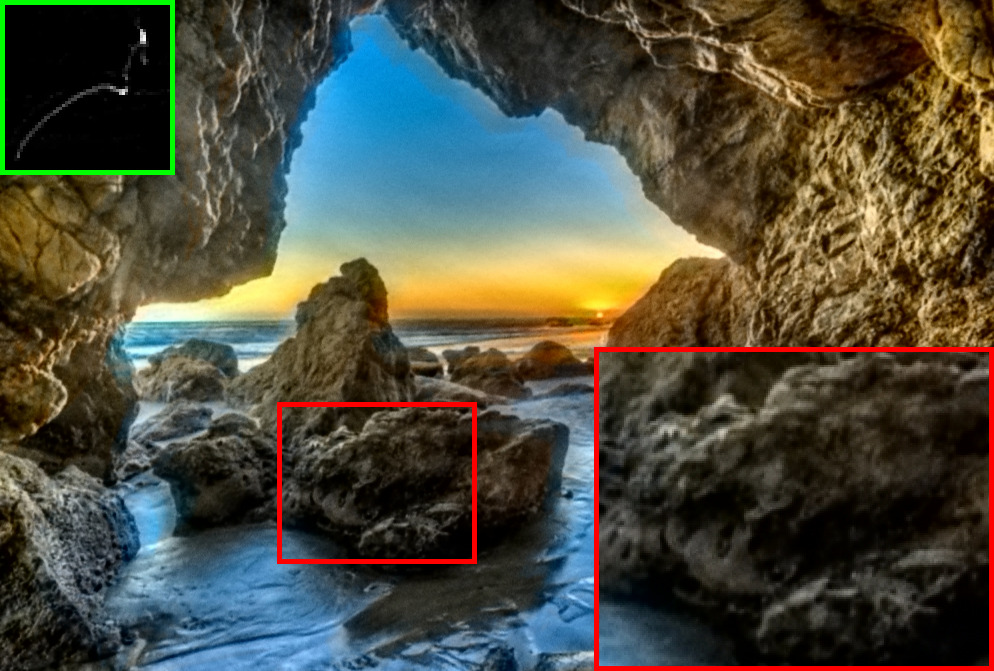}}
	\\
	\subfloat[Huo \cite{huo2023blind}]
	{\includegraphics[width=0.24\textwidth]{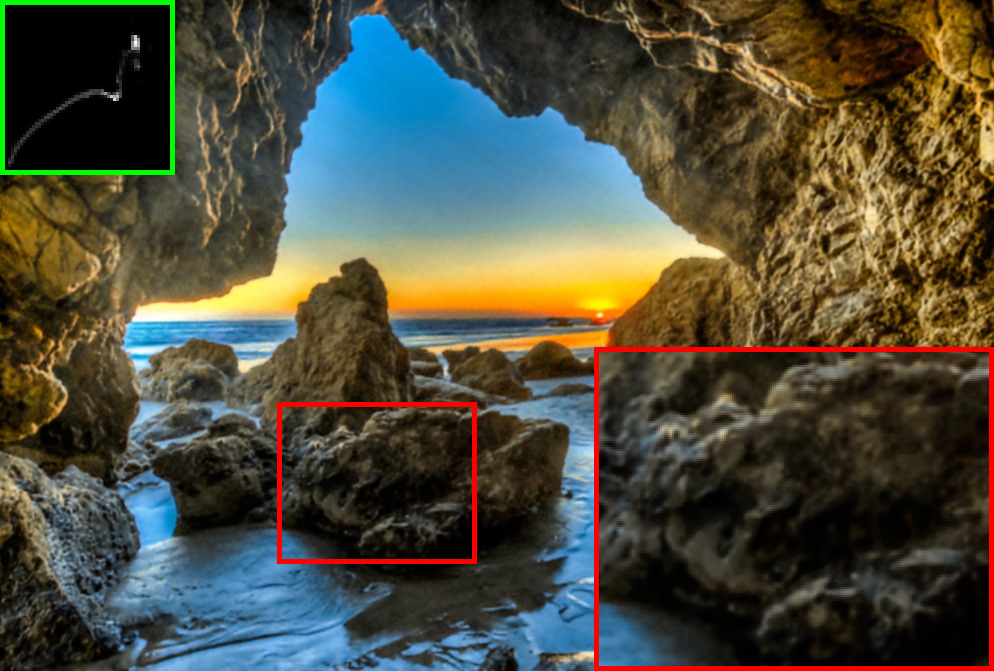}}\
	\subfloat[Li \cite{li2023self}]
	{\includegraphics[width=0.24\textwidth]{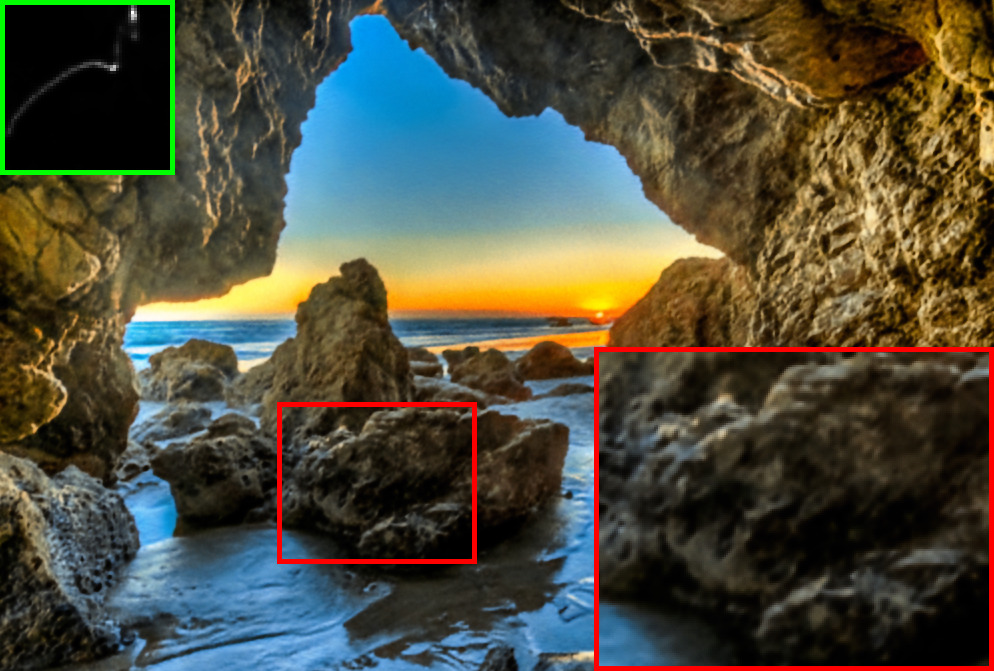}}\
	\subfloat[Ours]
	{\includegraphics[width=0.24\textwidth]{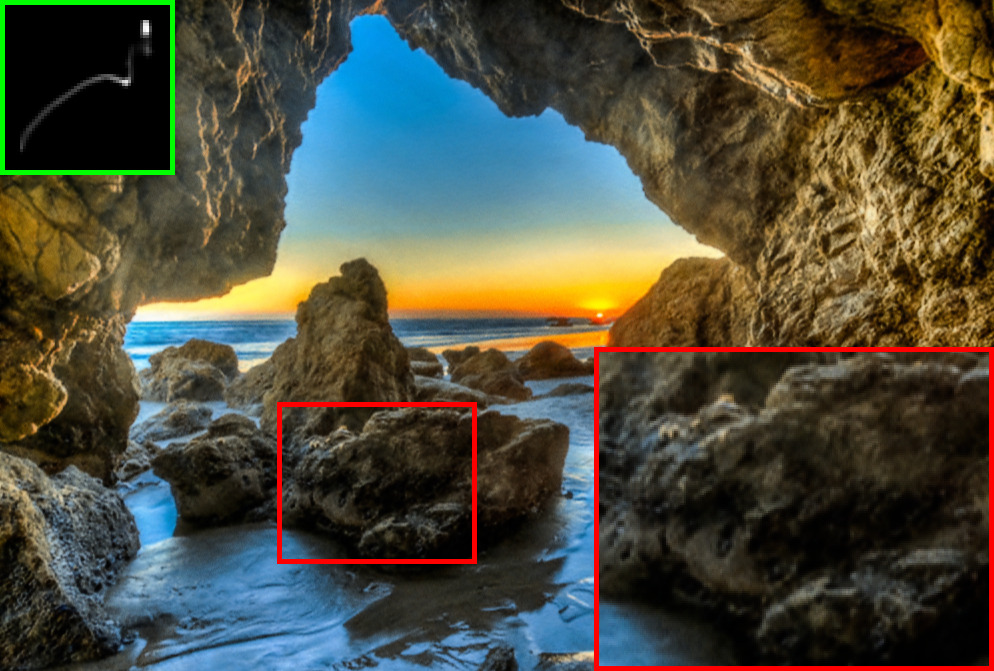}}\
	\subfloat[Ground truth]
	{\includegraphics[width=0.24\textwidth]{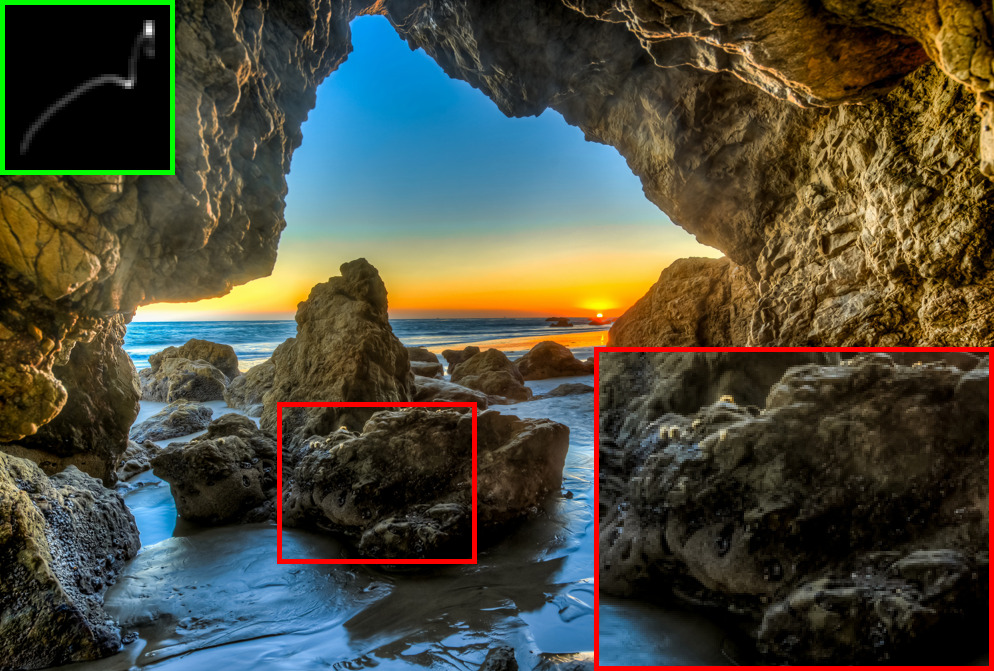}}
	\caption{Visual results on the dataset by Lai et al. \cite{lai2016comparative}. The estimated blur kernel is placed on the top-left corner of each method if available.}
	\label{fig_lai3}
\end{figure}

\begin{figure}[t]
	\centering
	\subfloat[Blurred]
	{\includegraphics[width=0.24\textwidth]{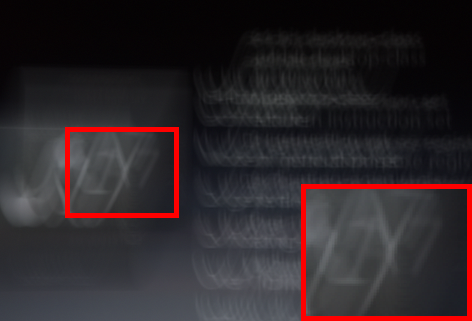}}\
	\subfloat[Cho \cite{cho2009fast}]
	{\includegraphics[width=0.24\textwidth]{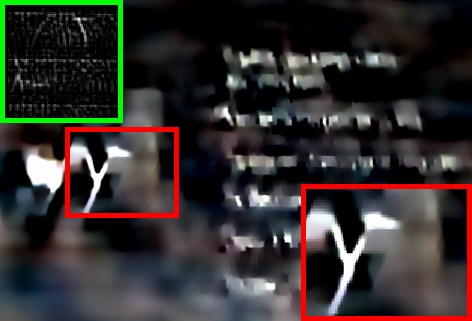}}\
	\subfloat[Krishnan \cite{krishnan2011blind}]
	{\includegraphics[width=0.24\textwidth]{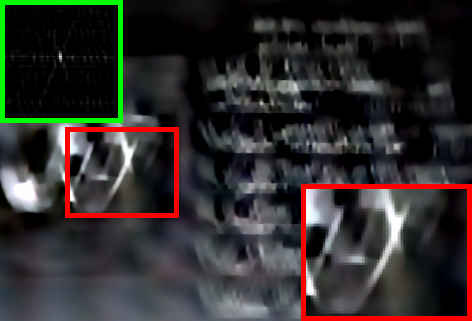}}\
	\subfloat[Xu \cite{xu2013unnatural}]
	{\includegraphics[width=0.24\textwidth]{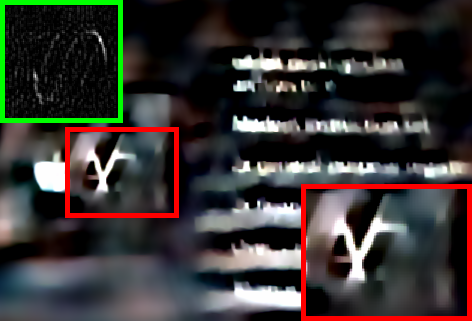}}
	\\
	\subfloat[Perrone \cite{perrone2014total}]
	{\includegraphics[width=0.24\textwidth]{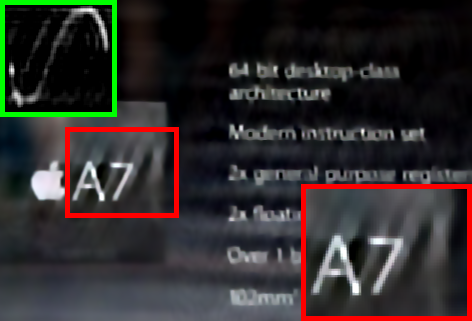}}\
	\subfloat[Pan \cite{pan2016blind}]
	{\includegraphics[width=0.24\textwidth]{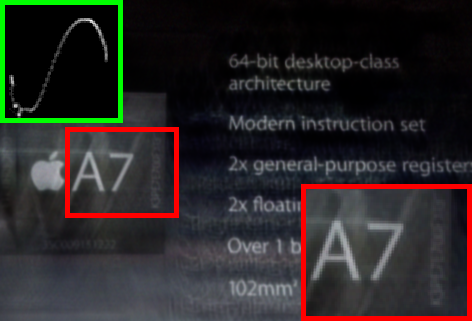}}\
	\subfloat[Dong \cite{dong2017blind}]
	{\includegraphics[width=0.24\textwidth]{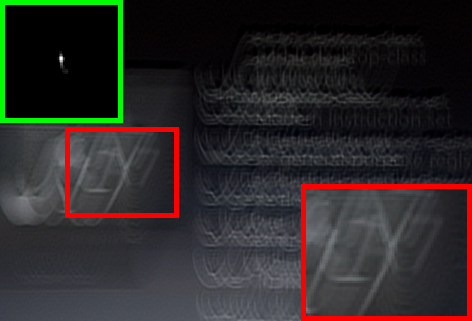}}\
	\subfloat[Tao \cite{tao2018scale}]
	{\includegraphics[width=0.24\textwidth]{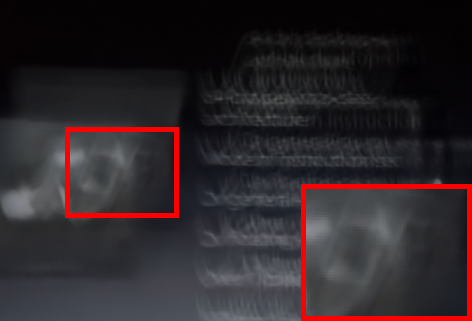}}
	\\
	\subfloat[Kupyn \cite{kupyn2019deblurgan}]
	{\includegraphics[width=0.24\textwidth]{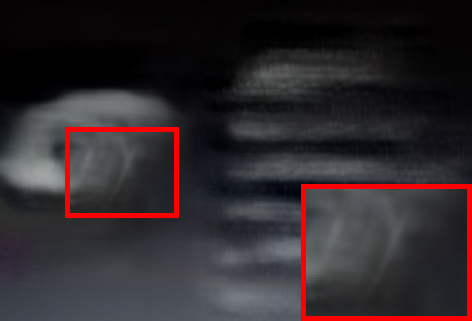}}\
	\subfloat[Kaufman \cite{kaufman2020deblurring}]
	{\includegraphics[width=0.24\textwidth]{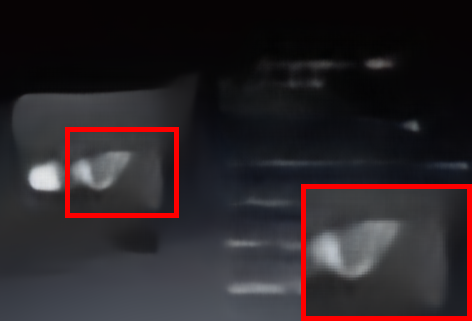}}\
	\subfloat[Zamir \cite{zamir2022restormer}]
	{\includegraphics[width=0.24\textwidth]{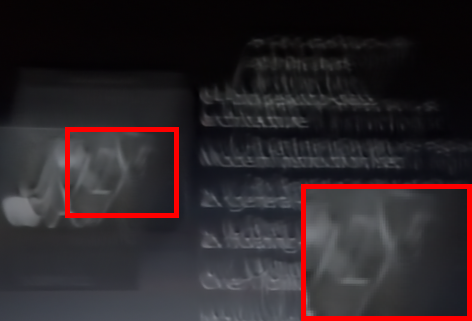}}\
	\subfloat[Ren \cite{ren2020neural}]
	{\includegraphics[width=0.24\textwidth]{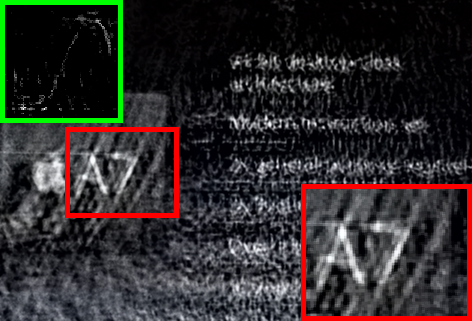}}
	\\
	\subfloat[Huo \cite{huo2023blind}]
	{\includegraphics[width=0.24\textwidth]{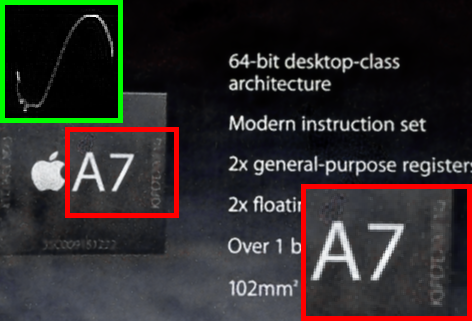}}\
	\subfloat[Li \cite{li2023self}]
	{\includegraphics[width=0.24\textwidth]{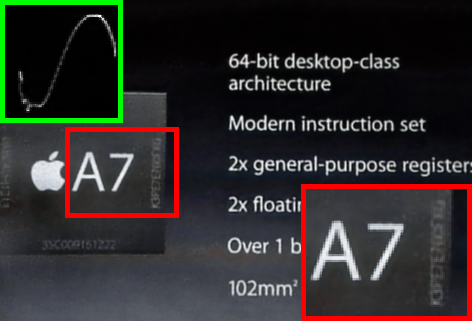}}\
	\subfloat[Ours]
	{\includegraphics[width=0.24\textwidth]{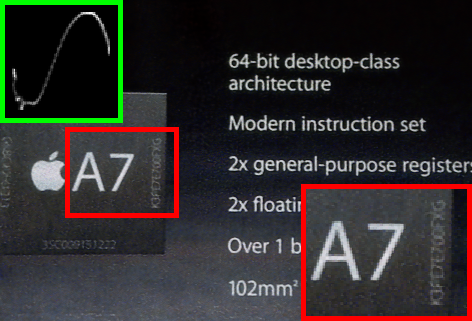}}\
	\subfloat[Ground truth]
	{\includegraphics[width=0.24\textwidth]{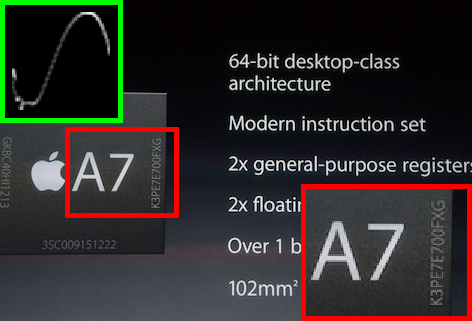}}
	\caption{Visual results on the dataset by Lai et al. \cite{lai2016comparative}. The estimated blur kernel is placed on the top-left corner of each method if available.}
	\label{fig_lai4}
\end{figure}

Figures. \ref{fig_syn2}-\ref{fig_syn4} provide visual results on our synthetic dataset, and Figs. \ref{fig_lai2}-\ref{fig_lai4} provide visual results on the benchmark by Lai et al. \cite{lai2016comparative}. The higher visual quality of the results produced by our method against other competing ones can be observed.

\section{Results and discussions on real blurry image}
\begin{figure}[htbp]
	\centering
	\subfloat[Blurred]
	{\includegraphics[width=0.24\textwidth]{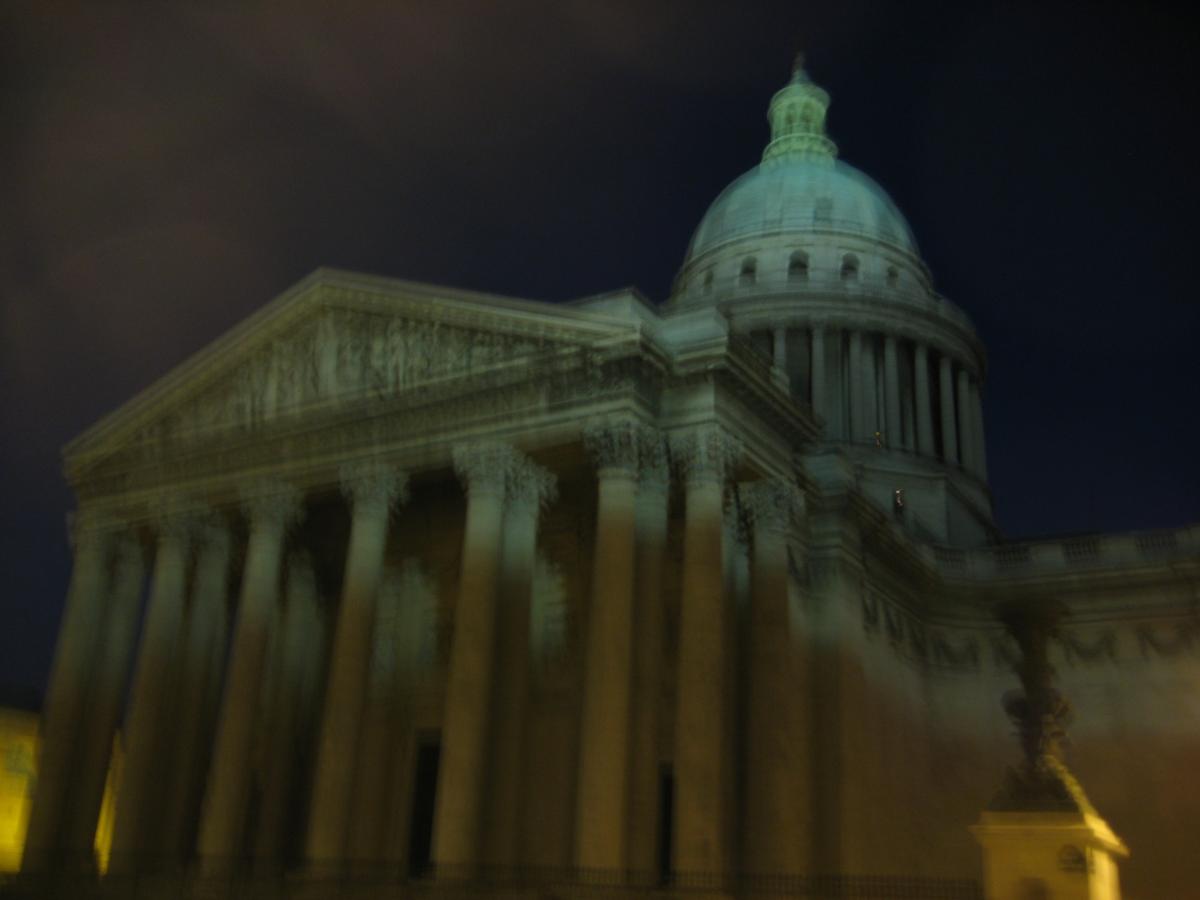}}\
	\subfloat[Pan \cite{pan2016blind}]
	{\includegraphics[width=0.24\textwidth]{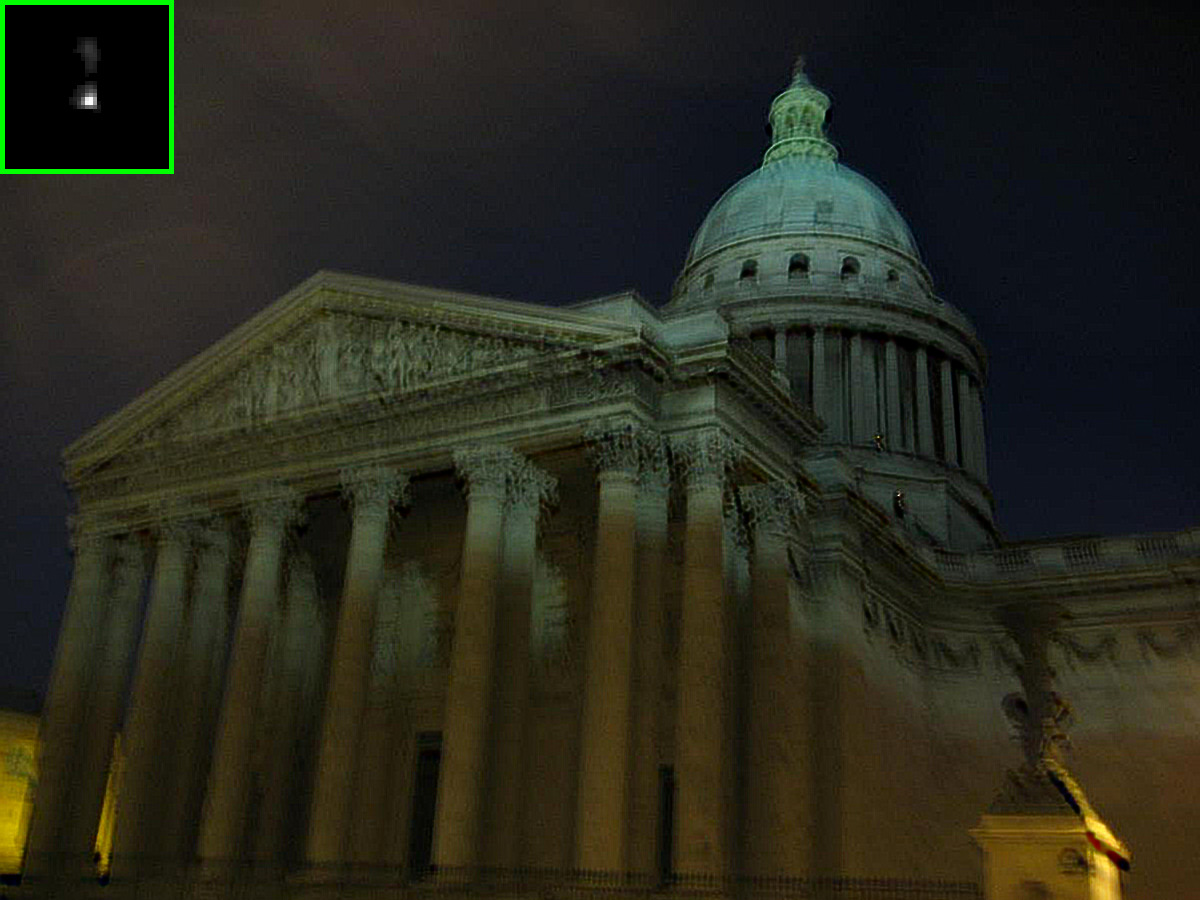}}\
	\subfloat[Dong \cite{dong2017blind}]
	{\includegraphics[width=0.24\textwidth]{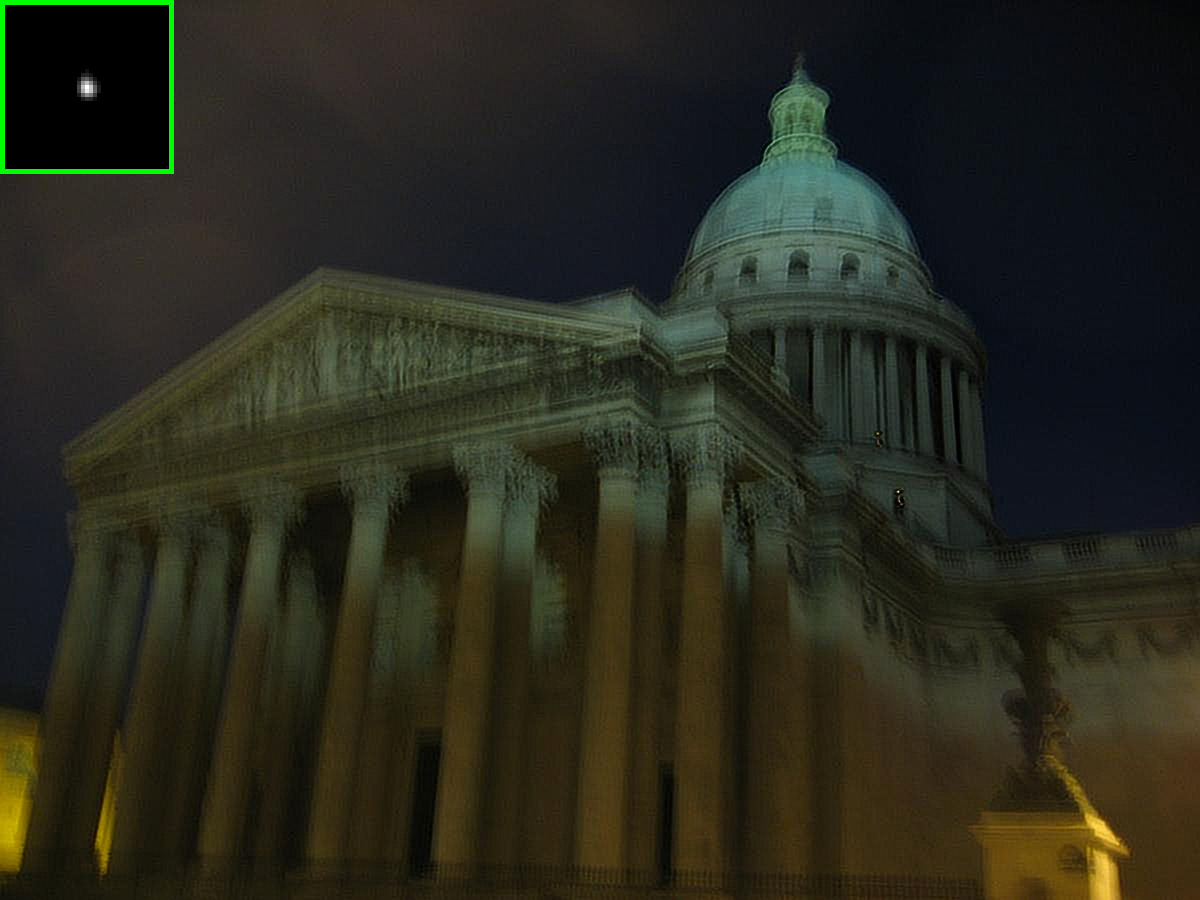}}\
	\subfloat[Tao \cite{tao2018scale}]
	{\includegraphics[width=0.24\textwidth]{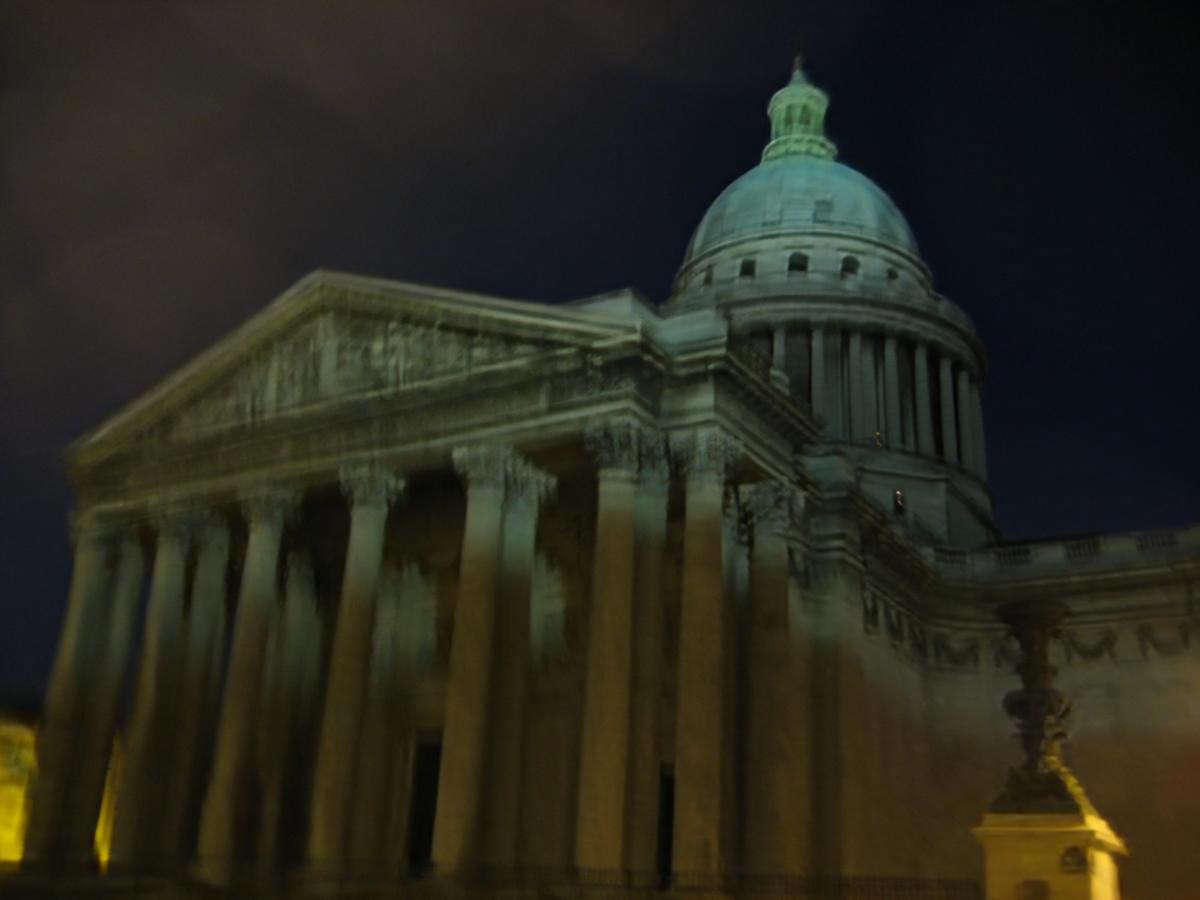}}
	\\
	\subfloat[Kupyn \cite{kupyn2019deblurgan}]
	{\includegraphics[width=0.24\textwidth]{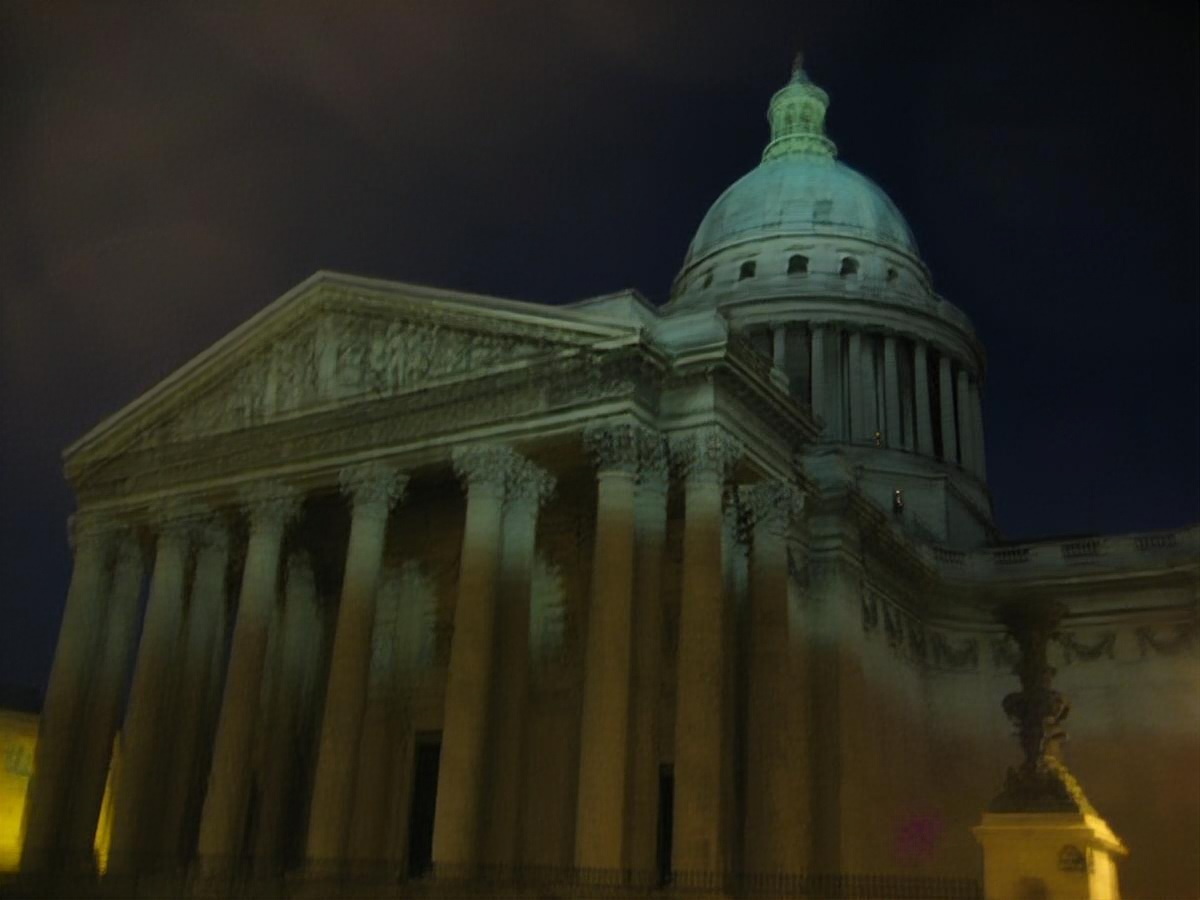}}\
	\subfloat[Kaufman \cite{kaufman2020deblurring}]
	{\includegraphics[width=0.24\textwidth]{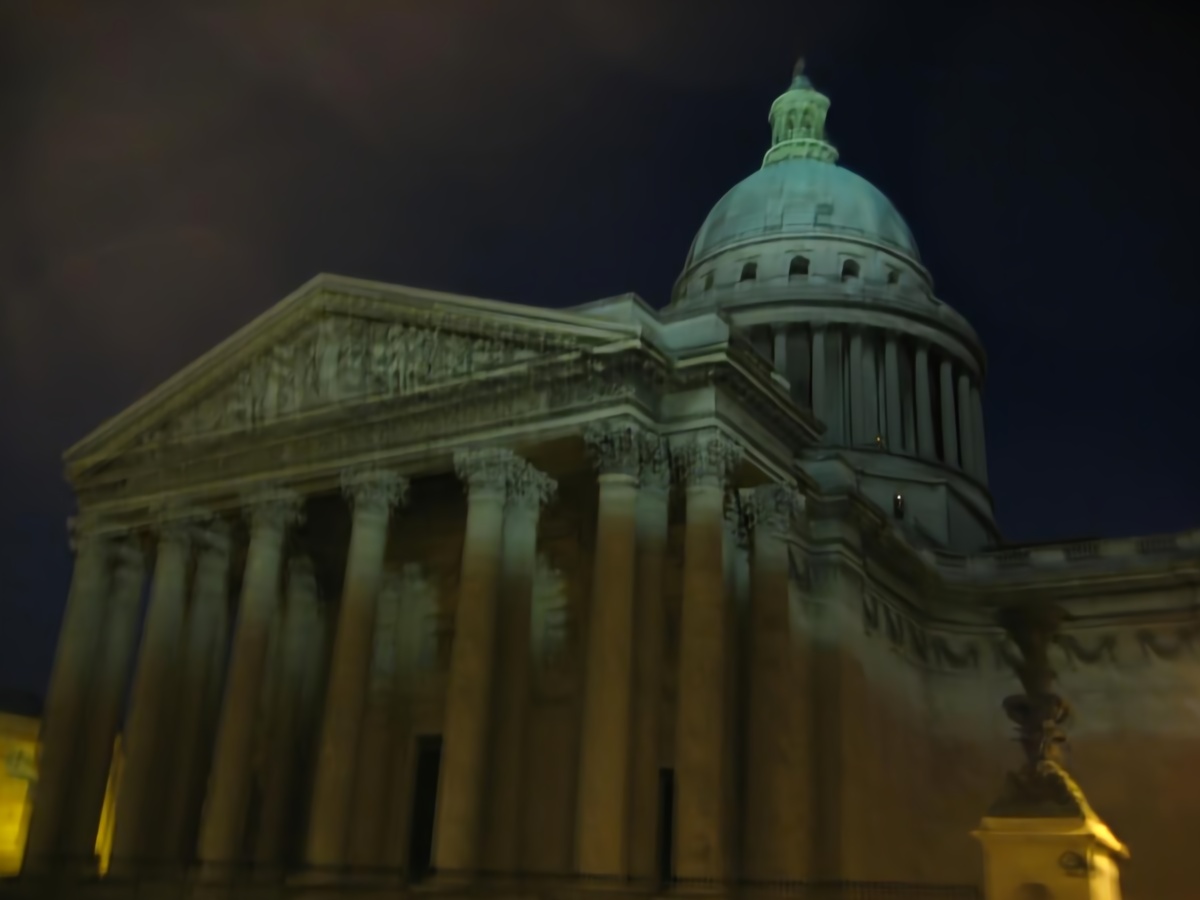}}\
	\subfloat[Zamir \cite{zamir2022restormer}]
	{\includegraphics[width=0.24\textwidth]{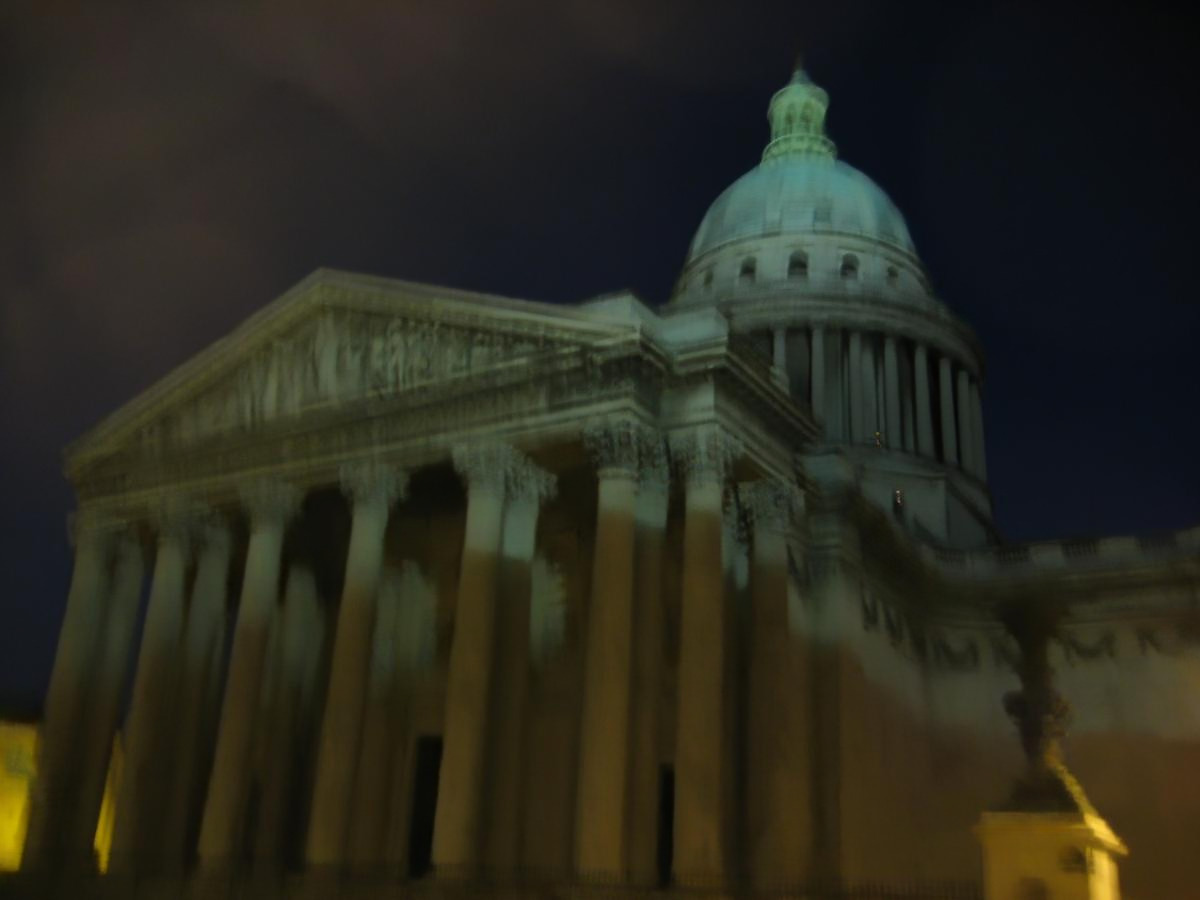}}\
	\subfloat[Ren \cite{ren2020neural}]
	{\includegraphics[width=0.24\textwidth]{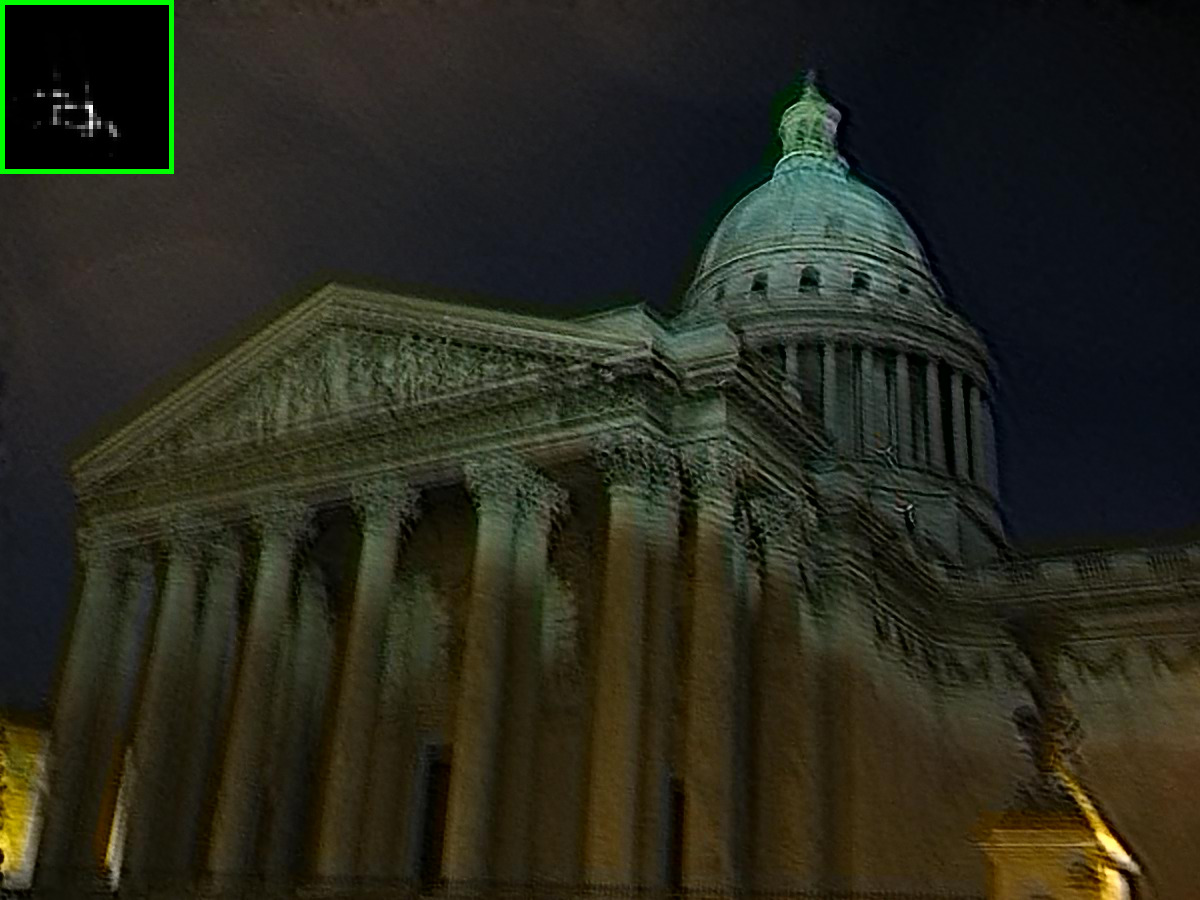}}
	\\
	\subfloat[Huo \cite{huo2023blind}]
	{\includegraphics[width=0.24\textwidth]{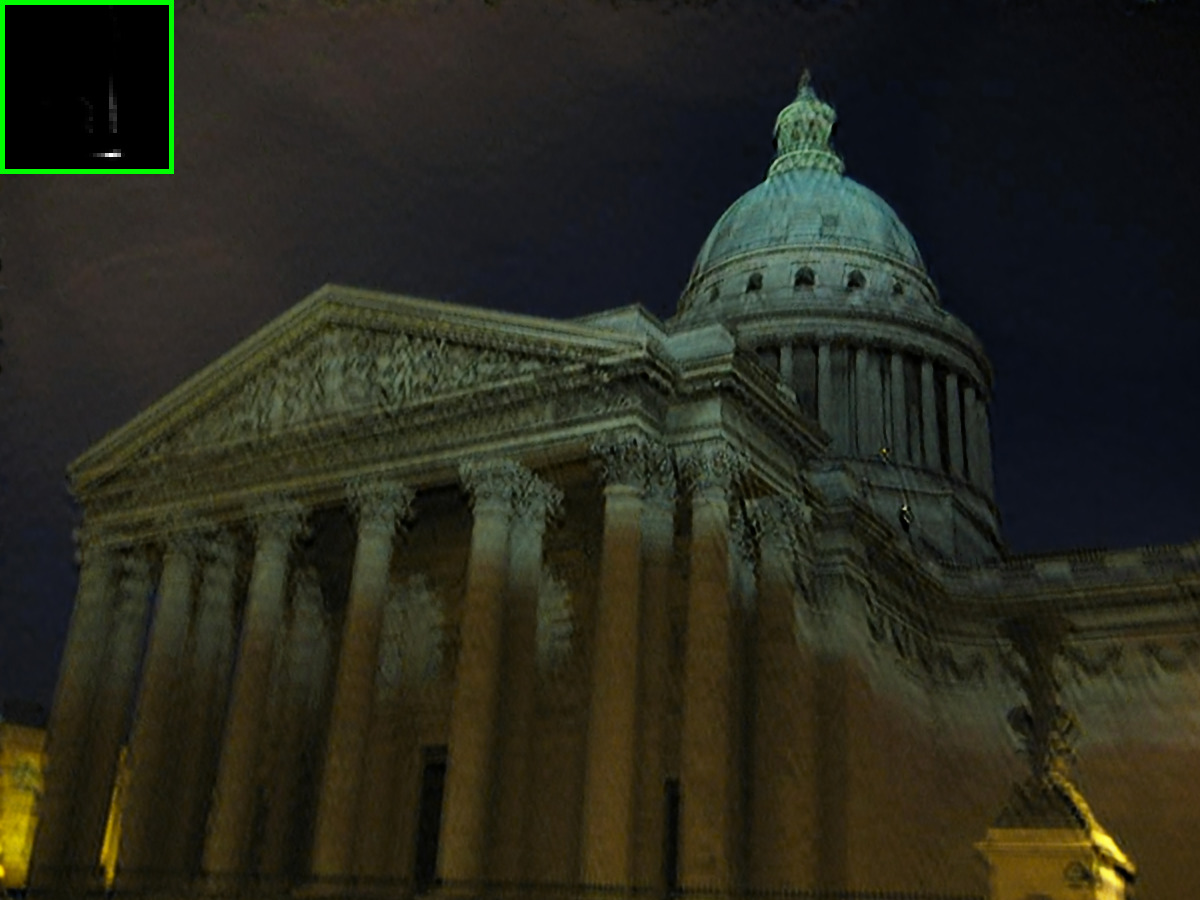}}\
	\subfloat[Li \cite{li2023self}]
	{\includegraphics[width=0.24\textwidth]{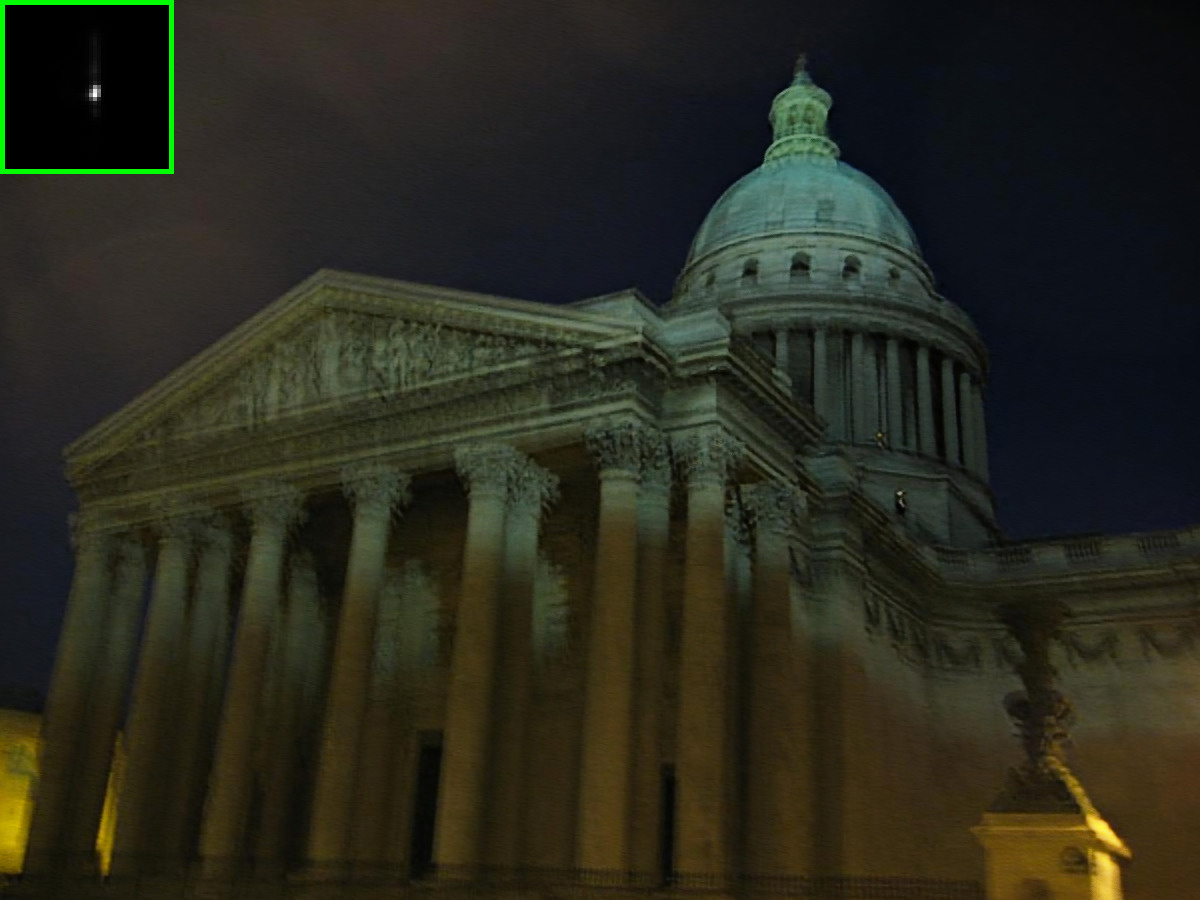}}\
	\subfloat[Ours]
	{\includegraphics[width=0.24\textwidth]{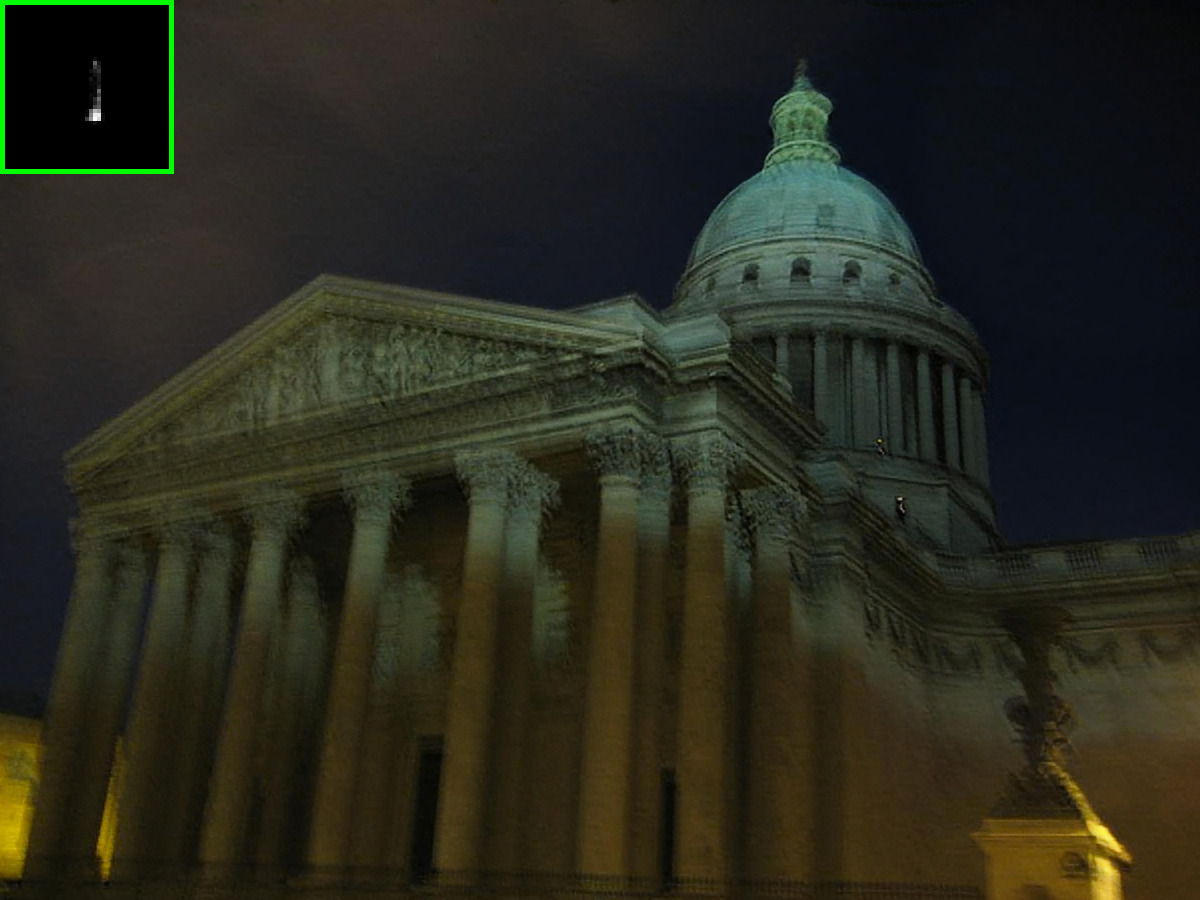}}
	\caption{Example deblur results of competing methods on the real blurry image provided by Lai et al. \cite{lai2016comparative}. The estimated blur kernel is placed on the top-left corner of each method if available.}
	\label{fig_lai_real2}
\end{figure}

In addition to the example shown in the main text, we show BID results of comparison methods on another real blurry image by Lai et al. \cite{lai2016comparative} in Fig. \ref{fig_lai_real2}. It can be seen from this example and the one in the main text that, on these very challenging images, none of the methods generate perfect results. Specifically, the results of traditional model-based methods are with severe ghost effects, while supervised deep learning methods over-smooth the image. Comparatively, deep prior-based methods generate better results, especially in recovering image details. Among them, our method has relatively higher or at least comparable visual quality, showing its potential in dealing with blurry images that are with unknown complex blur kernels.

\section{Limitations}\label{sec:limitation}
Though the proposed method has achieved better performance compared to existing approaches, it still has limitations. One limitation is that the pre-training stage in the proposed framework requires a large amount of training data, including the synthesized blur kernels for training the kernel generator, and also the blur kernel-image pairs for the kernel initializer. This can result in two drawbacks: one is that the training cost could be large, and the other is that the generalization performance on the unseen types of blur kernels might be limited. These drawbacks could be alleviated by using the meta-learning strategy \cite{maml} for few-shot learning, with which we can pre-train base models, i.e., the kernel generator and kernel initializer, and then fine-tune them in the few-shot setting to let them fast adapt to specific deblur scenarios.

Another limitation is about the BID model (Eq. (6) of the main text). Specifically, in this model, we consider DIP as the only prior for the to-be-deblurred image, while ignoring other image priors that have been exhaustively explored in traditional image reconstruction problems, such as smoothness. This drawback can limit the performance of our method. For example, as shown in Fig. 8 of the main text, though looks better than that of other competing methods, the result of our method still contains unexpected non-smooth areas. This limitation could be alleviated by more precise prior modeling for the image, and the strategy used in \cite{sr_yue}, for instance, can be considered.

Besides, our method is also limited by its blurred assumption. In this work, we only consider the case that the blurry image is degraded by uniform motion blur, while the blurring mechanism could be much more complex in real scenarios. In fact, we have tried to apply our method to the dataset provided by K{\"o}hler \etal \cite{kohler_dataset}, within which the non-uniform blur is involved, but failed to obtain satisfactory results. Nevertheless, it is possible to adopt the strategy proposed by Li \etal \cite{li2023self} to generalize our method to the non-uniform blur situations, which is a future direction of our study.

\end{document}